\documentclass[10pt]{article} 
\usepackage[preprint]{tmlr}
\usepackage{microtype}
\usepackage{graphicx}
\usepackage{subcaption}
\usepackage{booktabs} 
\usepackage{multirow} 
\usepackage{booktabs}
\usepackage{longtable}
\usepackage{siunitx}
\usepackage{hyperref}
\usepackage{wrapfig}
\usepackage{float}
\usepackage{tikz}
\usetikzlibrary{shapes.geometric, positioning, arrows.meta, calc}

\usepackage{algorithm,algpseudocode}

\usepackage{amsmath}
\usepackage{amssymb}
\usepackage{mathtools}
\usepackage{amsthm}
\usepackage{makecell}

\usepackage[utf8]{inputenc} 
\usepackage[T1]{fontenc}    
\usepackage{hyperref}       
\usepackage{url}            
\usepackage{booktabs}       
\usepackage{amsfonts}       
\usepackage{nicefrac}       
\usepackage{microtype}      
\usepackage{xcolor}         

\newcommand{\df}{\delta_f}
\newcommand{\dl}{\delta_l}

\title{Invertible continuous latent dynamic for long-term data assimilation in complex physical systems}

\author{%
\begin{minipage}[t]{0.24\textwidth}
\centering
\textbf{Rares Grozavescu}$^{*}$\\[-1pt]
\small University of Cambridge\\[-1pt]
\small\textit{rg625@cam.ac.uk}
\end{minipage}
\hfill
\begin{minipage}[t]{0.24\textwidth}
\centering
\textbf{Etienne Meunier}$^{*}$\\[-1pt]
\small Inria, Paris\\[-1pt]
\small\textit{etienne.meunier@inria.fr}
\end{minipage}
\hfill
\begin{minipage}[t]{0.24\textwidth}
\centering
\textbf{Pengyu Zhang}\\[-1pt]
\small University of Cambridge\\[-1pt]
\small\textit{pz281@cam.ac.uk}
\end{minipage}
\hfill
\begin{minipage}[t]{0.24\textwidth}
\centering
\textbf{Mark Girolami}\\[-1pt]
\small University of Cambridge\\[-1pt]
\small\textit{mag29@cam.ac.uk}
\end{minipage}%
}

\begin{document}

\maketitle
\begin{abstract}
Forward forecasting and data assimilation are the two important aspects in physical simulation: one propagates the state forward, the other recovers unknown states from sparse observations. Learned surrogates are normally built and benchmarked for forward forecasting, however, whether a surrogate could attain good performance in data assimilation tasks is valuable as well, as inverse problems are of paramount importance in the scientific domain. In this paper, we propose a continuous-time Koopman autoencoder whose latent dynamics obey $\frac{dz}{dt} = \mathbf{K}_{\mathrm{cont}} z$, yielding closed-form inference via $z(\tau) = \exp(\mathbf{K}_{\mathrm{cont}} \tau) z(0)$ at any horizon $\tau$ in a single step. This decouples forecast cost from forecast length at inference time, showing long-term stability and high efficiency in forward simulation, and also supports data assimilation as gradient-based optimization with cost independent of the assimilation window. Experiments are performed on the Kuramoto–Sivashinsky equation and a transient flow, and we compare our method against a range of baselines on the forward problem, including diffusion models and operator-learning models, and obtain a 110$\times$ inference speedup over strong diffusion baselines. We further test these baselines on an initial-state inference data assimilation task, and find that a strong forecaster does not guarantee a strong assimilator, while the continuous-time Koopman autoencoder achieves both higher accuracy and efficiency than surrogates of comparable forward performance.
\end{abstract}

\section{Introduction}

There are two complementary problems in physical simulation. The forward problem focuses on forecasting the long-term evolution of physical systems, such as turbulent flows and atmospheric dynamics, which requires models that remain stable over hundreds of steps. Traditional numerical solvers \citep{ghase2017les,slotnick2014cfd} are prohibitively expensive at these horizons. Data-driven surrogates based on autoregressive rollouts \citep{li2021fourier,kohl2026benchmarking} provide a cheaper alternative, but accumulate errors step-by-step, with inference cost scaling linearly with forecast length. Data assimilation aims to recover unknown states from sparse observations. Classical approaches use a numerical solver as the forward model and are correspondingly slow, while learned forward surrogates inherit the same autoregressive rollout and error accumulation \citep{tang2020deep,chattopadhyay2022towards}.

To address this, we propose a continuous-time Koopman autoencoder based on Koopman theory \citep{koopman1931hamiltonian}. By embedding the nonlinear state into a latent space with linear dynamics, $dz/dt = \mathbf{K}{\mathrm{cont}} z$, the state at any future time $\tau$ is given by $z(\tau) = \exp(\mathbf{K}{\mathrm{cont}} \tau) z(0)$—a single matrix-vector product regardless of horizon. This enables data assimilation as a single-step gradient-based optimization for initial-state recovery, with cost independent of $\tau$. Unlike prior Koopman methods \citep{lusch2018koopman,azencot2020consistent}, which learn a discrete operator tied to a fixed $\Delta t$, our continuous-time formulation naturally handles irregularly sampled observations without resampling or interpolation.

In this paper, we study forward forecasting and data assimilation together, comparing a range of models on both tasks. On the forward problem, we evaluate forecasting accuracy, inference speed, and long-term stability; on the data assimilation problem, we evaluate initial-state recovery across a range of settings. Our contributions are:

  \begin{itemize}
      \item We introduce a continuous-time Koopman autoencoder whose linear latent dynamics
      admit a closed-form solution, enabling single-step inference at any horizon. We
      demonstrate that it handles irregular sampling and unseen time steps.
      \item On realistic fluid flows and chaotic systems, it remains stable over long rollouts
      at a fraction of the inference cost of diffusion baselines.
      \item We demonstrate that our model enables fast and accurate data assimilation,
      outperforming autoregressive and score-based baselines at initial-state recovery with
      a cost independent of the assimilation window.
  \end{itemize}

\section{Related work}

\textbf{Koopman autoencoders.} Koopman theory \citep{koopman1931hamiltonian} represents nonlinear dynamics as a linear operator acting on observables, while Koopman autoencoders (KAEs) learn these observables with an autoencoder whose latent dynamics are linear \citep{takeishi2017learning,lusch2018koopman,morton2018deep,otto2019linearly}. Many variants have since been proposed \citep{azencot2020consistent,nayak2025tckae,xiong2024koopman,halder2026reduced}, but most learn a discrete operator $\mathbf{A}$ tied to the training timestep $\Delta t$. Continuous-time latent dynamics, as in neural ODEs \citep{chen2018neural}, allow irregular time steps but typically require numerical integration. In contrast, our linear latent dynamics admit a closed-form solution. Related continuous-time Koopman models \citep{menier2025interpretable,buzhardt2025relationship} either retain nonlinear terms requiring integration or focus on highly idealized, low-dimensional systems.

\textbf{Forward surrogates and data assimilation.} Neural operators such as FNO \citep{li2021fourier} and convolutional U-Nets \citep{ronneberger2015u} provide efficient autoregressive forecasting, but error accumulation limits long-horizon reliability \citep{brandstetter2022message}. Diffusion models \citep{kohl2026benchmarking} improve sample quality but incur stochastic error accumulation and high per-step cost. Learned surrogates have also been used to replace numerical solvers in ensemble and variational data assimilation \citep{chattopadhyay2022towards,penny2022integrating,adrian2025data,hatfield2021building}. \citet{frion2023leveraging,frion2024neural} showed that a Koopman autoencoder can provide an effective prior for variational assimilation, but focused on low-dimensional, near-periodic satellite time series with a regularized discrete-time operator.

\section{Background on latent models and data assimilation}
\label{sec:background}

\subsection{Continuous-Time Koopman Autoencoders}
Koopman operator theory shifts focus from nonlinear state-space evolution $x_{t+\Delta t} = f(x_t)$, $x_t \in \mathbb{R}^{N_d}$, to an infinite-dimensional space of measurement functions where dynamics are linear. Since finding a finite invariant subspace analytically is intractable for chaotic PDEs, Koopman Autoencoders (KAEs) learn this mapping from data: an encoder $\mathcal{E}$ maps the state to a latent representation $z_t \in \mathbb{R}^{N_z}$, a finite-dimensional operator propagates the latent dynamics, and a decoder $\mathcal{D}$ reconstructs the physical state.

Standard discrete-time KAEs learn a fixed-step operator $\mathbf{A}$ such that $z_{t+\Delta t} = \mathbf{A}z_t$, tying the latent dynamics to the training resolution and requiring regular sampling. Our approach instead trains the continuous generator directly, with latent dynamics governed by a linear ODE whose exact solution is $z(t_0+\tau)=e^{\mathbf{K}_{\text{cont}}\tau}z(t_0)$. This closed-form solution enables single-step inference at any horizon, independent of the training timestep, and naturally accommodates irregularly sampled observations. A full derivation is provided in Appendix~\ref{appendix:koppman_derivation}.

\subsection{Data assimilation with a latent based model}

In data assimilation, we infer a physical state $x \in \mathbb{R}^{T \times P}$ from observations $y = H(x) + \varepsilon$, where $H$ is the observation operator and $\varepsilon$ the observation noise. This can be formulated as Bayesian inference over a \emph{control variable} $c$, which determines the full state through a mapping $c \mapsto x(c)$. The posterior is
\begin{equation}
\log p(c \mid y) = \log p(y \mid x(c)) + \log p(c) - \log p(y),
\end{equation}
where the first two terms correspond to the likelihood and prior.

We consider two main families of data assimilation methods, which differ primarily in their choice of control variable and mapping $c \mapsto x(c)$.

\textbf{Score model based assimilation.} Score-based methods \cite{rozet2023score, chung2022diffusion} use a noisy version of the full spatio-temporal state as the control variable, $c \triangleq x_\sigma$. A generative model provides the prior score $\nabla_c \log p_\sigma(c)$, which is combined with the likelihood gradient to steer the state towards the posterior.

\textbf{Simulation based assimilation.} Simulation-based methods instead use a low-dimensional control variable, typically the initial state $x_0$, which a forecasting model maps to the full trajectory $x \in \mathbb{R}^{T \times P}$. This includes 4D-Var \citep{bouttier1999four,navon2009data}, using either numerical solvers \citep{courtier1994strategy} or learned emulators \citep{fablet2021learning}. Optimisation is performed on $x_0$, with gradients propagated through the forecast. Although effective, this can be computationally expensive and suffer from gradient instabilities over long assimilation windows.

Emulator-based assimilation also imposes an \emph{implicit prior}: the assimilated states are restricted to those reachable by the forecasting model. This can enforce physically consistent trajectories, but limits assimilation to states within the model’s predictive range.

\begin{figure}[H]
\centering
\includegraphics[width=0.8\linewidth]{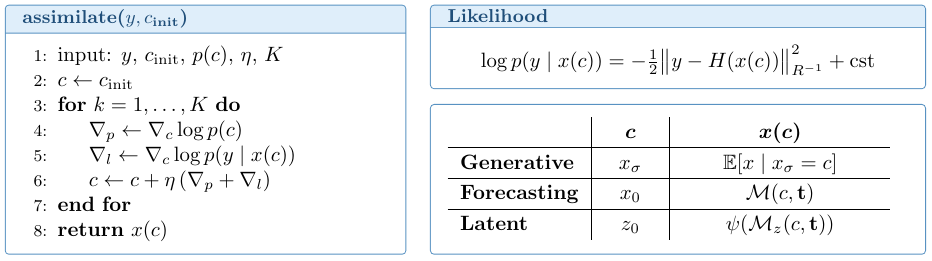}
\vspace{-6pt}
\caption{\small Optimisation backbone shared by the three assimilation methods (left), and their choice of control variable $c$ and mapping $x(c)$ (right).}
\vspace{-10pt}
\label{alg:da}
\end{figure}

\textbf{Optimisation for data assimilation.} All three approaches share the same optimisation backbone (Algorithm~\ref{alg:da}), updating the control variable using the prior and likelihood gradients. They differ only in the choice of control variable and mapping $c \mapsto x(c)$. For emulator-based assimilation, the central challenge is therefore computing gradients through the forecasting model, which we discuss next.

\subsection{Sensitivity in forecasting models}\label{sec:sensitivity}

As described above, forecasting-based assimilation requires the gradient of the log-likelihood $\log p(y \mid x(c))$ with respect to the control variable, which means differentiating through the forecasting model $\mathcal{M}$. By the chain rule, this gradient is the product of the likelihood gradient with respect to the state and the \emph{sensitivity} of the forecast to its initial condition, $\partial x_t / \partial x_0$. Beyond data assimilation, forecast sensitivities are used in many applications, for instance to estimate the impact of individual observations on a forecast \citep{langland2004estimation}, to compute optimal perturbations (singular vectors) for ensemble forecasting \citep{buizza1995singular}, or to quantify the uncertainty of an estimated initial state.

Sensitivities can be computed for any differentiable autoregressive model with backpropagation through time, but a latent linear model has a strong advantage for this task. For an autoregressive model $x_{k+1} = f(x_k)$ with time step $\Delta t$, and a Koopman model with encoder $\phi$, decoder $\psi$ and continuous latent dynamic $z_t = e^{At} z_0$, $z_0 = \phi(x_0)$, the sensitivities at time $t = K \Delta t$ can be written as : 
$$
\underbrace{\frac{\partial x_t}{\partial x_0} = J_f(x_{K-1}) \, J_f(x_{K-2}) \cdots J_f(x_0)}_{\text{autoregressive}}
\qquad \qquad
\underbrace{\frac{\partial x_t}{\partial x_0} = J_\psi(z_t) \, e^{At} \, J_\phi(x_0)}_{\text{Koopman}}
$$
The Koopman formulation does not involve a product of nonlinear Jacobians: the dependence on the horizon is entirely carried by the matrix exponential, evaluated once, while the encoder and decoder are only differentiated at the two ends of the trajectory. This has two consequences. First, \emph{the cost of a sensitivity query does not grow with the number of forecast steps}, where backpropagation through an autoregressive rollout has a time and memory cost linear in $K$. Second, long products of Jacobians are prone to exploding or vanishing gradients \citep{bengio1994learning, pascanu2013difficulty}, and small errors in the one-step derivative  compound multiplicatively along the rollout. The Koopman model is not immune to unstable gradients, but they are entirely controlled by the spectrum of $A$: constraining the real parts of its eigenvalues  is much easier than controlling the product of $K$ state-dependent Jacobians.

\subsection{Illustrative experiment}

We illustrate these differences on the double-well system $\dot{x} = r \, (x - x^3)$, whose flow and sensitivity are known in closed form. We train three emulators with the same rollout loss, without any supervision on derivatives: an autoregressive MLP $x_{k+1} = f_\theta(x_k)$, its residual variant $x_{k+1} = x_k + \Delta t \, f_\theta(x_k)$, and a Koopman model with a continuous latent space (Section~\ref{alg:kae}).

\begin{figure}[t]
    \centering
    \includegraphics[width=\linewidth]{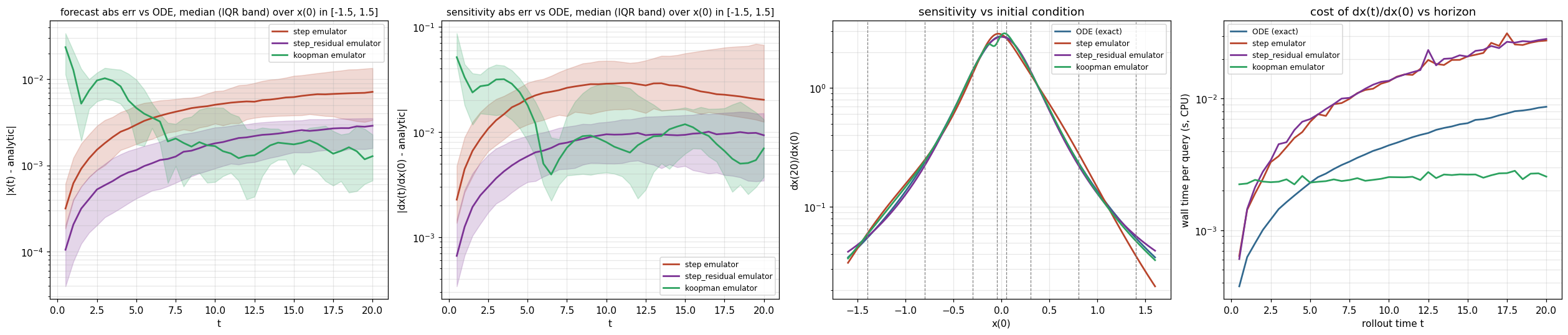}
    \caption{\small Autoregressive, residual autoregressive and Koopman emulators on the double-well system. From left to right: forecast error and sensitivity error over time; sensitivity as a function of the initial condition; cost of a sensitivity query as a function of the horizon.}
    \label{fig:demograd}
\end{figure}

As shown in Figure~\ref{fig:demograd}, the three models reach a similar forecasting accuracy and recover the analytic sensitivity well. On this simple dissipative system, the approaches thus give similar gradients, but at a very different cost: the cost of a sensitivity query grows linearly with the horizon for autoregressive models, while it stays constant for the Koopman model. On longer rollouts or chaotic systems, however, the product of one-step Jacobians is expected to suffer from vanishing or exploding gradients \citep{bengio1994learning, pascanu2013difficulty, mikhaeil2022difficulty}, whereas for the Koopman model this issue reduces to controlling the spectrum of $A$. In paragraph \ref{exp:da}, we show both this gap in computational cost and a gap in assimilation performance.

\section{Method}
\textbf{Training.}\label{subsection:training}
The training procedure follows a recurrent rollout scheme. Given an input sequence context $\{ x_{t_{i-1}}, x_{t_i} \}$, the model predicts a future trajectory of length $N$, denoted as $\{ \hat{x}_{t_{i+1}}, \hat{x}_{t_{i+2}}, \dots, \hat{x}_{t_{i+N}} \}$. A detailed architecture overview is provided Figure~\ref{fig:architecture}. The training objective is defined as a weighted combination of several loss components that enforce accurate reconstruction, long-horizon prediction, latent-space consistency, and structural regularization. All losses are computed as mean squared errors (MSE) unless stated otherwise.

\textbf{Overall objective.}
The final training objective is given by
\begin{equation}
\label{eq:total_loss}
\mathcal{L}_{\mathrm{total}}
=
\mathcal{L}_{\mathrm{recon}}
+
\alpha\,\mathcal{L}_{\mathrm{pred}}
+
\beta\,\mathcal{L}_{\mathrm{latent}}
+
\lambda_{\mathrm{phys}}\,\mathcal{L}_{\mathrm{phys}},
\end{equation}
where $\alpha$ and $\beta$ control the relative contributions of the rollout prediction and latent consistency losses, respectively, and $\lambda_{\mathrm{phys}}$ weights the optional physics-conditioned regularization terms. More details about how the hyperparameters were chosen can be found in Appendices~\ref{app:loss_wights} to ~\ref{app:phys_loss}. An ablation study removing different components of the loss can be found in Table~\ref{tab:da_ablation}

\textbf{Inference and Integration via Matrix Exponentiation.} \label{sec:inference_exp} A major advantage of restricting the latent dynamics to a linear Koopman form is that, at inference time, we can leverage the analytical solution of the learned system through matrix exponentiation. This bypasses iterative ODE solvers entirely and yields efficient long-horizon forecasting. Given the linear system in \eqref{eq:method_ode}, the exact solution for any future state $z_\tau$ given $z_0$ is defined by the matrix exponential $z_{\tau} = \exp(\mathbf{K}_{\text{cont}}\tau) z_0$. This formulation allows us to predict the state at an arbitrary future time $\tau$ in a single computational step, independent of the training step size $\Delta t$, providing our model the ability to perform zero-shot temporal super-resolution as well.

\section{Experiments and Results}


\begin{figure}[t]
    \centering
    \begin{subfigure}[t]{0.53\linewidth}
        \centering
        \includegraphics[width=\linewidth]{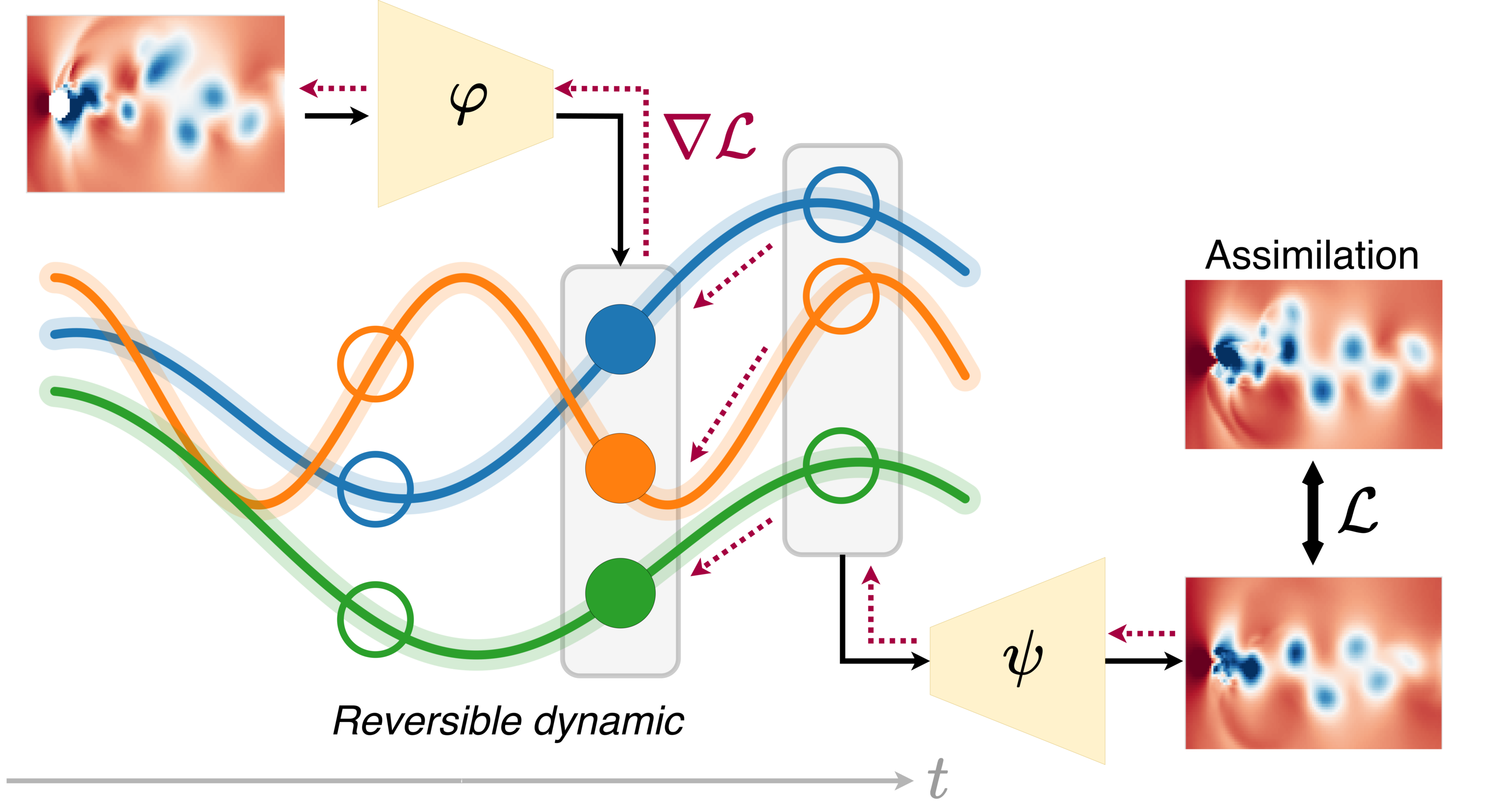}
        \label{fig:boxplot}
    \end{subfigure}
    \hfill
    \begin{subfigure}[t]{0.45\linewidth}
        \centering
        \includegraphics[width=\linewidth]{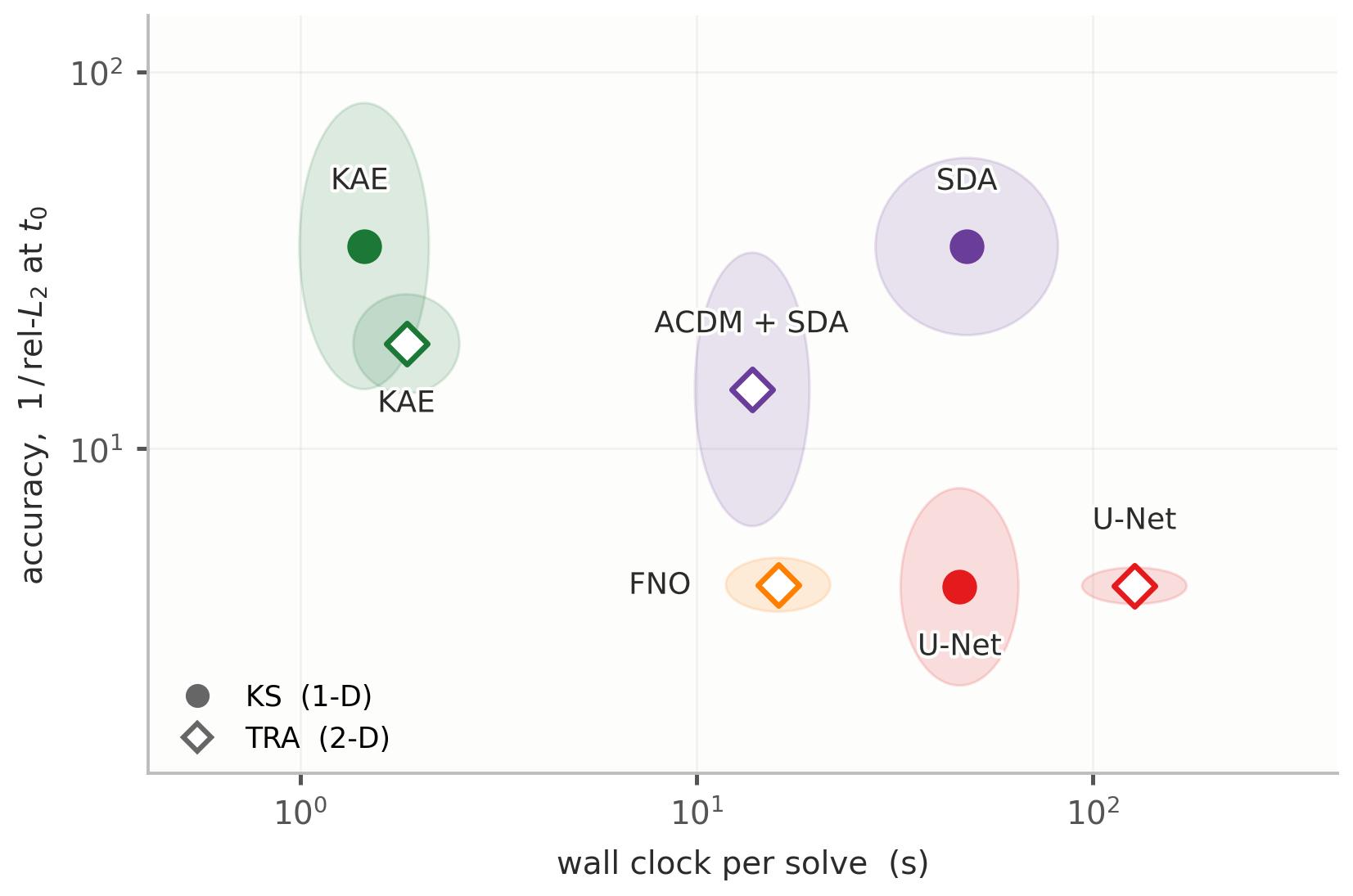}
        \label{fig:cost-accuracy}
    \end{subfigure}
    \caption{Left : summary figure of the method, $\varphi$ represent the encoder network from the physical space to continuous latent space and $\psi$ the decoder from latent space to observation. Forward prediction path is depicted as black arrow and gradients for a Cost against accuracy for every method in both regimes. Each marker is an aggregate over the whole set of experiments at $\df=9$. The spread is variation \emph{across conditions}, and its horizontal extent is set mainly by the $\dl$.}
    \label{fig:cost_accuracy_both}
\end{figure}

\paragraph{Is KAE a stable forward surrogate?}

We first evaluate the models as forward surrogates. We consider two fluid benchmarks from Autoregressive Conditional Diffusion Models (ACDM) \citet{kohl2026benchmarking}: incompressible wake flow ($Re \in [100,1000]$) and transonic cylinder flow ($Ma \in [0.50,0.90]$). Models predict $N=8$ future states from two conditioning frames. We also evaluate the Kuramoto--Sivashinsky (KS) equation with irregular timesteps. Details of the datasets and preprocessing are given in Appendices~\ref{app:inc_flow_data} to ~\ref{app:ks_data}.

\paragraph{Long-Horizon Forecasting Stability and Cost \label{subsection:results}}

Table~\ref{tab:quantitative_split_comparison_small} reports the main results. On incompressible flow, KAE outperforms ACDM on both extrapolation regimes, $Inc_{low}$ and $Inc_{high}$. For transonic flow, ACDM and other baselines generally achieve lower short-term errors on $Tra_{int}$ and $Tra_{ext}$. At the 240-step horizon, however, KAE gives the best stability among the compared methods. It is also over $110\times$ faster than ACDM at inference and uses fewer training epochs. Refiner and FNO-32 diverge at long horizons.

\begin{table*}[t]
\centering
\resizebox{0.8\textwidth}{!}{%
\begin{tabular}{@{}l cc cc cc cc cc cc@{}}
\toprule

\textbf{Method}
& {$Inc_{low}$}
& {$Inc_{high}$}
& {$Tra_{ext}$}
& {$Tra_{int}$}
& {$Tra_{long}$}
& {\textbf{Cost}} \\

\midrule

ResNet-dil \citep{he2016deep}
& $1.6\pm1.8$
& $1.5\pm0.8$
& $1.7\pm1.0$
& $1.7\pm1.4$
& $22.0\pm2.4$
& $3.46$ \\

FNO$_{16}$ \citep{li2021fourier}
& $2.8\pm3.1$
& $8.9\pm3.8$
& $4.8\pm1.2$
& $5.5\pm2.6$
& $20.8\pm2.0$
& $1.17$ \\

TF$_{Enc}$ \citep{vaswani2017attention}
& $1.5\pm1.7$
& $\mathbf{0.6\pm0.3}$
& $3.3\pm1.2$
& $6.2\pm4.2$
& $22.2\pm3.9$
& $0.60$ \\

Refiner \citep{lippe2023pde}
& $\mathbf{1.3\pm1.4}$
& $3.5\pm2.2$
& $5.4\pm2.1$
& $7.1\pm2.1$
& \text{Div.}
& $10.31$ \\

ACDM \citep{kohl2026benchmarking}
& $1.7\pm2.2$
& $\mathbf{0.8\pm0.4}$
& $2.3\pm1.4$
& $\mathbf{2.7\pm2.1}$
& $22.6\pm4.0$
& $126.57$ \\

KAE (Ours)
& $1.3\pm1.7$
& $2.9\pm1.1$
& $\mathbf{2.2\pm0.9}$
& $5.2\pm2.4$
& $\mathbf{14.9\pm1.3}$
& $\mathbf{1.15}$ \\

\bottomrule
\end{tabular}%
}

\caption{Forecasting accuracy and inference cost across incompressible and
transonic flow regimes. Lower MSE and LSiM indicate better performance.
MSE values are scaled by $10^{-4}$, $10^{-5}$, and $10^{-3}$ for
$Inc_{low}$, $Inc_{high}$, and the transonic regimes, respectively.
Full results and ablations are given in
Table~\ref{tab:quantitative_split_comparison} in Appendix~\ref{appendix:full_comparison}.}

\label{tab:quantitative_split_comparison_small}
\end{table*}



\textbf{Long Testing (1000 Steps):}
In a 1000-step rollout, ACDM develops large phase and spatial errors, while KAE remains bounded (Figure~\ref{fig:1000_metrics}). KAE gradually loses fine-scale detail but preserves the main flow structure, as also seen in the shorter 240-step rollout (Figure~\ref{fig:very_long_rollout}). Further error, spectral, and extrapolation analyses are provided in Appendices~\ref{appendix:error_analysis} and \ref{app:spectra_bias}–\ref{appendix:long_rollout}.

\begin{figure*}[t]
    \centering
    \begin{subfigure}[b]{0.48\textwidth}
        \centering
        \includegraphics[width=\textwidth]{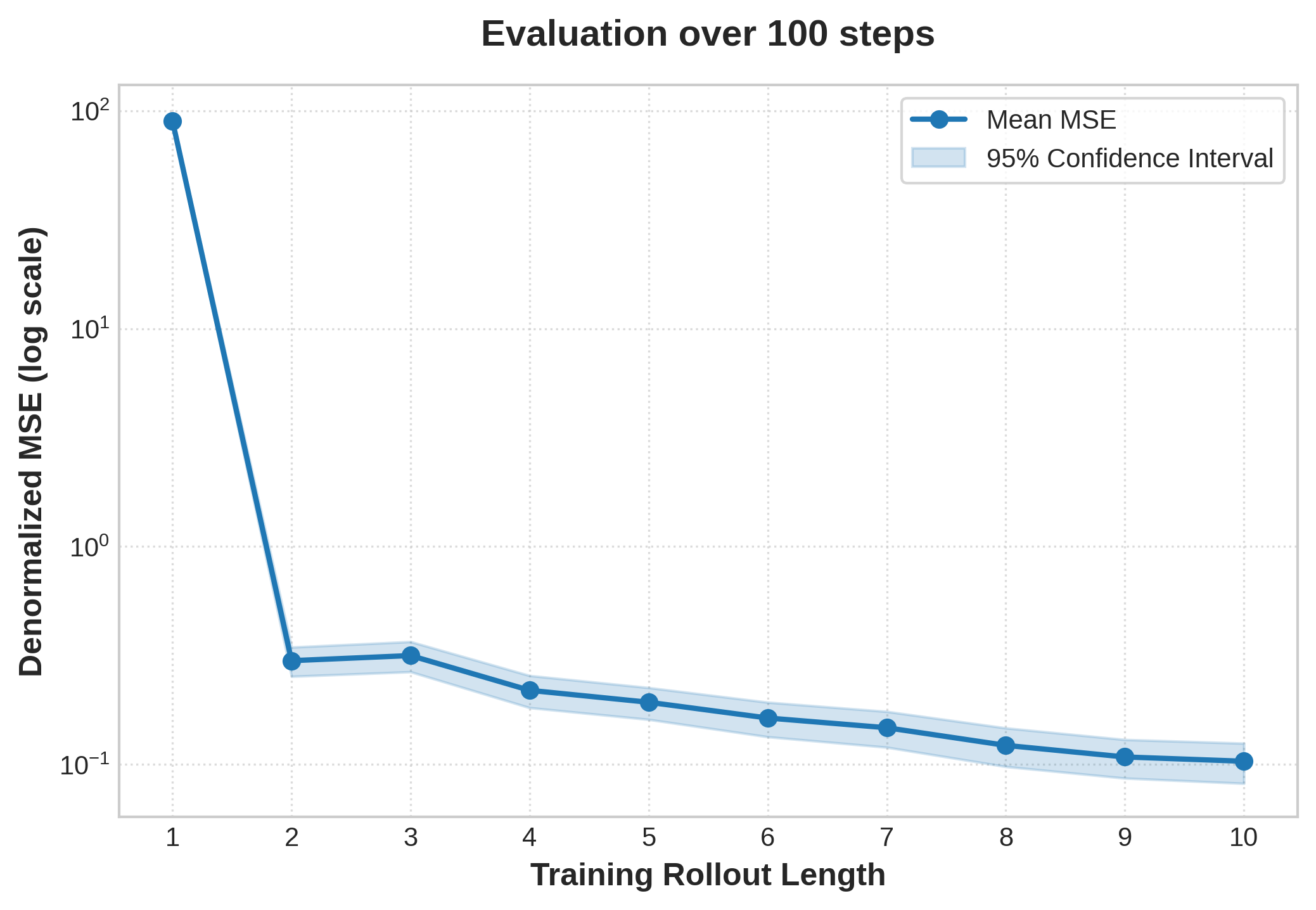}
        \caption{Failure at Rollout $R=1$}
        \label{fig:ks_failure}
    \end{subfigure}
    \hfill
    \begin{subfigure}[b]{0.48\textwidth}
        \centering
        \includegraphics[width=\textwidth]{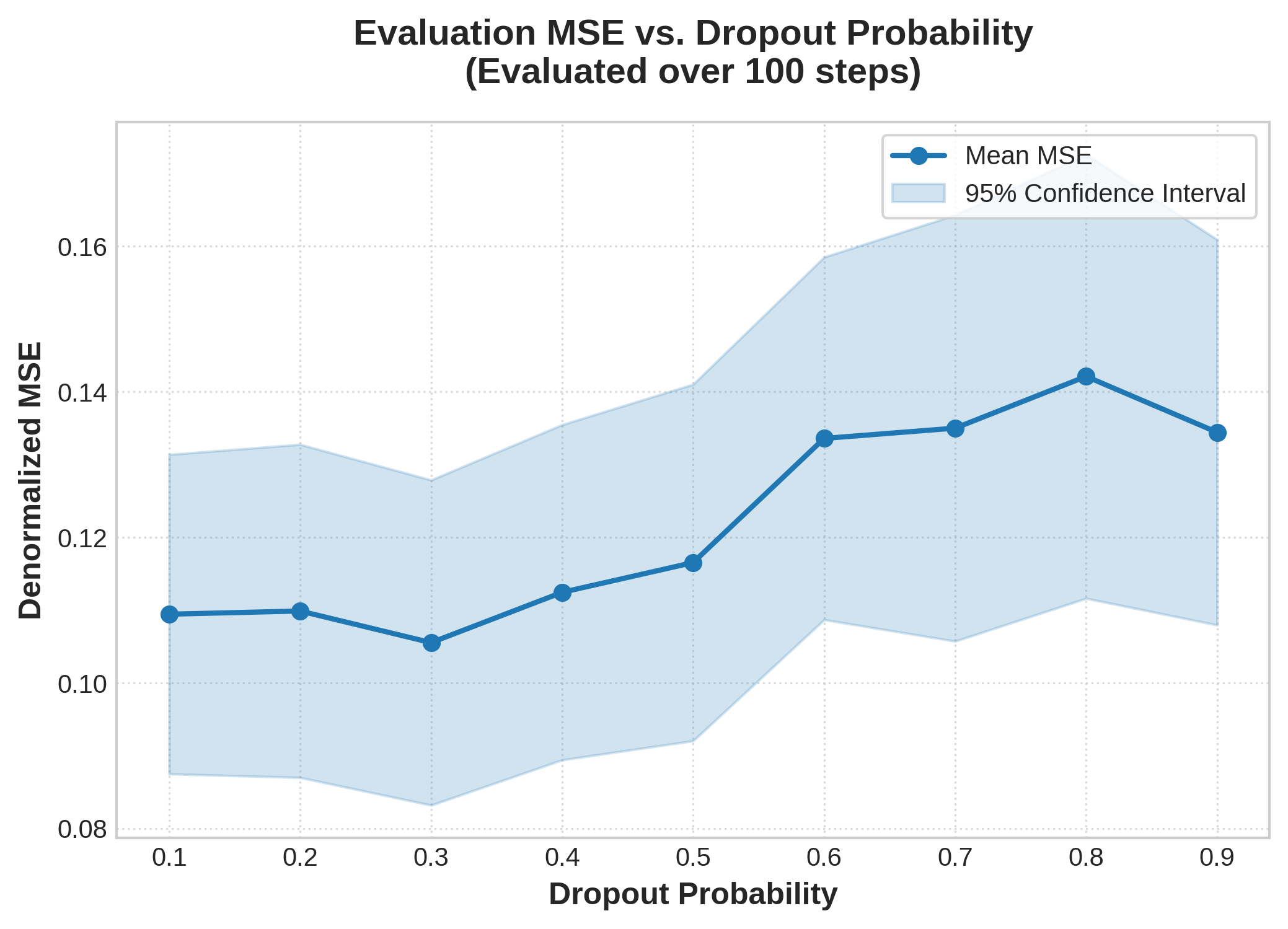}
        \caption{Irregular-Timestep Robustness}
        \label{fig:ks_dropout}
    \end{subfigure}
   \caption{KS stability analysis. (a) Single-step training ($R=1$) causes catastrophic divergence during 100-step evaluation. (b) The continuous generator maintains stable performance even when 80\% of training frames are randomly discarded.}
    \label{fig:ks_results_combined}
\end{figure*}

\textbf{Effect of rollout training:}
For the chaotic KS equation, training with a single rollout step ($R=1$) leads to divergence during 100-step evaluation (Figure~\ref{fig:ks_results_combined}a). Increasing the training rollout to $R=2$ substantially reduces the error, with further improvements up to $R=10$. Thus, multi-step training is important for long-term stability.

\paragraph{Recovery of the Continuous-Time Koopman Operator}

\begin{figure*}[t]
    \centering
    \begin{subfigure}[b]{0.33\textwidth}
        \centering
        \includegraphics[width=\textwidth]{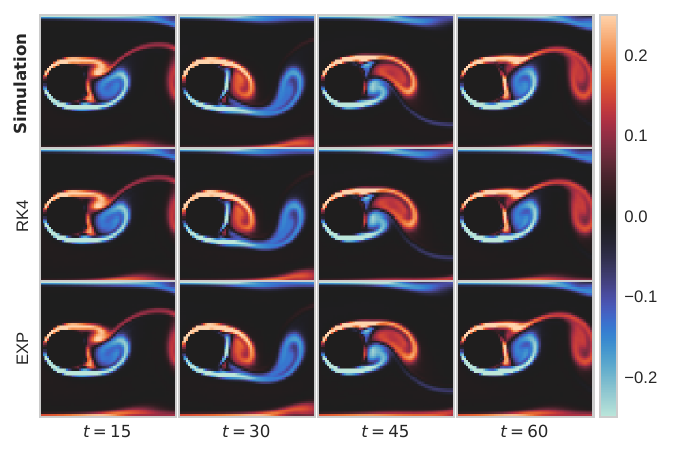}
        \caption{Vorticity (RK4 vs Exp)}
        \label{fig:inc_high_exp}
    \end{subfigure}
    \hfill
    \begin{subfigure}[b]{0.33\textwidth}
        \centering
        \includegraphics[width=\textwidth]{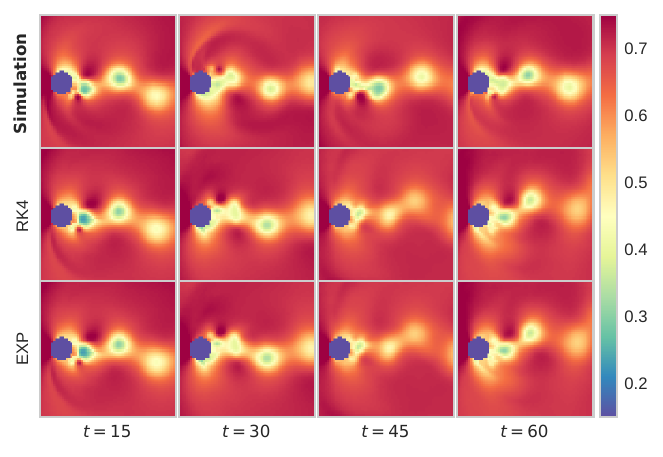}
        \caption{Pressure (RK4 vs Exp)}
        \label{fig:tra_extrap_exp}
    \end{subfigure}
    \hfill
    \begin{subfigure}[b]{0.295\textwidth}
        \centering
        \includegraphics[width=\textwidth]{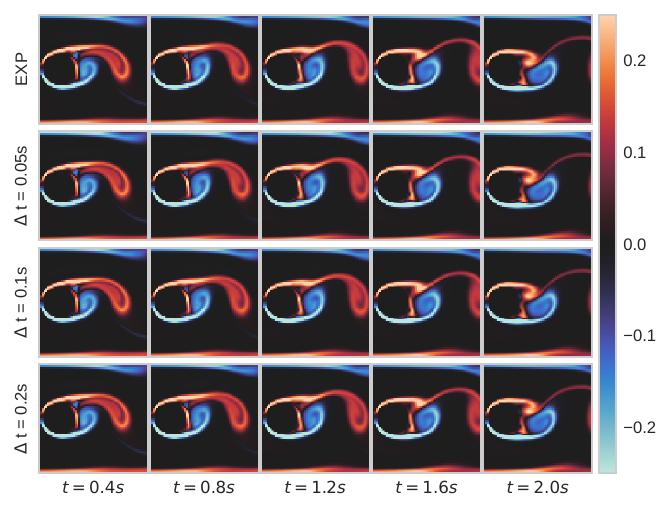}
        \caption{Variable $\Delta t$ Inference}
        \label{fig:delta_t_comparison}
    \end{subfigure}
    \vspace{-0.5em}
    \caption{Validation of continuous-time latent dynamics. (a,b) Close agreement between RK4 integration and the matrix exponential for incompressible and transonic regimes. (c) Rollouts at three unseen integration step sizes ($\Delta t \in \{0.05, 0.1, 0.2\}$ s), demonstrating zero-shot temporal generalization.}
    \label{fig:integrator_and_temporal}
\end{figure*}

We test whether the learned $\mathbf{K}_{\mathrm{cont}}$ represents a continuous-time generator. RK4 integration and the closed-form matrix exponential produce closely matching trajectories across the incompressible and transonic datasets (Figures~\ref{fig:inc_high_exp}--\ref{fig:tra_extrap_exp}). The eigenvalue spectrum is also mostly in the stable half-plane, $\mathrm{Re}(\lambda)<0$ (Appendix~\ref{appendix:eigenvalues}). The continuous-time formulation and its relationship to discrete Koopman consistency are further discussed in Appendix~\ref{appendix:consistency_equivalence}, while numerical solver behavior and physical stiffness are analyzed in Appendix~\ref{appendix:ode_solvers}. These results support the use of $\mathbf{K}_{\mathrm{cont}}$ as a continuous-time operator rather than a discrete approximation. The model also supports inference at unseen timesteps. Figure~\ref{fig:delta_t_comparison} shows rollouts using $\Delta t \in \{0.05,0.1,0.2\}$ without retraining.

\paragraph{Generalization to Unseen Dynamics via Physics Conditioning}
We test whether the physics-conditioned LoRA operator allows the model to generalize to unseen flow regimes. Figure~\ref{fig:inc_fieldwise} shows MSE across Reynolds numbers from $100$ to $1000$. KAE gives smoother error changes across the range, while ACDM shows larger variation, especially at high Reynolds numbers and for pressure. On $Tra_{ext}$, KAE achieves a low, comparable MSE. Removing physics conditioning increases the error by two orders of magnitude. This shows that the physics-conditioned LoRA is important for generalization across flow regimes. Full ablations are provided in Appendix~\ref{appendix:ablations}.

\paragraph{Training on Irregularly Sampled Data \label{subsec:q4}}

Finally, we test whether KAE can train on irregularly sampled data. We randomly remove KS frames with dropout probabilities $p\in[0.1,0.9]$. The model remains stable across all dropout levels; even with 90\% dropout, the 100-step MSE increases only slightly (Figure~\ref{fig:ks_results_combined}b). This shows that the continuous formulation does not require a fixed timestep during training. Further results and ablations are given in Appendix~\ref{appendix:irregular_timestep}.

\begin{figure*}[t]
    \centering
    \begin{subfigure}[b]{0.54\textwidth}
        \centering
        \includegraphics[width=\textwidth]{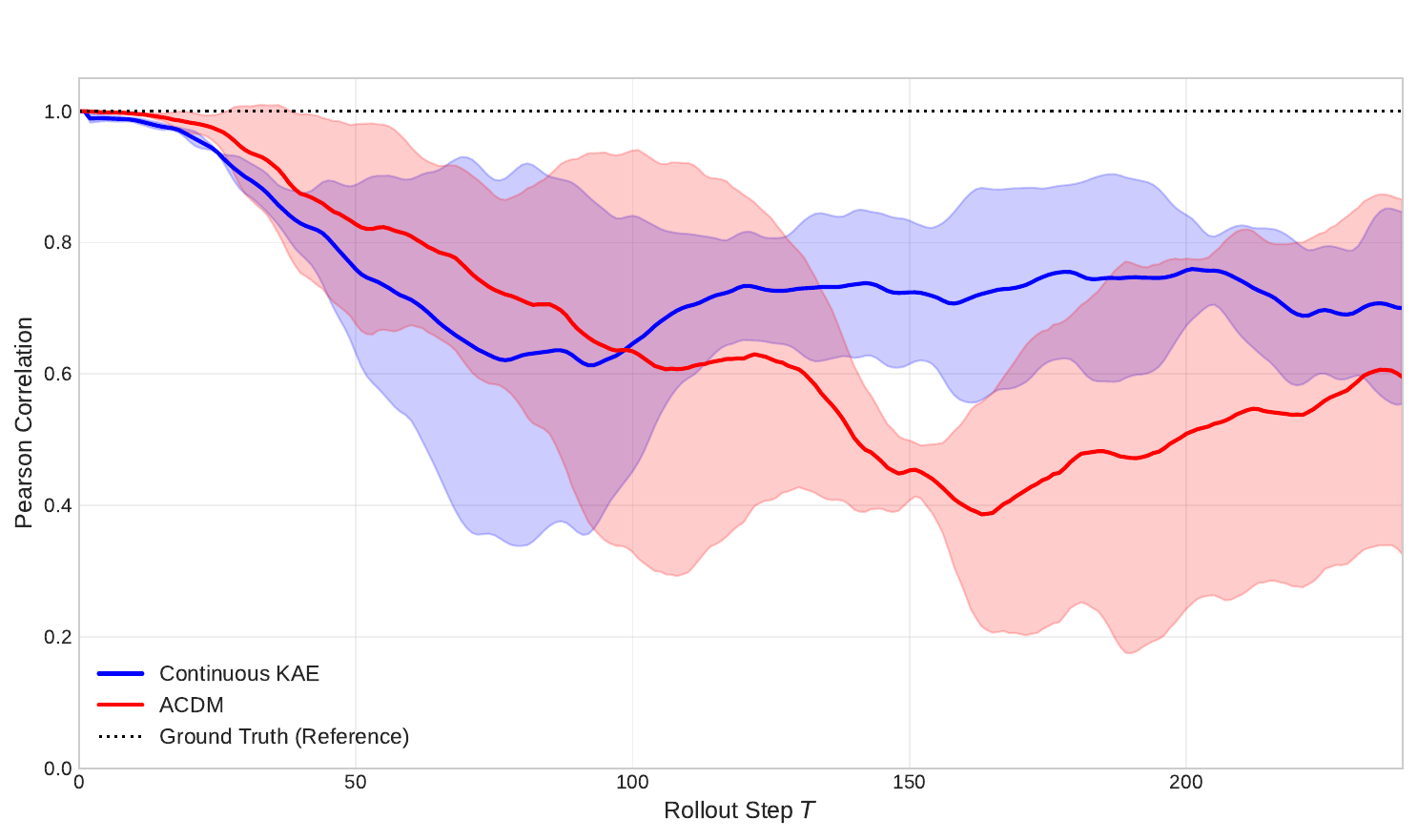}
        \caption{Spatial Correlation (240 Steps)}
        \label{fig:spatial_corr_250}
    \end{subfigure}
    \hfill
    \begin{subfigure}[b]{0.44\textwidth}
        \centering
        \includegraphics[width=\textwidth]{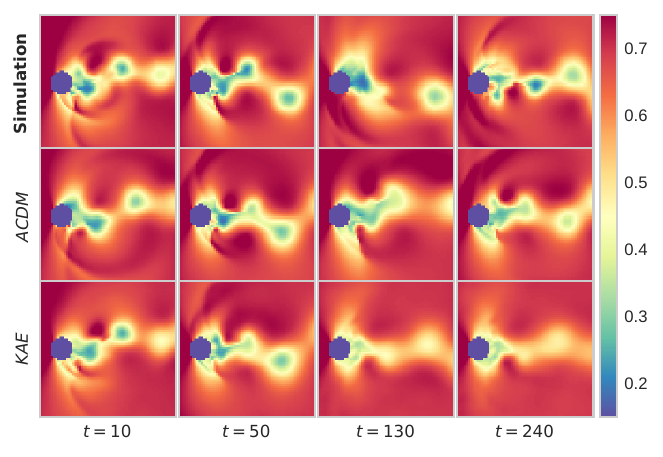}
        \caption{Long Rollout (240 Steps)}
        \label{fig:stability_visuals}
    \end{subfigure}
    \vspace{-0.5em}
    \caption{Evaluation of long-horizon structural stability in the chaotic Transonic regime. (a) Spatial Pearson correlation over a 240-step rollout, demonstrating that KAE maintains stable structural alignment while the diffusion baseline diverges. (b) Visual snapshots over a 240-step rollout. While ACDM compounds stochastic errors into numerical noise, KAE diffuses smoothly into a stable limit cycle.}
    \label{fig:very_long_rollout}
\end{figure*}

\paragraph{Does forecasting accuracy predict assimilation accuracy?}
\label{exp:da}

\begin{wrapfigure}{r}{0.40\textwidth}
    \centering
    \includegraphics[width=0.38\textwidth]{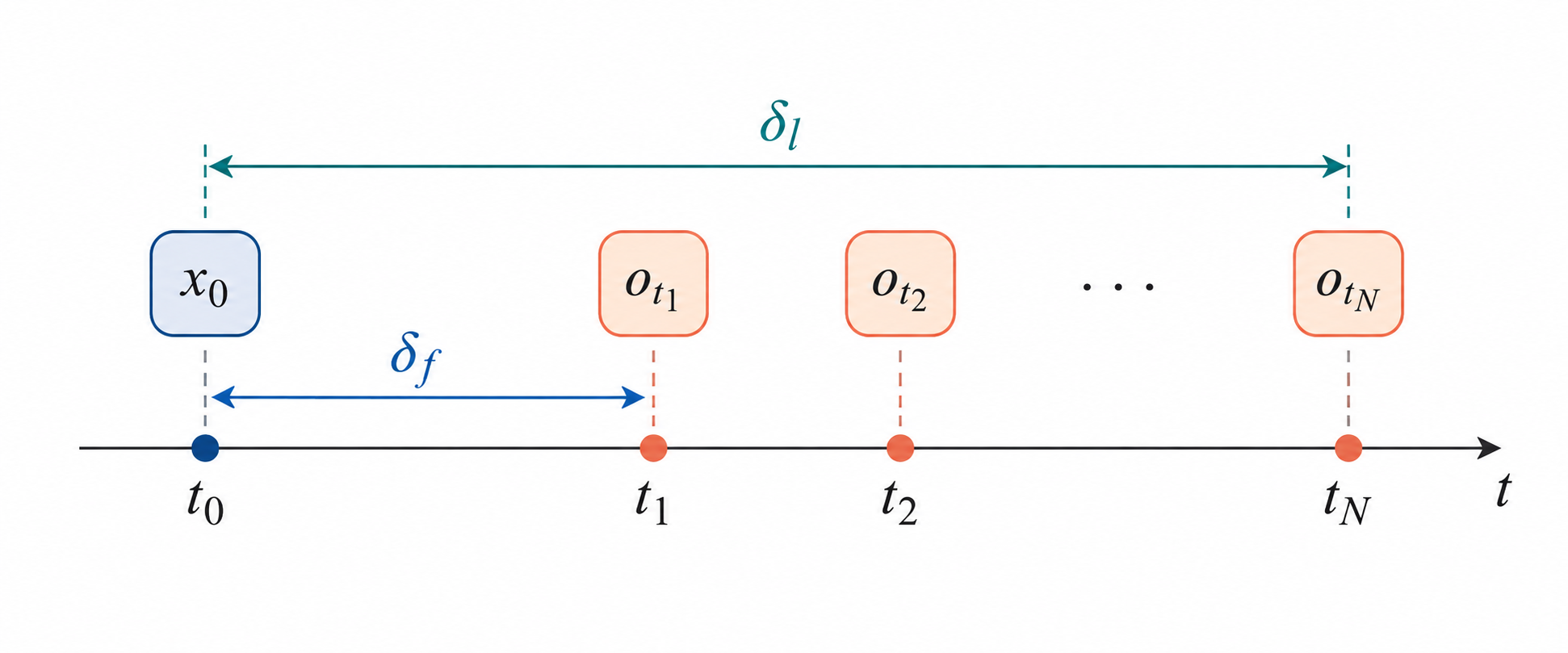}
    \caption{Qualitative representation of the parameters considered
    in the assimilation process.}
    \label{fig:deltafl}
\end{wrapfigure}

We next evaluate the same models as state estimators using variational data assimilation. The goal is to recover the unknown state $u(t_0)$ from observations collected after $t_0$. We vary the first-observation lead time $\delta_f$, while keeping the recovery horizon $\delta_l=25$ frames and the number of observations $N=5$ fixed unless stated otherwise. The $N$ observations span $[\delta_f,\delta_l]$, with the endpoints fixed and interior observations geometrically spaced.

For KAE, the observations are encoded and mapped back to $t_0$ using the learned continuous operator, $z_0=e^{-\mathbf{K}_{\mathrm{cont}}\tau_1}\mathcal{E}(y_1)$. For the autoregressive baselines, the nearest observation is used as the initial state. This warm start gives all methods an initial estimate before optimisation.

The transonic experiment provides a direct comparison between the two tasks. When given the true initial state, U-Net gives the lowest 20-step forecasting error, while KAE gives the highest. When the initial state must instead be recovered from observations, the ordering reverses: KAE gives the lowest error in both examples, while U-Net is among the worst. Details of algorithms implementations are in Appendix~\ref{app:da_algs}.

\begin{figure}[t]
    \centering
    \includegraphics[width=0.8\linewidth]{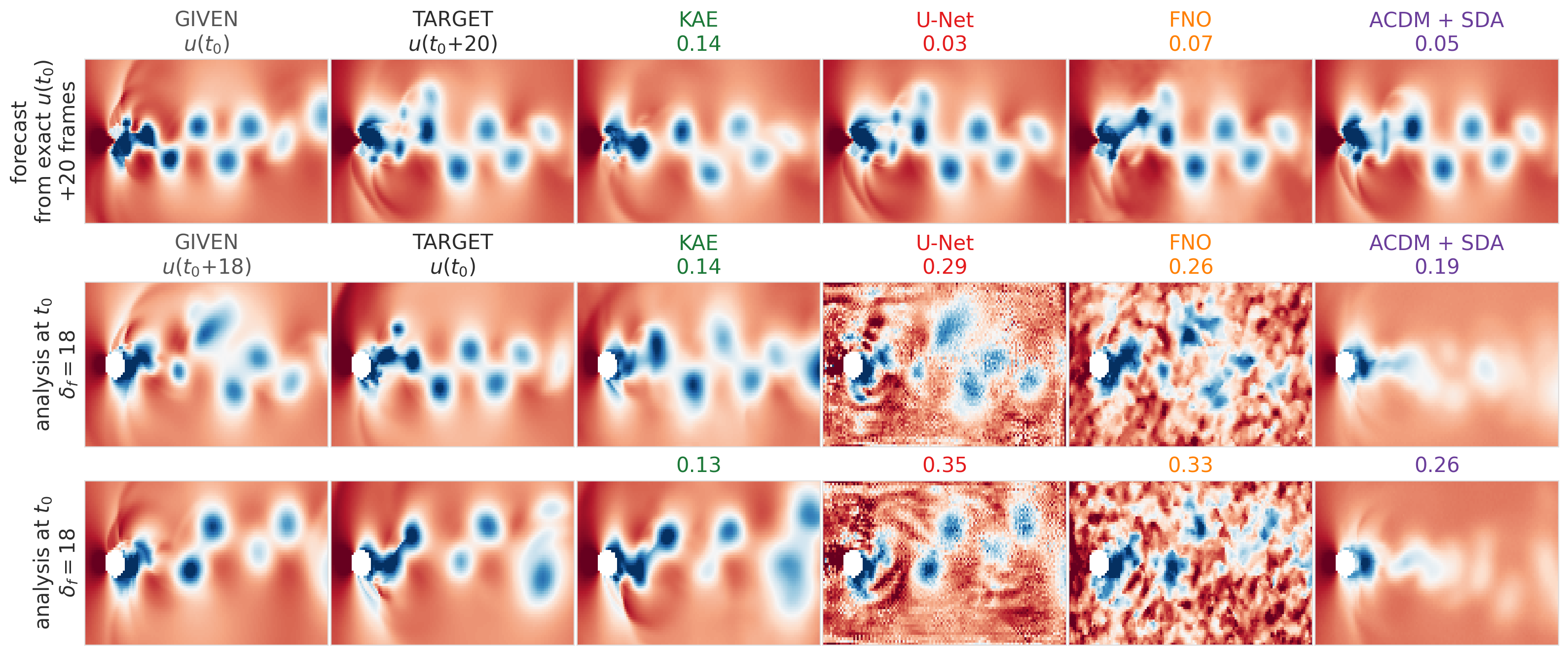}
    \caption{\textbf{Transonic fields: the same four models as forecasters and as assimilators.} Pressure is shown and the number above each panel is the relative $L_2$ over all four channels.
    \textbf{Top block --- forecasting.} Each model is handed the exact $u(t_0)$ and rolled forward $20$ frames; no assimilation and no optimisation.
    \textbf{Lower blocks --- assimilation.} $u(t_0)$ is never given. It is recovered from five observations, so the earliest sits $\df=18$ frames after the state being recovered.}
    \label{fig:tra_fields_qual}
\end{figure}

\paragraph{How does assimilation change as observations become less informative?}

The previous comparison uses a fixed assimilation configuration. We next vary the first-observation lead time $\delta_f$ to measure how the models behave as the state being recovered becomes increasingly separated from the available observations.

\paragraph{Kuramoto--Sivashinsky}\label{sec:da_ks}

\begin{figure*}[t]
    \centering

    \begin{subfigure}[b]{0.515\textwidth}
        \centering
        \includegraphics[width=\textwidth]{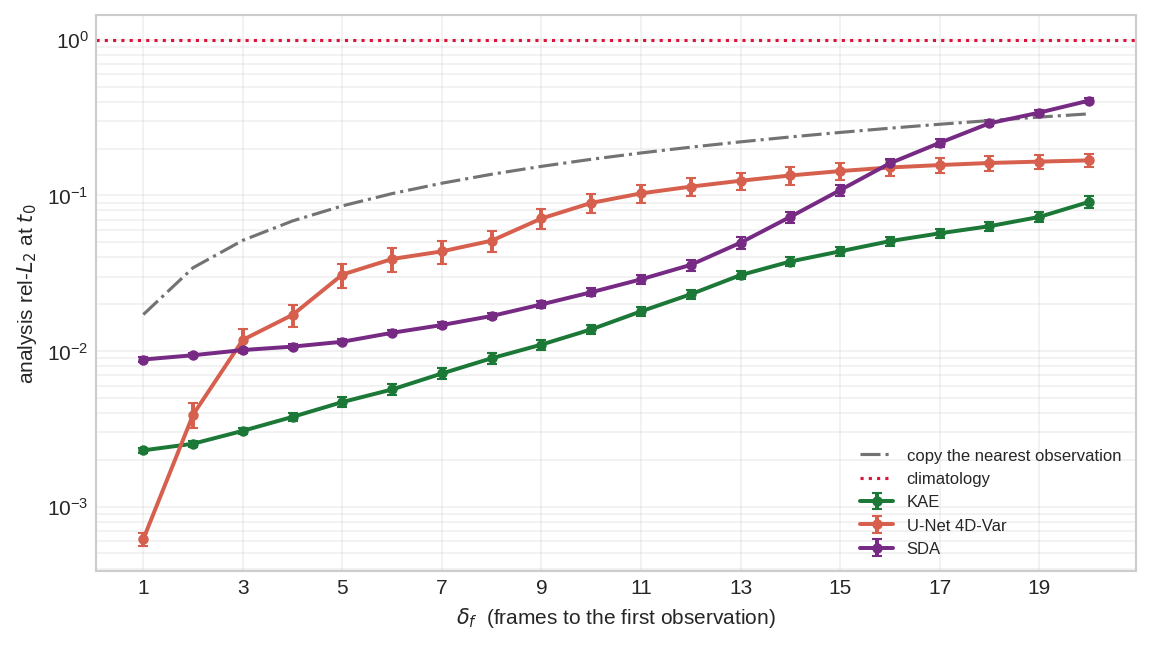}
        \caption{Kuramoto--Sivashinsky}
        \label{fig:ks_deltaf_fine}
    \end{subfigure}
    \hfill
    \begin{subfigure}[b]{0.465\textwidth}
        \centering
        \includegraphics[width=\textwidth]{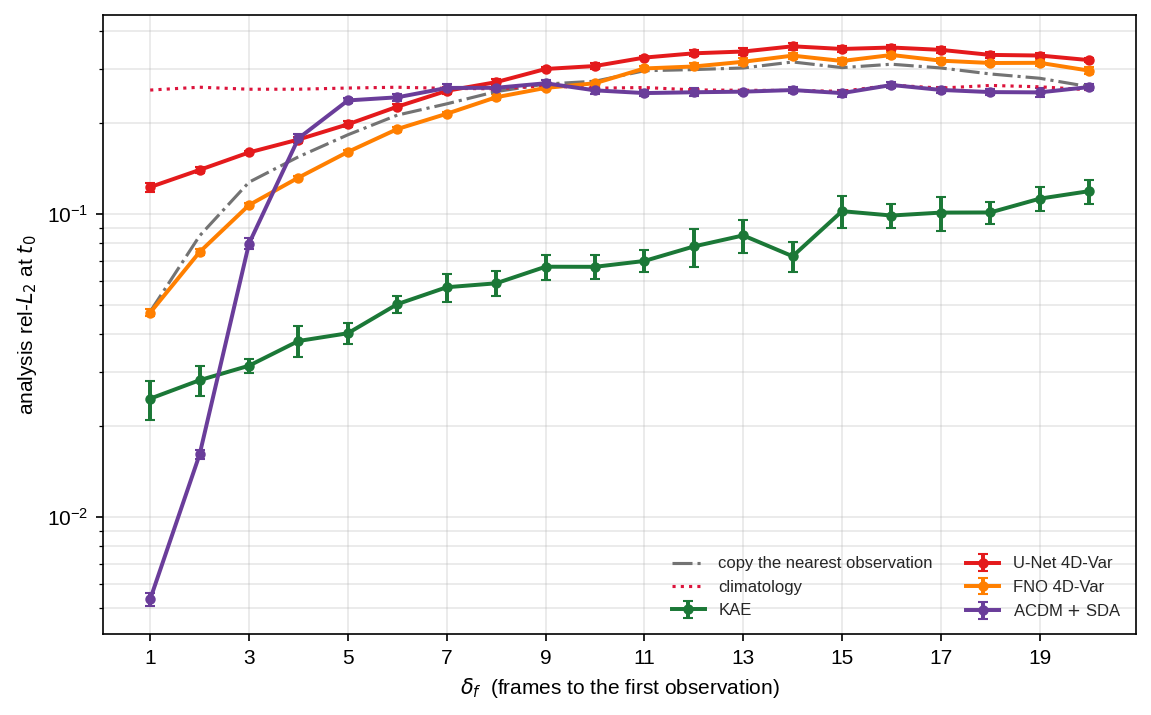}
        \caption{Transonic cylinder flow}
        \label{fig:tra_deltaf_fine}
    \end{subfigure}

    \caption{\textbf{Assimilation error versus first-observation lead time.}
    The first observation is moved from $\delta_f=1$ to $20$ frames while
    $\delta_l=25$ and $N=5$ remain fixed. 
    (a) On KS, KAE maintains lower assimilation error than SDA across the
    lead-time range, while SDA eventually becomes worse than copying the
    nearest observation.
    (b) On transonic flow, ACDM$+$SDA performs well at small $\delta_f$
    but degrades rapidly as the observation gap increases. KAE remains
    below the reference baselines across the full range.
    The dashed line shows the nearest-observation baseline and the dotted
    line shows climatology.}
    \label{fig:assimilation_leadtime}
\end{figure*}

\begin{figure}[t]
    \centering
    \includegraphics[width=0.9\linewidth]{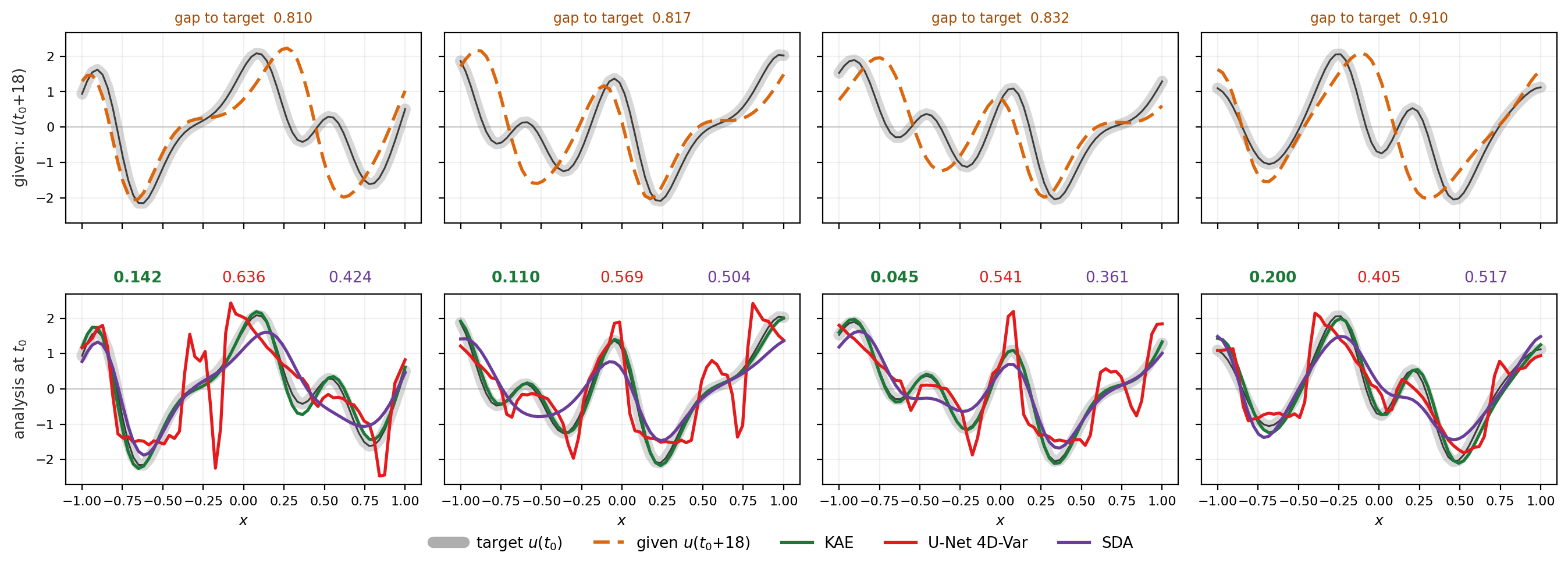}
     \caption{\textbf{KS states recovered at $t_0$, four trajectories.} 
     \emph{Top row:} the same target with the field each method was actually \emph{given} ($\df=18$), which is also where every solve starts.
     \emph{Bottom row:} the target $u(t_0)$, with all three analyses plotted on it; the three numbers above each panel are that panel's relative $L_2$, coloured to match the curve they belong to. We see a similar pattern in \citep{lippe2023pde}.}
    \label{fig:ks_fields_qual}
\end{figure}

We first test how assimilation changes as the first observation is moved further from the unknown initial state. We sweep $\delta_f$ from $1$ to $20$ frames while keeping $\delta_l=25$ and $N=5$ fixed (Figure~\ref{fig:ks_deltaf_fine}). Across the sweep, KAE improves over SDA by a more than twice. SDA eventually becomes worse than simply copying the nearest observation at large $\delta_f$. Figure~\ref{fig:ks_fields_qual} shows four recovered KS states. The models receive only observations after $t_0$; the true state is never provided.

\textbf{Experimental setup.}
The KS equation is $u_t=-uu_x-u_{xx}-u_{xxxx}$ on a periodic domain with 64 spatial points and $\Delta t=0.1$. Training uses 100 trajectories of 1000 frames. Assimilation uses 64 separate trajectories, with an additional 32 trajectories of 5000 frames for post-assimilation forecasts. The Lyapunov time is $T_L=22.6$ time units. KAE and U-Net are trained on the same data using 10-frame rollouts. The score model is trained on nine-frame segments. For assimilation, KAE and U-Net use 8000 Adam iterations, while SDA uses its fixed sampling procedure. Detailed KS assimilation settings are given in Appendix~\ref{app:da_ks}, including the convergence analysis in Figure~\ref{fig:ks_convergence_df9}; the corresponding KS ablations and parameter sweeps are reported in Appendix~\ref{app:da_ablations} and Figure~\ref{fig:ks_sweeps}.

\paragraph{Transonic Cylinder Flow}\label{sec:da_tra}

We repeat the lead-time experiment on transonic flow (Figure~\ref{fig:tra_deltaf_fine}). Here, $\delta_f$ again varies from $1$ to $20$ frames, while $\delta_l=25$ and $N=5$ remain fixed. The evaluation uses unseen intermediate Mach numbers $Ma\in\{0.66,0.67,0.68\}$.

\textbf{Experimental setup.}
The transonic dataset contains compressible flow past a cylinder at $Re=10^4$, with resolution $128\times64$ and four physical fields. The data are sampled at $\Delta t=0.1$ s. Training covers $Ma\in[0.53,0.63]\cup[0.69,0.89]$. Assimilation uses the unseen intermediate range $Ma\in\{0.66,0.67,0.68\}$.Full transonic model and optimization details are provided in Appendix~\ref{app:da:tra}, while the corresponding TRA sweeps and assimilation ablations are summarized in Figure~\ref{fig:tra_sweeps} and Appendix \ref{app:da_ablations}.

ACDM's error increases rapidly with lead time and is rapidly worse than KAE, and quickly reaching the climatology error. U-Net and FNO are also unable to improve consistently over simply copying the nearest observation. KAE is the only method that remains below both reference baselines across the full lead-time range.

\section{Discussion and Limitations}
\label{sec:discussion}
The experiments show that forecasting and data assimilation test different properties of learned dynamical surrogates. KAE provides stable long-horizon forecasts, but forecasting accuracy does not consistently predict assimilation accuracy. In the transonic experiments, U-Net achieves lower short-horizon forecasting error, while KAE achieves lower assimilation error, with the same separation observed in the controlled ablation. Thus, forecast accuracy alone does not characterize a surrogate’s usefulness for state estimation.

The continuous-time formulation propagates latent states directly to arbitrary observation times without intermediate autoregressive steps, making it suitable for irregularly spaced observations and long assimilation windows. Experiments with irregular sampling and unseen integration timesteps support this property.

There are several limitations. The assimilation experiments use a controlled variational state-recovery setting and only two fluid systems and the Kuramoto–Sivashinsky equation. Assimilation still requires iterative optimisation, so the computational benefit depends on the observation configuration and optimisation procedure. Finally, although KAE remains stable over long horizons, it loses fine-scale detail at 1000 steps. Further evaluation on more complex systems and observation settings is needed.




\section{Conclusion}
\label{sec:conclusion}

We introduced a continuous-time Koopman autoencoder for long-term forecasting and data assimilation in complex physical systems. Its continuous latent dynamics enable direct propagation to arbitrary timesteps and observation times, without intermediate autoregressive steps.

Across incompressible flow, transonic flow, and the Kuramoto–Sivashinsky equation, KAE provides stable long-horizon forecasts at low inference cost. More importantly, the experiments show that forecasting accuracy does not determine assimilation accuracy: models that perform well as forward predictors can perform poorly at initial-state recovery, while KAE can show the opposite behaviour. The controlled loss ablation confirms this separation within the same architecture.

These results highlight that forecasting and data assimilation should be evaluated as distinct properties of learned physical surrogates. For state recovery from observations, forward forecasting accuracy alone is not sufficient to characterize the usefulness of the learned dynamics.
\subsection*{AI Use Disclosure}

We used generative AI tools during the preparation of this work to assist with drafting sections of the manuscript, organizing and editing the presentation of the research, and developing and automating parts of the experimental workflow, including experiment execution and plot generation. All AI-assisted text was reviewed and modified by the authors before inclusion in the manuscript.

Generative AI tools were also used to reproduce existing implementation code and generate the Kuramoto--Sivashinsky datasets used in this work. The underlying dataset-generation procedure was based on the existing code and was reviewed and verified by the authors. AI tools were not used to independently conduct the research: experimental procedures, generated data, experimental outputs, plots, numerical results, and reported claims were checked by the authors.

All AI-assisted code and outputs were reviewed by the authors and verified against the intended experimental procedures. The authors take full responsibility for the final content of this work, including all text, code, datasets, experimental results, analyses, figures, and claims.

\bibliographystyle{plainnat} 
\bibliography{bibliography} 

@article{buzhardt2025relationship,
  title={On the relationship between Koopman operator approximations and neural ordinary differential equations for data-driven time-evolution predictions},
  author={Buzhardt, Jake and Constante-Amores, C Ricardo and Graham, Michael D},
  journal={Chaos: An Interdisciplinary Journal of Nonlinear Science},
  volume={35},
  number={4},
  year={2025},
  publisher={AIP Publishing}
}

@article{koopman1931hamiltonian,
  title={Hamiltonian systems and transformation in Hilbert space},
  author={Koopman, Bernard O},
  journal={Proceedings of the National Academy of Sciences},
  volume={17},
  number={5},
  pages={315--318},
  year={1931}
}

@article{penny2022integrating,
  title={Integrating recurrent neural networks with data assimilation for scalable data-driven state estimation},
  author={Penny, Stephen G and Smith, Timothy A and Chen, T-C and Platt, Jason A and Lin, H-Y and Goodliff, Michael and Abarbanel, Henry DI},
  journal={Journal of Advances in Modeling Earth Systems},
  volume={14},
  number={3},
  year={2022}
}

@article{adrian2025data,
  title={Data assimilation with machine learning surrogate models: A case study with FourCastNet},
  author={Adrian, Melissa and Sanz-Alonso, Daniel and Willett, Rebecca},
  journal={Artificial Intelligence for the Earth Systems},
  volume={4},
  number={3},
  pages={e240050},
  year={2025}
}

@article{frion2024neural,
  title={Neural Koopman prior for data assimilation},
  author={Frion, Anthony and Drumetz, Lucas and Dalla Mura, Mauro and Tochon, Guillaume and El Bey, Abdeldjalil Aissa},
  journal={IEEE Transactions on Signal Processing},
  volume={72},
  pages={4191--4206},
  year={2024}
}

@article{hatfield2021building,
  title={Building tangent-linear and adjoint models for data assimilation with neural networks},
  author={Hatfield, Sam and Chantry, Matthew and Dueben, Peter and Lopez, Philippe and Geer, Alan and Palmer, Tim},
  journal={Journal of Advances in Modeling Earth Systems},
  volume={13},
  number={9},
  pages={e2021MS002521},
  year={2021}
}

@article{chattopadhyay2022towards,
  title={Towards physics-inspired data-driven weather forecasting: integrating data assimilation with a deep spatial-transformer-based U-NET in a case study with ERA5},
  author={Chattopadhyay, Ashesh and Mustafa, Mustafa and Hassanzadeh, Pedram and Bach, Eviatar and Kashinath, Karthik},
  journal={Geoscientific Model Development},
  volume={15},
  number={5},
  pages={2221--2237},
  year={2022}
}

@article{xiong2024koopman,
  title={Koopman neural operator as a mesh-free solver of non-linear partial differential equations},
  author={Xiong, Wei and Huang, Xiaomeng and Zhang, Ziyang and Deng, Ruixuan and Sun, Pei and Tian, Yang},
  journal={Journal of Computational Physics},
  volume={513},
  pages={113194},
  year={2024},
  publisher={Elsevier}
}

@article{otto2019linearly,
  title={Linearly recurrent autoencoder networks for learning dynamics},
  author={Otto, Samuel E and Rowley, Clarence W},
  journal={SIAM Journal on Applied Dynamical Systems},
  volume={18},
  number={1},
  pages={558--593},
  year={2019},
  publisher={SIAM}
}

@article{morton2018deep,
  title={Deep dynamical modeling and control of unsteady fluid flows},
  author={Morton, Jeremy and Jameson, Antony and Kochenderfer, Mykel J and Witherden, Freddie},
  journal={Advances in neural information processing systems},
  volume={31},
  year={2018}
}

@article{tang2020deep,
  title={A deep-learning-based surrogate model for data assimilation in dynamic subsurface flow problems},
  author={Tang, Meng and Liu, Yimin and Durlofsky, Louis J},
  journal={Journal of Computational Physics},
  volume={413},
  pages={109456},
  year={2020},
  publisher={Elsevier}
}

@article{kohl2026benchmarking,
  title={Benchmarking autoregressive conditional diffusion models for turbulent flow simulation},
  author={Kohl, Georg and Chen, Li-Wei and Thuerey, Nils},
  journal={Neural Networks},
  pages={108641},
  year={2026},
  publisher={Elsevier}
}

@inproceedings{vaswani2017attention,
  title     = {Attention Is All You Need},
  author    = {Vaswani, Ashish and Shazeer, Noam and Parmar, Niki and Uszkoreit, Jakob and Jones, Llion and Gomez, Aidan N and Kaiser, {\L}ukasz and Polosukhin, Illia},
  booktitle = {Advances in Neural Information Processing Systems (NeurIPS)},
  volume    = {30},
  year      = {2017}
}

@inproceedings{chen2018neural,
  title={Neural ordinary differential equations},
  author={Chen, Ricky TQ and Rubanova, Yulia and Bettencourt, Jesse and Duvenaud, David K},
  booktitle={Advances in Neural Information Processing Systems (NeurIPS)},
  year={2018}
}

@article{menier2025interpretable,
  title={Interpretable learning of effective dynamics for multiscale systems},
  author={Menier, Emmanuel and Kaltenbach, Sebastian and Yagoubi, Mouadh and Schoenauer, Marc and Koumoutsakos, Petros},
  journal={Proceedings of the Royal Society A: Mathematical, Physical and Engineering Sciences},
  volume={481},
  number={2305},
  year={2025},
  publisher={The Royal Society}
}

@article{lusch2018koopman,
  author = {Bethany Lusch and J. Nathan Kutz and Steven L. Brunton},
  title = {Deep learning for universal linear embeddings of nonlinear dynamics},
  journal = {Nature Communications},
  year = {2018},
  volume = {9},
  issue = {1}
}

@inproceedings{takeishi2017learning,
    title = {Learning Koopman Invariant Subspaces for Dynamic Mode Decomposition}, 
    author = {Takeishi, Naoya and Kawahara, Yoshinobu and Yairi, Takehisa}, 
    booktitle = {Advances in Neural Information Processing Systems 30}, 
    pages = {1130--1140}, 
    year = {2017}}

@inproceedings{azencot2020consistent,
    author = {Azencot, Omri and Erichson, N. Benjamin and Lin, Vanessa and Mahoney, Michael W.},
    title = {Forecasting sequential data using consistent Koopman autoencoders},
    year = {2020},
    booktitle = {Proceedings of the 37th International Conference on Machine Learning},
    articleno = {45},
    numpages = {11},
    series = {ICML'20}
}

@article{nayak2025tckae,
  author = {Indranil Nayak and Ananda Chakrabarti and Mrinal Kumar and  Fernando L. Teixeira and Debdipta Goswami},
  title = {Temporally-consistent koopman autoencoders for forecasting dynamical systems},
  journal = {Scientific Reports},
  year = {2025},
  volume = {15},
  issue = {1}
}

@article{halder2026reduced,
  title={Reduced-order modeling of large-scale turbulence using Koopman $\beta$-variational autoencoders},
  author={Halder, Rakesh and Eiximeno, Benet and Lehmkuhl, Oriol},
  journal={Physics of Fluids},
  volume={38},
  number={1},
  year={2026},
  publisher={AIP Publishing}
}

@article{ghase2017les,
  author = {Masoud Ghasemian and Z. Najafian Ashrafi and Ahmad Sedaghat},
  title = {A review on computational fluid dynamic simulation techniques for darrieus vertical axis wind turbines},
  journal = {Energy Conversion and Management},
  year = {2017},
  volume = {149},
  pages = {87-100}
}

@inproceedings{slotnick2014cfd,
  title={CFD Vision 2030 Study: A Path to Revolutionary Computational Aerosciences},
  author={Jeffrey Slotnick and Abdollah Khodadoust and Juan Alonso and David Darmofal and William Gropp and Elizabeth Lurie and Dimitri Mavriplis},
  year={2014},
  booktitle={High-Performance Computing (HPC) User Forum},
}

@inproceedings{li2021fourier,
  title={Fourier Neural Operator for Parametric Partial Differential Equations},
  author={Li, Zongyi and Kovachki, Nikola and Azizzadenesheli, Kamyar and Liu, Burigede and Bhattacharya, Kaushik and Stuart, Andrew and Anandkumar, Anima},
  booktitle={International Conference on Learning Representations (ICLR)},
  year={2021}
}

@article{paszke2019pytorch,
  title={Pytorch: An imperative style, high-performance deep learning library},
  author={Paszke, Adam and Gross, Sam and Massa, Francisco and Lerer, Adam and Bradbury, James and Chanan, Gregory and Killeen, Trevor and Lin, Zeming and Gimelshein, Natalia and Antiga, Luca and others},
  journal={Advances in neural information processing systems},
  volume={32},
  year={2019}
}

@article{loshchilov2017decoupled,
  title={Decoupled weight decay regularization},
  author={Loshchilov, Ilya and Hutter, Frank},
  journal={arXiv preprint arXiv:1711.05101},
  year={2017}
}

@inproceedings{he2016identity,
  title={Identity mappings in deep residual networks},
  author={He, Kaiming and Zhang, Xiangyu and Ren, Shaoqing and Sun, Jian},
  booktitle={European conference on computer vision},
  pages={630--645},
  year={2016},
  organization={Springer}
}

@inproceedings{wu2018group,
  title={Group normalization},
  author={Wu, Yuxin and He, Kaiming},
  booktitle={Proceedings of the European conference on computer vision (ECCV)},
  pages={3--19},
  year={2018}
}

@inproceedings{woo2018cbam,
  title={Cbam: Convolutional block attention module},
  author={Woo, Sanghyun and Park, Jongchan and Lee, Joon-Young and Kweon, In So},
  booktitle={Proceedings of the European conference on computer vision (ECCV)},
  pages={3--19},
  year={2018}
}

@article{hu2022lora,
  title={Lora: Low-rank adaptation of large language models.},
  author={Hu, Edward J and Shen, Yelong and Wallis, Phillip and Allen-Zhu, Zeyuan and Li, Yuanzhi and Wang, Shean and Wang, Lu and Chen, Weizhu and others},
  journal={ICLR},
  volume={1},
  number={2},
  pages={3},
  year={2022}
}

@inproceedings{ronneberger2015u,
  title={U-net: Convolutional networks for biomedical image segmentation},
  author={Ronneberger, Olaf and Fischer, Philipp and Brox, Thomas},
  booktitle={International Conference on Medical image computing and computer-assisted intervention},
  pages={234--241},
  year={2015},
  organization={Springer}
}

@article{brandstetter2022message,
  title={Message passing neural PDE solvers},
  author={Brandstetter, Johannes and Worrall, Daniel and Welling, Max},
  journal={arXiv preprint arXiv:2202.03376},
  year={2022}
}

@article{lorenz1969predictability,
  title={The predictability of a flow which possesses many scales of motion},
  author={Lorenz, Edward N},
  journal={Tellus},
  volume={21},
  number={3},
  pages={289--307},
  year={1969},
  publisher={Taylor \& Francis}
}

@inproceedings{he2016deep,
  title     = {Deep Residual Learning for Image Recognition},
  author    = {He, Kaiming and Zhang, Xiangyu and Ren, Shaoqing and Sun, Jian},
  booktitle = {Proceedings of the IEEE Conference on Computer Vision and
               Pattern Recognition (CVPR)},
  pages     = {770--778},
  year      = {2016},
  doi       = {10.1109/CVPR.2016.90}
}

@article{lippe2023pde,
  title={Pde-refiner: Achieving accurate long rollouts with neural pde solvers},
  author={Lippe, Phillip and Veeling, Bas and Perdikaris, Paris and Turner, Richard and Brandstetter, Johannes},
  journal={Advances in Neural Information Processing Systems},
  volume={36},
  pages={67398--67433},
  year={2023}
}

@article{rozet2023score,
  title={Score-based data assimilation},
  author={Rozet, Fran{\c{c}}ois and Louppe, Gilles},
  journal={Advances in Neural Information Processing Systems},
  volume={36},
  pages={40521--40541},
  year={2023}
}

@article{courtier1994strategy,
  title={A strategy for operational implementation of 4D-Var, using an incremental approach},
  author={Courtier, PHILIPPE and Th{\'e}paut, J-N and Hollingsworth, Anthony},
  journal={Quarterly Journal of the Royal Meteorological Society},
  volume={120},
  number={519},
  pages={1367--1387},
  year={1994},
  publisher={Wiley Online Library}
}

@article{fablet2021learning,
  title={Learning variational data assimilation models and solvers},
  author={Fablet, Ronan and Chapron, Bertrand and Drumetz, Lucas and M{\'e}min, Etienne and Pannekoucke, Olivier and Rousseau, Fran{\c{c}}ois},
  journal={Journal of Advances in Modeling Earth Systems},
  volume={13},
  number={10},
  pages={e2021MS002572},
  year={2021},
  publisher={Wiley Online Library}
}

@inproceedings{chung2022diffusion,
  title     = {Diffusion Posterior Sampling for General Noisy Inverse Problems},
  author    = {Chung, Hyungjin and Kim, Jeongsol and Mccann, Michael T. and Klasky, Marc L. and Ye, Jong Chul},
  booktitle = {International Conference on Learning Representations},
  year      = {2023}
}

@misc{bouttier1999four,
  title        = {Data Assimilation Concepts and Methods},
  author       = {Bouttier, Fran{\c{c}}ois and Courtier, Philippe},
  howpublished = {ECMWF Meteorological Training Course Lecture Series},
  year         = {1999}
}

@incollection{navon2009data,
  title     = {Data Assimilation for Numerical Weather Prediction: A Review},
  author    = {Navon, Ionel M.},
  booktitle = {Data Assimilation for Atmospheric, Oceanic and Hydrologic Applications},
  editor    = {Park, Seon K. and Xu, Liang},
  publisher = {Springer},
  pages     = {21--65},
  year      = {2009}
}

@article{langland2004estimation,
  title   = {Estimation of Observation Impact Using the {NRL} Atmospheric Variational Data Assimilation Adjoint System},
  author  = {Langland, Rolf H. and Baker, Nancy L.},
  journal = {Tellus A},
  volume  = {56},
  number  = {3},
  pages   = {189--201},
  year    = {2004}
}

@article{buizza1995singular,
  title   = {The Singular-Vector Structure of the Atmospheric Global Circulation},
  author  = {Buizza, Roberto and Palmer, Tim N.},
  journal = {Journal of the Atmospheric Sciences},
  volume  = {52},
  number  = {9},
  pages   = {1434--1456},
  year    = {1995}
}

@article{bengio1994learning,
  title   = {Learning Long-Term Dependencies with Gradient Descent is Difficult},
  author  = {Bengio, Yoshua and Simard, Patrice and Frasconi, Paolo},
  journal = {IEEE Transactions on Neural Networks},
  volume  = {5},
  number  = {2},
  pages   = {157--166},
  year    = {1994}
}

@inproceedings{pascanu2013difficulty,
  title     = {On the Difficulty of Training Recurrent Neural Networks},
  author    = {Pascanu, Razvan and Mikolov, Tomas and Bengio, Yoshua},
  booktitle = {International Conference on Machine Learning},
  pages     = {1310--1318},
  year      = {2013}
}

@inproceedings{mikhaeil2022difficulty,
  title     = {On the Difficulty of Learning Chaotic Dynamics with {RNNs}},
  author    = {Mikhaeil, Jonas M. and Monfared, Zahra and Durstewitz, Daniel},
  booktitle = {Advances in Neural Information Processing Systems},
  volume    = {35},
  year      = {2022}
}

@inproceedings{frion2023leveraging,
    title     = {Leveraging Neural {K}oopman Operators to Learn Continuous Representations of Dynamical Systems from Scarce Data},
    author    = {Frion, Anthony and Drumetz, Lucas and Dalla Mura, Mauro and Tochon, Guillaume and A{\"i}ssa-El-Bey, Abdeldjalil},
    booktitle = {IEEE International Conference on Acoustics, Speech and Signal Processing (ICASSP)},
    pages     = {1--5},
    year      = {2023}
  }

\newpage
\appendix

\section{Detailed explanation of Koopman Autoencoders} \label{appendix:koppman_derivation}

Consider a discrete-time dynamical system with states $x_t \in \mathbb{R}^{N_d}$, where $x_t$ denotes the state of the system at time $t \in \mathbb{R}$. The state evolves to its next timestep $t+\Delta t$ according to a flow map $f: \mathbb{R}^{N_d} \to \mathbb{R}^{N_d}$:
\begin{equation}
    \label{eq:discrete_dynamics}
    x_{t+\Delta t} = f(x_t).
\end{equation}
Koopman operator theory shifts the focus from the state variables $x_t$ to a set of measurement functions $h:\mathbb{R}^{N_d} \to \mathbb{R}$, which are elements of an infinite-dimensional Hilbert space. The Koopman operator $\mathcal{K}$ acts on measurement functions $h$ linearly with the dynamics:
\begin{equation}
    \label{eq:koopmantheory}
    (\mathcal{K} h)(x_t) = h(f(x_t)) = h(x_{t+\Delta t}).
\end{equation}
In this formulation, the underlying nonlinear system \eqref{eq:discrete_dynamics} is transformed into a linear evolution in the infinite-dimensional space of observables. The challenge lies in choosing or constructing a suitable set of observables. KAEs address this by learning the observables from data, training an encoder which maps the state into a latent representation $\mathcal{E}: \mathbb{R}^{N_d} \to \mathbb{R}^{N_z}$,
\begin{equation}
    \label{eq:encoder}
    z_t = \mathcal{E}(x_t),
\end{equation}
which serves as a finite-dimensional approximation of the observable space. In this latent space, the evolution can be modelled as an approximately linear finite-dimensional Koopman operator $\mathbf{A}$,
\begin{equation}
    \label{eq:koopmanlatent}
    z_{t+\Delta t} \approx \mathbf{A} z_t.
\end{equation}
A decoder $\mathcal{D}$ then reconstructs the state from latent variables,
\begin{equation}
    \label{eq:decoder}
    \hat{x}_{t+\Delta t} = \mathcal{D}(z_{t+\Delta t}).
\end{equation}
In our continuous-time formulation, we replace the discrete operator with the linear ODE $dz/dt = \mathbf{K}_{\mathrm{cont}} z$, which allows inference at any horizon $\tau$ in a single matrix-vector product, regardless of $\Delta t$ used during training.

\subsection{Proposed Model}\label{app:proposed_model}

The proposed KAE consists of three main components: a dual-stream Transformer encoder, a parametric latent Koopman operator, and a CNN-based decoder (Appendix~\ref{app:architecure}-\ref{app:koop_dyn}), including the temporal history encoder in Appendix~\ref{app:history_enc}. Although the continuous formulation naturally supports irregularly sampled data (Paragraph~\ref{subsec:q4}), our experiments use uniformly spaced timesteps to enable direct comparison with existing baselines.

We employ a Transformer-based encoding architecture consisting of two symmetric streams: a history encoder ($\mathcal{E}_{\text{history}}$) and a present encoder ($\mathcal{E}_{\text{present}}$). To maintain consistency with benchmark evaluations, we formulate the temporal prediction as an Auto-Regressive process of order 2 (AR-2). Mathematically, we approximate a deterministic expected value mapping of the future state, rather than generating a full conditional probability distribution.

Fluid flows are inherently non-Markovian in observable space due to hidden state variables (e.g., pressure fields missing from velocity observations). Therefore, relying on a single spatial snapshot is dynamically insufficient. To construct a richer, fully Markovian initial state $z_{t_i}$ in the latent space, we aggregate temporal context. Specifically, the immediate past ($x_{t_{i-1}}$) and present ($x_{t_i}$) states are independently processed by the convolutional encoder. To break temporal symmetry and correctly capture flow directionality, these representations are then processed by a Transformer encoder \citep{vaswani2017attention} with sinusoidal positional encodings before being averaged across time:
\begin{equation}
    z_{t_i} = \text{MeanPooling}\Big(\text{Transformer}\big(\mathcal{E}_{\text{history}}(x_{t_{i-1}}), \mathcal{E}_{\text{present}}(x_{t_i})\big)\Big).
\end{equation}
A decoder $\mathcal{D}$ reconstructs the physical state from the latent vector, $\hat{x}_{t_i}=\mathcal{D}(z_{t_i}) $.

\subsubsection{Koopman Operator} \label{sec:koopman_operator}

Next, we discuss the structure of the continuous Koopman operator. While most KAE formulations operate in discrete time, directly mapping $z_t \to z_{t+\Delta t}$ as discussed in Section \ref{sec:background}, such updates are tied to the temporal resolution of training data. This dependence can hinder generalization across timesteps and reduce robustness in scenarios with different or irregular temporal resolutions. Our proposed continuous-time model interprets the latent evolution as an ODE, 
\begin{equation}
    \label{eq:method_ode}
    \frac{dz}{dt} = \mathbf{K}_{\text{cont}}(\phi) z,
\end{equation}
where $\mathbf{K}_{\text{cont}}$ governs the linear evolution in latent space. The state at a future time, $t_i+\Delta t$, is obtained by integrating this ODE:
\begin{equation}
    \label{eq:integration_ode}
    z_{t_i+\Delta t} = z_{t_{i+1}} = z_{t_i} + \int_{0}^{\Delta t}\frac{\mathrm{d} z}{\mathrm{d} \tau} d\tau,
\end{equation}
where any solver could used to conduct this integration stably during training, in practice we use fourth-order Runge–Kutta. During inference, exact matrix exponentiation could be used for maximum efficiency, which will be discussed in Section \ref{sec:inference_exp}.

Finally, we define the structure of the operator. To capture the dependency on physical parameters $\phi$, we express $\mathbf{K}_{\text{cont}}(\phi)$ as the sum of a static base dynamics matrix and a parametric adjustment:
\begin{equation}
\mathbf{K}_{\text{cont}}(\phi) = \mathbf{K}_{0} + \mathcal{N}_\psi(\phi).
\end{equation}
While conditioning neural surrogate models on physical parameters is common in PDE forecasting, incorporating this conditioning directly into a Koopman operator has received comparatively little attention. Therefore, an innovation in our approach is explicitly conditioning the continuous-time Koopman operator on these external control parameters, denoted by $\phi$. In our setting, $\phi$ represents physical parameters such as Reynolds or Mach numbers. Unlike traditional Koopman approaches where a single static operator is learned for the entire domain, our operator adapts dynamically based on the specific physical regime. Here, the base dynamics are governed by a learnable matrix $\mathbf{K}_0$, which captures the global, invariant behavior shared across all flow regimes. $\mathcal{N}_\psi$ is a Low-Rank Adaptation (LoRA \citep{hu2022lora}) module parameterized by $\psi$. We utilize a LoRA parameterization rather than a full-rank conditional matrix due to strict computational constraints. For a latent dimension $N_z$, a full-rank regime-conditioned operator requires $O(N_z^2)$ parameters, leading to severe VRAM fragmentation during batched training. LoRA reduces this to $O(2rN_z)$, acting as a structural regularizer that prevents $\mathbf{K}_{\text{cont}}$ from deviating too radically. 
This parameterization implicitly constrains regime-specific dynamics to remain close to a shared global flow manifold, improving stability when extrapolating across unseen physical regimes.
During training, we inject small Gaussian noise into the conditioning parameters $\phi$. This acts as a regularizer that improves robustness to sparse parameter sampling and discretization artifacts. In practice, this encourages the learned Koopman operator to vary smoothly across parameter space, improving interpolation between regimes.

\section{Data Details}
\label{appendix:datasets_extended}

To rigorously evaluate the proposed continuous-time Koopman framework, we introduce a suite of dynamical system datasets ranging from canonical fluid flow benchmarks to chaotic spatiotemporal PDEs. All models were trained on the public data from  \citep{kohl2026benchmarking} or on synthetical, locally generated datasets simulating the Kuramoto–Sivashinsky equation. For a thorough comparison, we used the exact same split in terms of train/evaluation. The incompressible model was trained on all the sequences with Reynolds numbers in the $[200, 900]$ interval with a step of $10$, and evaluated on the strictly disjoint $Inc_{low} = [100, 180]$, $Inc_{high} = [920, 1000]$ extreme intervals.

The transonic model was also trained on the sequences with Mach numbers in the $[0.53,0.54,...,0.62,0.63] \cup [0.69,0.70,...,0.89,0.90]$ and evaluated on the sequences with $Tra_{ext} = [0.50,0.51,0.52]$, $Tra_{int} = [0.66,0.67,0.68]$ for $60$ time steps and $Tra_{long} = [0.64,0.65]$ for $240$ time steps. The main difference in training and loading the data is that we treat them in an exhaustive manner. We follow the same stride (a sub-sampling factor of $2$), but we load the data such that the previous sequence's prediction can be initial conditions for the following sequence. 

Moreover, we used a different structure. The original Diffusion Model was using a simple tensor with multiple channels, each of them representing a different variable ($v_x$, $v_y$, etc.). For expressiveness we used TensorDicts, which are PyTorch's \citep{paszke2019pytorch} version of dictionaries, where the values are tensors. Each of the channels from the original formulation was translated into a specific variable.

\subsection{Incompressible Wake Flow}\label{app:inc_flow_data}
This dataset consists of a fully developed incompressible Karman vortex street behind a cylindrical obstacle with varying Reynolds numbers $Re \in [100,1000]$, comprising 91 simulations in total \citep{kohl2026benchmarking}. For training, both the baseline models and our approach utilize simulations within the range $Re \in [200, 900]$. There are two distinct test sets used to measure out-of-distribution extrapolation: $Inc_{low}$, covering $Re \in [100, 180]$, and $Inc_{high}$, covering $Re \in [920, 1000]$. Both test sets use a prediction horizon of $T=60$ steps. While the raw datasets employ a time step of $\Delta t = 0.05\text{s}$, we follow the baseline protocol and subsample the trajectories by a factor of 2, resulting in an effective time step of $\Delta t = 0.1\text{s}$ for model training and evaluation.

\subsection{Transonic Cylinder Flow}\label{app:tra_flow_data}
Transonic flows exhibit highly chaotic behavior, characterized by the formation of shock waves that dynamically interact with the fluid flow \citep{kohl2026benchmarking}. The dataset consists of 40 sequences of fully developed compressible Karman vortex streets at a fixed Reynolds number $Re=10,000$, with Mach numbers varying across the range $Ma \in [0.50, 0.90]$ . Models are trained on sequences with $Ma \in [0.53,0.63] \cup [0.69,0.90]$. For evaluation, two test sets, $Tra_{ext}$ and $Tra_{int}$, represent simulations with $Ma \in [0.50,0.52]$ and $Ma \in [0.66,0.68]$ respectively, with standard rollout steps of $T=60$. To test extreme stability over roughly 8 vortex shedding periods, an additional long rollout case, $Tra_{long}$, is defined with $Ma \in [0.64,0.65]$ and $T=240$. Similar to the incompressible case, the raw simulation interval is $\Delta t = 0.05\text{s}$, subsampled to $\Delta t = 0.1\text{s}$ for our experiments.

\subsection{1D Kuramoto-Sivashinsky (KS) Equation}\label{app:ks_data}
To rigorously validate our continuous-time formulation's ability to handle irregularly sampled data and chaotic phase turbulence, we evaluate our model on the Kuramoto-Sivashinsky (KS) equation. The dynamics are governed by the following partial differential equation:
$$u_t = -uu_x - u_{xx} - u_{xxxx} + f(x, t)$$
where $f(x, t)$ represents an optional external forcing term. For our primary experiments, we simulate the unforced system ($f = 0$). The spatial domain is defined with periodic boundary conditions over a length $L = 22.0$, a standard regime for evaluating chaotic KS dynamics. 

The system is numerically integrated using a pseudospectral method for spatial discretization ($n_x = 64$ points, utilizing the 2/3 rule for dealiasing the nonlinear term) and an explicit Runge-Kutta method (\texttt{RK45}, $\text{rtol}=10^{-6}$, $\text{atol}=10^{-8}$) for time stepping. We generate 100 independent trajectories, each integrated for 1000 steps with a sampling interval of $\Delta t = 0.1$. To align with standard spatial resolutions and convolutional frameworks, the final outputs are downsampled to 64 spatial grid points and cast with a dummy spatial dimension. The resulting dataset is split into 80\% training, 10\% validation, and 10\% testing subsets.

\section{Full Models comparison}
\label{appendix:full_comparison}

\begin{table*}[t]
\centering
\resizebox{0.98\textwidth}{!}{%
\begin{tabular}{@{}l cc cc cc cc cc cc@{}}
\toprule
\multirow{2}{*}{\textbf{Method}} 
& \multicolumn{2}{c}{\textbf{$Inc_{low}$}} 
& \multicolumn{2}{c}{\textbf{$Inc_{high}$}}
& \multicolumn{2}{c}{\textbf{$Tra_{ext}$}}
& \multicolumn{2}{c}{\textbf{$Tra_{int}$}}
& \multicolumn{2}{c}{\textbf{$Tra_{long}$}}
& \multirow{2}{*}{\makecell{\textbf{Avg. Step} \\ \textbf{(ms)}}}
& \multirow{2}{*}{\makecell{\textbf{Mean VRAM} \\ \textbf{(MB)}}} \\
\cmidrule(lr){2-3}
\cmidrule(lr){4-5}
\cmidrule(lr){6-7}
\cmidrule(lr){8-9}
\cmidrule(lr){10-11}
& \makecell{MSE \\ $(\times 10^{-4})$} & \makecell{LSiM \\ $(\times 10^{-2})$}
& \makecell{MSE \\ $(\times 10^{-5})$} & \makecell{LSiM \\ $(\times 10^{-2})$}
& \makecell{MSE \\ $(\times 10^{-3})$} & \makecell{LSiM \\ $(\times 10^{-1})$}
& \makecell{MSE \\ $(\times 10^{-3})$} & \makecell{LSiM \\ $(\times 10^{-1})$}
& \makecell{MSE \\ $(\times 10^{-3})$} & \makecell{LSiM \\ $(\times 10^{-1})$} 
& & \\ 
\midrule

ResNet 
& $10 \pm 9.1$ & $17 \pm 7.8$
& $16 \pm 3.0$ & $5.9 \pm 1.6$
& $2.3 \pm 0.9$ & $1.4 \pm 0.2$
& $1.8 \pm 1.0$ & $\mathbf{1.0 \pm 0.3}$
& $24.2 \pm 4.6$ & $7.6 \pm 1.7$ 
& $3.67$ & $188.0$ \\

ResNet-dil 
& $1.6 \pm 1.8$ & $7.7 \pm 5.5$
& $1.5 \pm 0.8$ & $2.6 \pm 0.7$ 
& $1.7 \pm 1.0$ & $1.2 \pm 0.3$ 
& $1.7 \pm 1.4$ & $1.0 \pm 0.5$ 
& $22.0 \pm 2.4$ & $5.5 \pm 2.3$ 
& $3.46$ & $\mathbf{178.6}$ \\

$\text{FNO}_{16}$ 
& $2.8 \pm 3.1$ & $8.8 \pm 7.1$
& $8.9 \pm 3.8$ & $2.5 \pm 1.2$
& $4.8 \pm 1.2$ & $3.4 \pm 1.1$
& $5.5 \pm 2.6$ & $2.6 \pm 1.1$
& $20.8 \pm 2.0$ & $11.5 \pm 1.1$ 
& $1.17$ & $184.1$ \\

$\text{FNO}_{32}$ 
& $160 \pm 50$ & $80 \pm 5.4$
& $1k \pm 140$ & $57 \pm 4.9$
& $4.9 \pm 1.9$ & $3.6 \pm 0.9$
& $6.8 \pm 3.4$ & $3.1 \pm 1.1$
& \text{Diverged} & \text{Diverged} 
& $1.17$ & $183.9$ \\

$\text{TF}_{Enc}$ 
& $1.5 \pm 1.7$ & $6.3 \pm 4.2$
& $0.6 \pm 0.3$ & $1.0 \pm 0.3$ 
& $3.3 \pm 1.2$ & $1.8 \pm 0.3$ 
& $6.2 \pm 4.2$ & $2.2 \pm 0.7$
& $22.2 \pm 3.9$ & $3.8 \pm 0.4$ 
& $0.60$ & $3448.6$ \\

$\text{TF}_{MGN}$ 
& $5.7 \pm 4.3$ & $13 \pm 6.4$
& $10 \pm 2.9$ & $3.5 \pm 0.4$
& $3.9 \pm 1.0$ & $1.8 \pm 0.3$
& $6.3 \pm 4.4$ & $2.2 \pm 0.7$
& $18.9 \pm 4.5$ & $4.0 \pm 0.2$ 
& $0.69$ & $3498.0$ \\

$\text{TF}_{VAE}$ 
& $5.4 \pm 5.5$ & $13 \pm 7.2$
& $14 \pm 19$ & $4.1 \pm 1.4$ 
& $4.1 \pm 0.9$ & $2.4 \pm 0.2$ 
& $7.2 \pm 3.0$ & $2.7 \pm 0.6$ 
& $20.6 \pm 2.1$ & $4.0 \pm 0.2$ 
& $\mathbf{0.30}$ & $13749.9$ \\

U-Net 
& $1.0 \pm 1.1$ & $5.8 \pm 3.2$
& $2.7 \pm 0.6$ & $2.6 \pm 0.6$
& $3.1 \pm 2.1$ & $3.9 \pm 2.8$
& $2.3 \pm 2.0$ & $3.3 \pm 2.8$
& $30.3 \pm 6.1$ & $9.1 \pm 1.2$ 
& $6.19$ & $183.7$ \\

$\text{U-Net}_{ut}$  
& $\mathbf{0.8 \pm 1.1}$ & $\mathbf{4.5 \pm 4.0}$
& $\mathbf{0.2 \pm 0.1}$ & $\mathbf{0.5 \pm 0.2}$
& $1.6 \pm 0.7$ & $\mathbf{1.1 \pm 0.2}$
& $\mathbf{1.5 \pm 1.5}$ & $1.0 \pm 0.5$
& $22.2 \pm 3.6$ & $3.8 \pm 0.4$ 
& $6.16$ & $184.1$ \\

$\text{U-Net}_{tn}$  
& $1.0 \pm 1.0$ & $5.6 \pm 3.1$ 
& $0.9 \pm 0.6$ & $1.5 \pm 0.6$
& $\mathbf{1.4 \pm 0.8}$ & $1.1 \pm 0.3$
& $1.8 \pm 1.1$ & $1.0 \pm 0.4$
& $22.4 \pm 3.9$ & $3.9 \pm 0.4$ 
& $6.16$ & $184.1$ \\

Refiner 
& $1.3 \pm 1.4$ & $7.1 \pm 4.2$
& $3.5 \pm 2.2$ & $2.5 \pm 1.0$
& $5.4 \pm 2.1$ & $2.3 \pm 0.5$
& $7.1 \pm 2.1$ & $3.0 \pm 1.7$
& \text{Diverged} & $8.8 \pm 3.3$ 
& $10.31$ & $642.4$ \\

$\text{ACDM}_{ncn}$ 
& $0.9 \pm 0.8$ & $6.6 \pm 2.7$
& $5.6 \pm 2.7$ & $3.6 \pm 1.2$
& $4.1 \pm 1.9$ & $1.9 \pm 0.6$
& $2.8 \pm 1.3$ & $1.7 \pm 0.4$
& $22.8 \pm 3.8$ & $8.2 \pm 3.0$ 
& $126.3$ & $649.2$ \\

ACDM 
& $1.7 \pm 2.2$ & $6.9 \pm 5.6$
& $0.8 \pm 0.4$ & $1.0 \pm 0.3$
& $2.3 \pm 1.4$ & $1.3 \pm 0.3$
& $2.7 \pm 2.1$ & $1.3 \pm 0.6$
& $22.6 \pm 4.0$ & $\mathbf{3.8 \pm 0.4}$ 
& $126.57$ & $659.2$ \\

KAE
& $1.3 \pm 1.7$ & $6.1 \pm 4.8$
& $2.9 \pm 1.1$ & $1.7 \pm 0.3$
& $2.2 \pm 0.9$ & $1.8\pm 0.3$
& $5.2 \pm 2.4$ & $2.1 \pm 0.6$
& $\mathbf{14.9 \pm 1.3}$ & $4.1 \pm 0.3$ 
& $1.15$ & $2751.3$ \\

Exponential KAE
& $1.3 \pm 1.7$ & $6.1 \pm 4.8$
& $2.9 \pm 1.1$ & $1.7 \pm 0.3$
& $2.2 \pm 0.9$ & $1.8\pm 0.3$
& $5.2 \pm 2.4$ & $2.1 \pm 0.6$
& $\mathbf{14.9 \pm 1.3}$ & $4.1 \pm 0.3$ 
& $1.15$ & $2751.3$ \\

\bottomrule
\end{tabular}%
}
\caption{Quantitative comparison across incompressible and transonic flow regimes alongside computational efficiency. Metrics reported include MSE and LSiM, as well as average per-step inference time and VRAM footprint. This view highlights KAE's state-of-the-art long-horizon stability ($Tra_{long}$, evaluated over an extended 240-step horizon) combined with its massive inference speedup over diffusion baselines.}
\label{tab:quantitative_split_comparison}
\end{table*}

\section{Implementation and Model Details}
The entire model is implemented and trained using PyTorch \citep{paszke2019pytorch}. All wights in the architecture were optimized using AdamW \citep{loshchilov2017decoupled} with $\beta_1 = 0.9$ and $\beta_2 = 0.999$. The learning rate follows a Cosine Warm-up schedule, linear increasing for the first $20$ epochs from $0.0$ to $5\times 10^{-4}$, then decreasing on a cosine curve to $10^{-5}$ for the following $180$ epochs, making it $200$ epochs in total for training, with a batch size of $64$. 

We also follow an end-to-end training procedure, learning the encoder, operator and decoder simultaneously. While testing different architectures and training schemes, we have experimented with curriculum learning too. As the model can be rather complex, especially for large latent dimensionality, while the data, more specifically the transonic flow, can be very chaotic and rough, we tried training the Auto-Encoder's core (encoder and decoder) for the first $50$ epochs alone, freezing the operator's side, as we first wanted to build a smooth manifold for the data. For the following $100$ epochs, we have frozen the encoder and decoder and only trained the operator. Finally, for the remaining epochs, we fine-tuned them together. However, this proved challenging as regularizing the latent space was not as straightforward as expected, leaving the linear operator to struggle.

Our model follows a modular $encoder\mbox{--} latent\mbox{--} dynamic\mbox{--} decoder$ design, combining convolutional feature extraction with physics-conditioned Koopman operators for latent-space temporal evolution. The architecture is specifically designed for stability, interpretability, and robustness in strongly conditioned fluid dynamics regimes.

\begin{figure*}[ht]
    \centering
    \resizebox{0.7\textwidth}{!}{%
    \begin{tikzpicture}[
        >=Stealth,
        node distance=0.35cm and 0.5cm,
        scale=0.75, transform shape,
        tensor/.style={draw, rectangle, rounded corners=2pt, minimum width=1.0cm, minimum height=0.5cm, align=center, fill=white, font=\small},
        base_trap/.style={draw, trapezium, trapezium angle=70, minimum height=0.6cm, align=center, font=\footnotesize},
        encoder/.style={base_trap, shape border rotate=180, minimum width=1.5cm, fill=blue!5},
        decoder/.style={base_trap, minimum width=0.9cm, fill=red!5},
        op/.style={draw, circle, inner sep=0pt, minimum size=0.3cm, fill=white},
        label/.style={font=\tiny}
    ]
        \node[tensor] (xt) {$x_{t_i}$};
        \node[encoder, below=0.3cm of xt] (E_pres) {$\mathcal{E}_{present}$};
        \draw[->] (xt) -- (E_pres);
        \node[op, below=0.6cm of E_pres] (sum) {$+$};
        \node[tensor, below=0.3cm of sum] (zt0) {$z_{t_i}$};
        \draw[->] (sum) -- (zt0);
        \node[decoder, below=0.35cm of zt0] (D0) {$\mathcal{D}$};
        \draw[->] (zt0) -- (D0);
        \node[tensor, below=0.3cm of D0] (xhat0) {$\hat{x}_{t_i}$};
        \draw[->] (D0) -- (xhat0);
        \node[encoder, left=0.4cm of E_pres] (E_hist) {$\mathcal{E}_{history}$};
        \node[tensor, above=0.3cm of E_hist] (xt_prev) {$x_{t_{i-1}}$};
        \draw[->] (xt_prev) -- (E_hist);
        \draw[->] (E_pres.south) -- (sum.north) node[midway, right, font=\tiny, xshift=1pt] {$\times \frac{1}{2}$};
        \draw[->] (E_hist.south) -- (E_hist.south |- sum.west) node[midway, left, font=\tiny] {$\times \frac{1}{2}$} -- (sum.west);
        \node[tensor, right=1.4cm of zt0] (zt1) {$z_{t_{i+1}}$};
        \draw[->, thick] (zt0) -- (zt1) node[midway, above, font=\scriptsize] {$\mathcal{K}_{cont}$};
        \node[tensor, right=1.4cm of zt1] (zt2) {$z_{t_{i+2}}$};
        \draw[->, thick] (zt1) -- (zt2);
        \node[right=0.8cm of zt2] (dots) {$\cdots$};
        \draw[->, thick] (zt2) -- (dots);
        \node[tensor, right=0.8cm of dots] (zt8) {$z_{t_{i+N}}$};
        \draw[->, thick] (dots) -- (zt8);
        \node[decoder, below=0.35cm of zt1] (D1) {$\mathcal{D}$};
        \node[tensor, below=0.3cm of D1] (xhat1) {$\hat{x}_{t_{i+1}}$};
        \draw[->] (zt1) -- (D1); \draw[->] (D1) -- (xhat1);
        \node[decoder, below=0.35cm of zt2] (D2) {$\mathcal{D}$};
        \node[tensor, below=0.3cm of D2] (xhat2) {$\hat{x}_{t_{i+2}}$};
        \draw[->] (zt2) -- (D2); \draw[->] (D2) -- (xhat2);
        \node[decoder, below=0.35cm of zt8] (D8) {$\mathcal{D}$};
        \node[tensor, below=0.3cm of D8] (xhat8) {$\hat{x}_{t_{i+N}}$};
        \draw[->] (zt8) -- (D8); \draw[->] (D8) -- (xhat8);
    \end{tikzpicture}%
    }
    \caption{Architecture overview. The history encoder and present encoder (top) produce the initial latent state $z_{t_i}$, which is propagated by the continuous Koopman operator $\mathcal{K}_{\mathrm{cont}}$ and decoded at each rollout step.}
    \label{fig:architecture}
\end{figure*}

\subsection{Encoder Architecture \label{app:architecure}}

The spatial encoder is a convolutional residual encoder operating on input fields augmented with explicit coordinate information. For each input snapshot of shape $C \times H \times W$, two additional coordinate channels corresponding to normalized spatial coordinates are concatenated, resulting in $C+2$ input channels. This coordinate injection enables the model to reason about absolute spatial position without relying on implicit convolutional biases.

The encoder backbone begins with an initial $3\times3$ convolution, followed by a sequence of pre-activation residual blocks \citep{he2016identity} with Group Normalization \citep{wu2018group} and SiLU activations. Each resolution level consists of one downsampling residual block with stride $2$, and one refinement residual block with stride $1$.

We use three resolution levels with channel widths $[64, 128, 256]$, leading to an effective downsampling factor of $2^3=8$. All convolutional and linear layers in the encoder employ spectral normalization, which we found crucial for stabilizing gradients in downstream Koopman dynamics learning.

At the lowest spatial resolution, we apply a Convolutional Block Attention Module (CBAM) \citep{woo2018cbam}, consisting of sequential channel-wise and spatial attention. This allows the encoder to focus on dynamically active regions such as vortices while suppressing background noise.

The resulting feature map is flattened and projected into a latent representation via a linear layer. This latent representation significantly improves numerical stability of the Koopman operator. When late fusion conditioning is enabled, the flattened spatial features are concatenated with a physics embedding before the final projection.

Conditioning and Physics Parameter Embeddings
Physical parameters (e.g. Reynolds number, Mach number, forcing terms) are incorporated through radial basis function (RBF)–style expansions, mapping low-dimensional scalars into higher-dimensional embeddings. The expansion type is configurable and shared across encoder, decoder, and dynamics modules. During training, small Gaussian noise can be injected into the conditioning variables to improve robustness to discretization artifacts and sparse parameter sampling.

\subsubsection{Temporal History Encoder}\label{app:history_enc}
For scenarios involving multiple input timesteps, we introduce a history encoder that aggregates temporal context before latent evolution. Each timestep is independently encoded using the convolutional encoder backbone, producing a sequence of latent vectors. These vectors are then processed by a Transformer encoder \citep{vaswani2017attention} with sinusoidal positional encodings. The Transformer output is averaged across time to obtain a single latent context vector, which serves as the initial condition for the Koopman dynamics. This design allows the model to capture temporal correlations without explicitly unrolling convolutional operations over time.

\subsection{Decoder Architecture}\label{app:dec_arch}

The architectural asymmetry between the temporal Transformer encoder and the convolutional decoder is highly intentional. While the Transformer globally aggregates temporal dependencies across the 1D time axis, the decoder must reconstruct the physical 2D flow fields. Retaining CNNs in the decoder enforces the spatial translation invariance and local inductive biases necessary to cleanly map the latent state back to the physical grid without artifacts. The latent vector is first linearly expanded and reshaped into a low-resolution feature map. Upsampling proceeds through a sequence of resolution levels $[256, 128, 64]$, each consisting of a conditioned residual block, and a resize–convolution upsampling block (nearest-neighbor upsampling followed by a $3\times3$ convolution).

To integrate physical parameters, we employ Adaptive Group Normalization (AdaGN) within the decoder residual blocks. Here, scale and shift parameters of the normalization layers are modulated by the physics embeddings, enabling strong, spatially uniform conditioning while preserving convolutional locality. Dropout is applied inside conditioned residual blocks for regularization, using element-wise dropout to avoid suppressing entire physical channels. The final output is produced via a $3\times3$ convolution mapping back to the original number of physical variables.

\subsection{Koopman Dynamics Module}\label{app:koop_dyn}
Latent evolution is governed by a physics-conditioned Koopman operator implemented in continuous-time form. We strictly enforce a linear Koopman parameterization, where the dynamics are governed by a base linear operator with low-rank, condition-dependent updates predicted by a hyper-network (LoRA-style adaptation) \citep{hu2022lora}. Avoiding a nonlinear MLP ensures the global linearity of the latent space remains intact, which is mathematically required to utilize the $O(1)$ matrix exponential at inference.

Condition dependence is handled via a hyper-network that predicts operator modifications from the physics embeddings. A dissipative inductive bias is encouraged through dissipative initialization and spectral regularization, though stability is not enforced as a hard constraint---it emerges empirically from rollout training.

For continuous dynamics, we model
$\frac{d\mathbf{z}}{dt} = K({\phi}) \mathbf{z}$,
and integrate using a fourth-order Runge–Kutta (RK4) scheme by default, with optional implicit midpoint or Radau IIA solvers for stiff regimes. All implicit solvers internally upcast to $float32$ to ensure numerical robustness.

\section{Training Objective and Loss Function Details}
\label{appendix:loss}

To ensure that the learned latent dynamics are not only accurate in terms of Euclidean error but also physically consistent, stable, and topologically faithful to the fluid flows, we employ a composite loss function. The total objective $\mathcal{L}_{\text{total}}$ is a weighted sum of reconstruction accuracy, temporal rollout consistency, latent space regularization, and physics-conditioned constraints:

\begin{equation}
    \mathcal{L}_{\text{total}} = \mathcal{L}_{\text{recon}} + \alpha \mathcal{L}_{\text{pred}} + \beta \mathcal{L}_{\text{latent}} + \lambda_{\text{phys}} \mathcal{L}_{\text{phys}} 
\end{equation}

\subsection{Loss Weighting Strategy}\label{app:loss_wights}
The scalar weights $\alpha, \beta$, and $\lambda_{phys}$ are critical for balancing the trade-offs between spatial reconstruction fidelity, temporal stability, and adherence to theoretical Koopman constraints.

\begin{itemize}
    \item \textbf{Reconstruction Base ($\mathcal{L}_{recon}$):} Kept at a unit weight, this anchors the overall scale of the loss. It is heavily prioritized in the early epochs to ensure the encoder and decoder construct a valid spatial manifold before temporal dynamics are strictly enforced.
    \item \textbf{Rollout Prediction ($\alpha$):} This weight dictates the model's resistance to autoregressive error accumulation. A sufficiently high $\alpha$ forces the Koopman operator to learn a globally stable trajectory rather than a greedy one-step mapping. However, if $\alpha$ dominates too early in training, the network struggles to converge on the underlying spatial representations. 
    \item \textbf{Latent Consistency ($\beta$):} This parameter controls the strictness of the theoretical Koopman constraints (detailed below). Balancing $\beta$ is crucial: if it is too low, the latent space evolves non-linearly, risking divergence during long rollouts; if it is too high, it over-constrains the autoencoder, resulting in overly smoothed spatial reconstructions that fail to capture complex flow features. 
    \item \textbf{Physics Regularization ($\lambda_{phys}$):} Because spatial gradients and spectral amplitudes operate on different numerical scales than standard pixel-wise MSE, $\lambda_{phys}$ scales these high-frequency penalties. It acts as a structural fine-tuning mechanism to sharpen edges and correct phase shifts once the base dynamics are established.
\end{itemize}

Below, we detail the mathematical formulation of the latent and physics-insipred components implemented in our framework.

\subsection{Latent Space Consistency}\label{appendix:consisteny_loss}

To ensure the latent manifold respects the theoretical properties of the Koopman operator, we enforce structural constraints via $\mathcal{L}_{\text{latent}}$. This includes three specific sub-terms:

\textbf{1. Forward-Backward Linearity Consistency.} 
Inspired by the discrete-time constraints of \citet{azencot2020consistent}, we enforce a continuous-time generalization of operator invertibility. If $\mathbf{K}(\Delta t)$ propagates the state forward, integrating backward via $\mathbf{K}(-\Delta t)$ must recover the previous state. This ensures the continuous generator does not learn a trivial "shrink-to-zero" solution.
\begin{equation}\label{eq:consistency_loss}
    \mathcal{L}_{\text{lin}} = \underbrace{\| \mathbf{K}(z_t, \Delta t) - z_{t+1} \|_2^2}_{\text{Forward}} + \underbrace{\| \mathbf{K}(z_{t+1}, -\Delta t) - z_t \|_2^2}_{\text{Backward}}.
\end{equation}

\textbf{2. Directional Cosine Similarity.} 
To decouple magnitude errors (decay) from directional errors (dynamics), we enforce alignment between the predicted latent vector update and the true encoder trajectory:
\begin{equation}
    \mathcal{L}_{\text{cos}} = 1 - 
    \frac{z_{t+1} \cdot z_{t+1}^{\text{target}}}{\|z_{t+1}\| \|z_{t+1}^{\text{target}}\|}
\end{equation}

\textbf{3. Energy Conservation Regularization.}
Ideally, the "energy" (norm) of the latent state should evolve smoothly. We penalize abrupt changes in the latent norm to encourage smooth trajectories:
\begin{equation}
    \mathcal{L}_{\text{energy}} = \left( \|z_{t+1}\|_2 - \|z_t\|_2 \right)^2.
\end{equation}
This constraint acts as a soft regularizer against rapid latent norm explosion beyond the training horizon, rather than enforcing strict energy conservation which would contradict the desired dissipative nature of the linear ODE. Without it, the continuous latent norm remains unconstrained beyond the training horizon, leading to catastrophic divergence when rolling out 30x beyond the training data. The total latent loss is given by $\mathcal{L}_{\text{latent}} = \mathcal{L}_{\text{lin}} + w_{cos}\mathcal{L}_{\text{cos}} + \mathcal{L}_{\text{energy}}$.

\subsection{Physics-Inspired Loss}\label{app:phys_loss}

Standard MSE losses often result in "pacing" errors (phase shift) or blurred edges. We mitigate this using a physics-conditioned loss $\mathcal{L}_{\text{phys}}$ comprising Sobolev norms and spectral analysis.

\textbf{Temporal Sobolev Loss (Velocity Matching).} Enforces consistency in the time-derivative (velocity) of the flow:
\begin{equation}
    \mathcal{L}_{\text{time}} = \left\| \frac{\partial \hat{x}}{\partial t} - \frac{\partial x}{\partial t} \right\|_2^2 \approx \frac{1}{\Delta t^2} \left\| (\hat{x}_{t+1} - \hat{x}_t) - (x_{t+1} - x_t) \right\|_2^2.
\end{equation}

\textbf{Spatial Sobolev Loss (Structure Matching).} Enforces consistency in spatial gradients to preserve sharp edges (e.g., shock waves):
\begin{equation}
    \mathcal{L}_{\text{space}} = \|\nabla_x \hat{x} - \nabla_x x\|_2^2 + \|\nabla_y \hat{x} - \nabla_y x\|_2^2.
\end{equation}

\textbf{Spectral Consistency Loss.} 
To correct phase errors and ensure the model captures the correct shedding frequencies, we compute the loss in the frequency domain using the Fast Fourier Transform ($\mathcal{F}$). This term penalizes discrepancies in both amplitude (energy spectrum) and phase:
\begin{equation}
    \mathcal{L}_{\text{spectral}} = \| |\mathcal{F}(\hat{x})| - |\mathcal{F}(x)| \|_1.
\end{equation}

\section{Error Analysis}
\label{appendix:error_analysis}
In this appendix, we provide a detailed quantitative and qualitative comparison between our method and ACDM, focusing on accuracy, stability, and robustness across flow regimes. Beyond aggregate error metrics, we emphasize distributional and temporal error characteristics to better understand when and why our approach outperforms diffusion-based forecasting.

We visualize prediction errors using a combination of time-resolved error curves, error bar plots, and violin plots. Error bars report the mean and standard deviation of normalized $\ell_2$ errors across test trajectories, highlighting both average performance and variability. Violin plots are used to capture the full error distribution over space and time, revealing differences in tail behavior and robustness that are not visible from mean metrics alone.

Across all evaluated scenarios, our model exhibits consistently lower variance in prediction error, indicating improved stability under long-horizon rollouts. In particular, the error distributions produced by ACDM show heavier tails, corresponding to occasional but severe prediction failures, whereas our Koopman-based model yields tighter, more concentrated distributions. This effect is especially pronounced in transonic regimes, where diffusion-based models tend to suffer from mode collapse or over-smoothing.

We further include timestep-sensitivity plots, where the same trained model is evaluated under varying rollout step sizes. These visualizations demonstrate that our continuous-time latent dynamics maintain accuracy across a wide range of timesteps, while ACDM performance degrades significantly when evaluated outside its training resolution. This highlights a key advantage of the continuous-time formulation in terms of temporal generalization.

Finally, spatial error maps are presented to illustrate qualitative differences in failure modes. Our model localizes errors primarily in dynamically active regions such as shocks or vortex cores, while maintaining low background error. In contrast, ACDM exhibits spatially diffuse errors that accumulate over time, consistent with the absence of an explicit latent evolution operator.

\begin{figure*}[ht!]
    \centering
    \begin{subfigure}{0.48\textwidth}
        \centering
        \includegraphics[width=\linewidth]{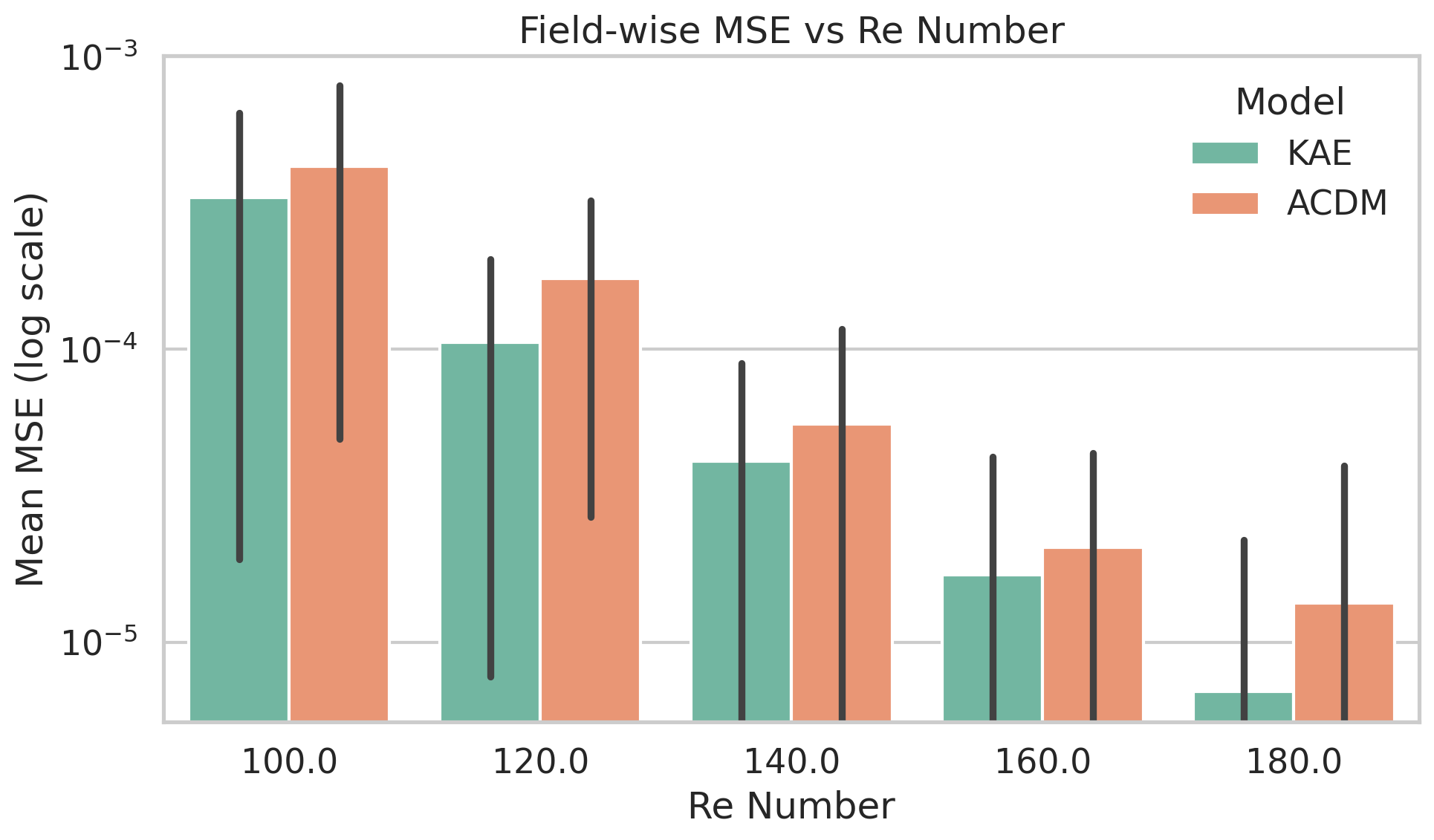}
        \caption{Low Reynolds regime}
        \label{fig:inc_lowRe_bar}
    \end{subfigure}
    \hfill
    \begin{subfigure}{0.48\textwidth}
        \centering
        \includegraphics[width=\linewidth]{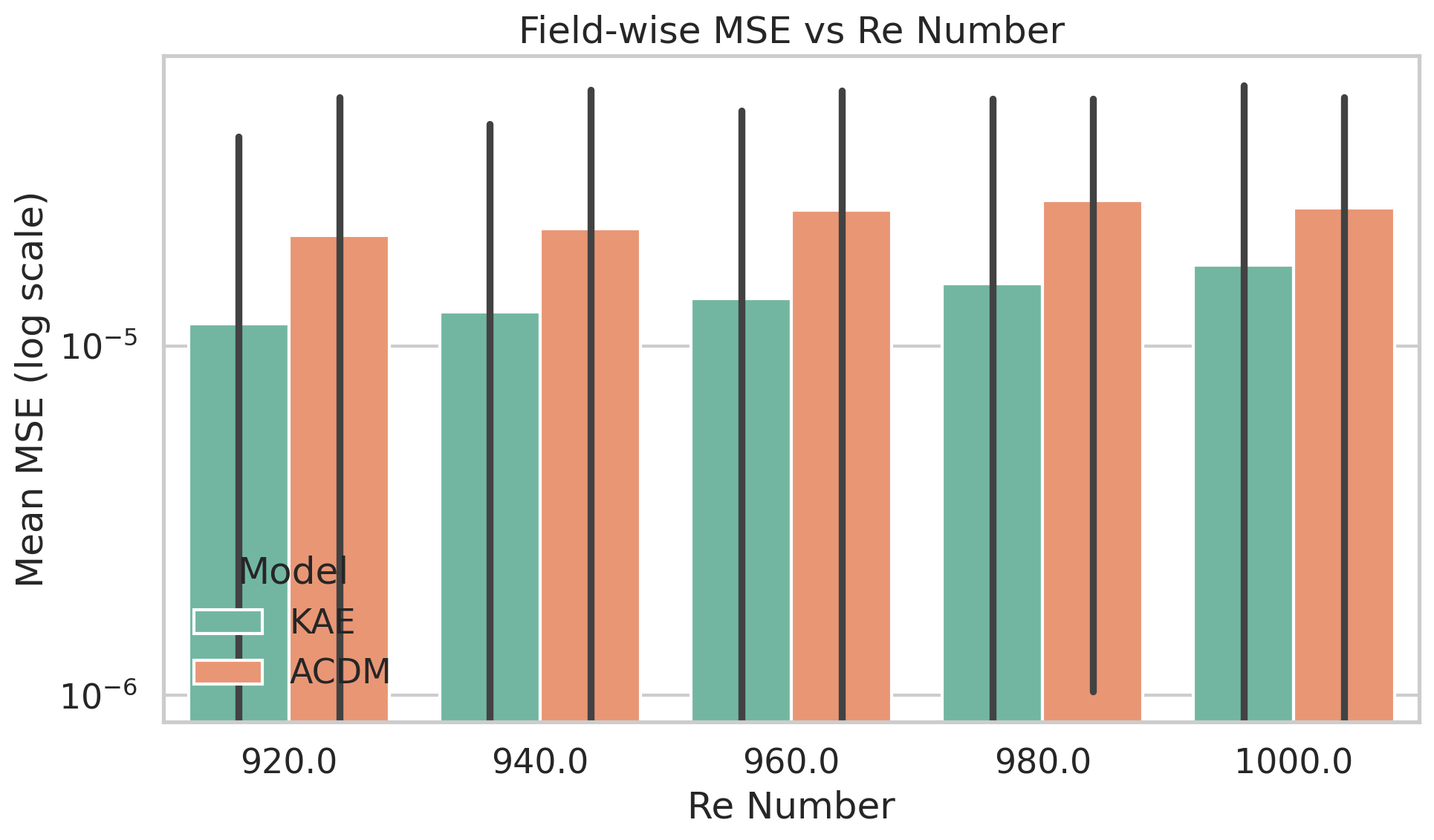}
        \caption{High Reynolds regime}
        \label{fig:inc_highRe_bar}
    \end{subfigure}

    \caption{
    Field-averaged mean squared error (MSE) comparison between KAE and ACDM for incompressible flows.
    Error bars denote one standard deviation across test trajectories.
    Our method consistently achieves lower error and reduced variance, with the performance gap widening in the high-Re regime where chaotic dynamics dominate.
    }
    \label{fig:inc_overview}
\end{figure*}

\begin{figure}[ht!]
    \centering
    \begin{subfigure}{0.48\textwidth}
        \centering
        \includegraphics[width=\linewidth]{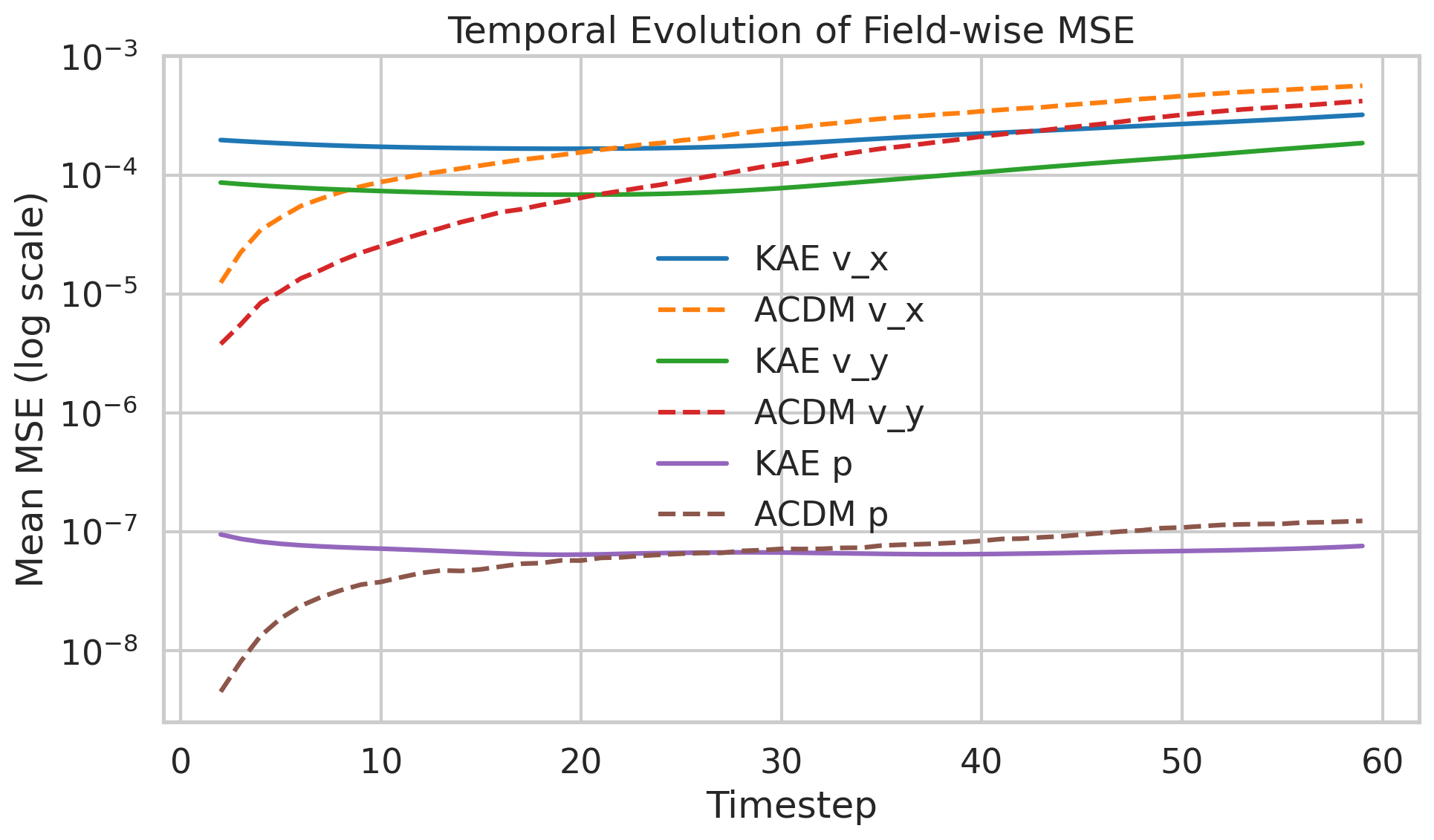}
        \caption{Low Reynolds regime}
        \label{fig:inc_temporal_lowRey}
    \end{subfigure}
    \hfill
    \begin{subfigure}{0.48\textwidth}
        \centering
        \includegraphics[width=\linewidth]{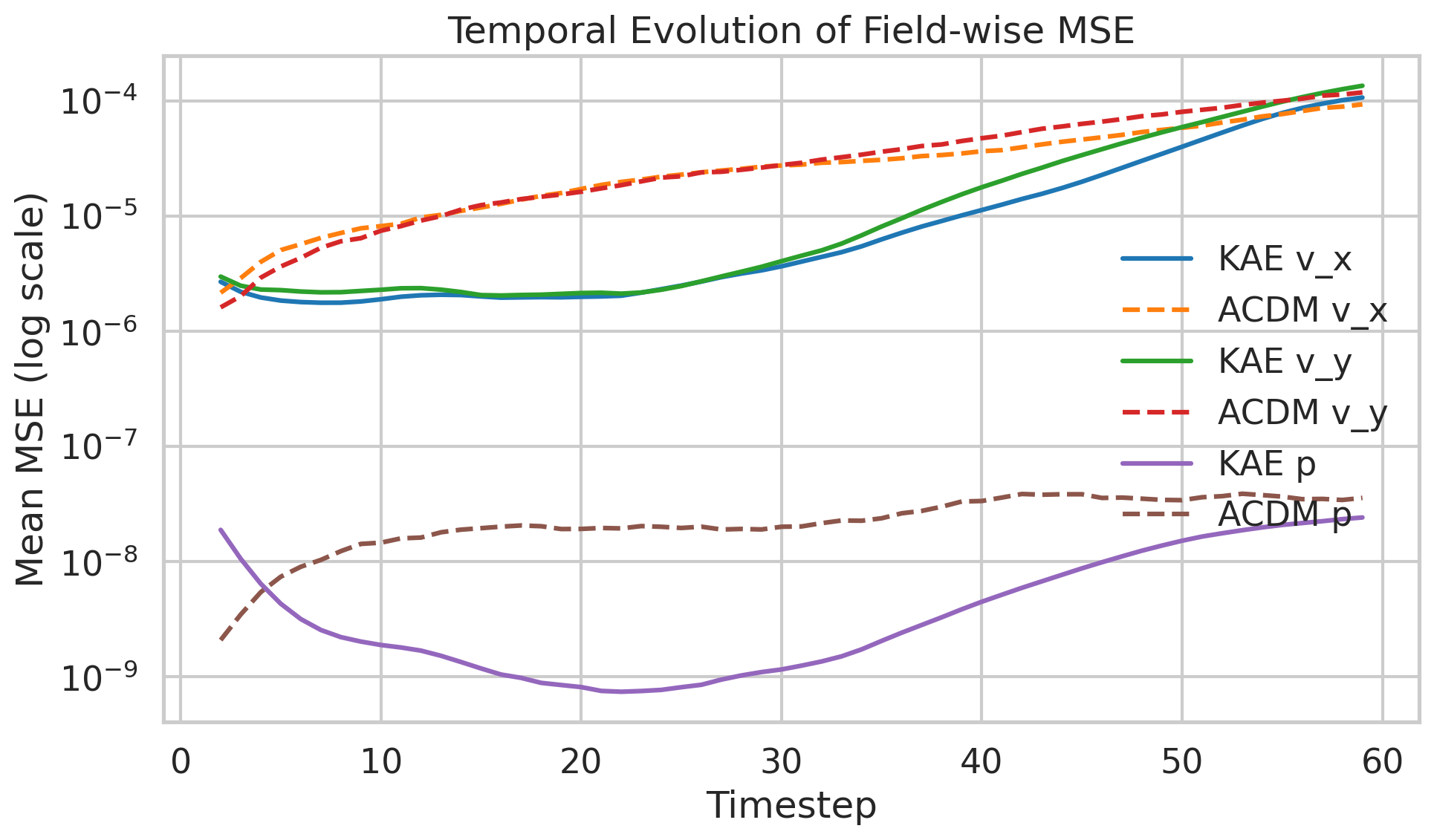}
        \caption{High Reynolds regime}
        \label{fig:inc_temporal_highRey}
    \end{subfigure}
    \caption{
    Temporal evolution of field-wise MSE for incompressible flows.
    Errors are shown on a logarithmic scale.
    KAE maintains stable error growth over long horizons, while ACDM exhibits accelerated error accumulation, indicative of compounding stochastic prediction drift.
    }
    \label{fig:inc_temporal}
\end{figure}

\begin{figure*}[ht!]
    \centering
    \begin{subfigure}{\textwidth}
        \centering
        \includegraphics[width=\linewidth]{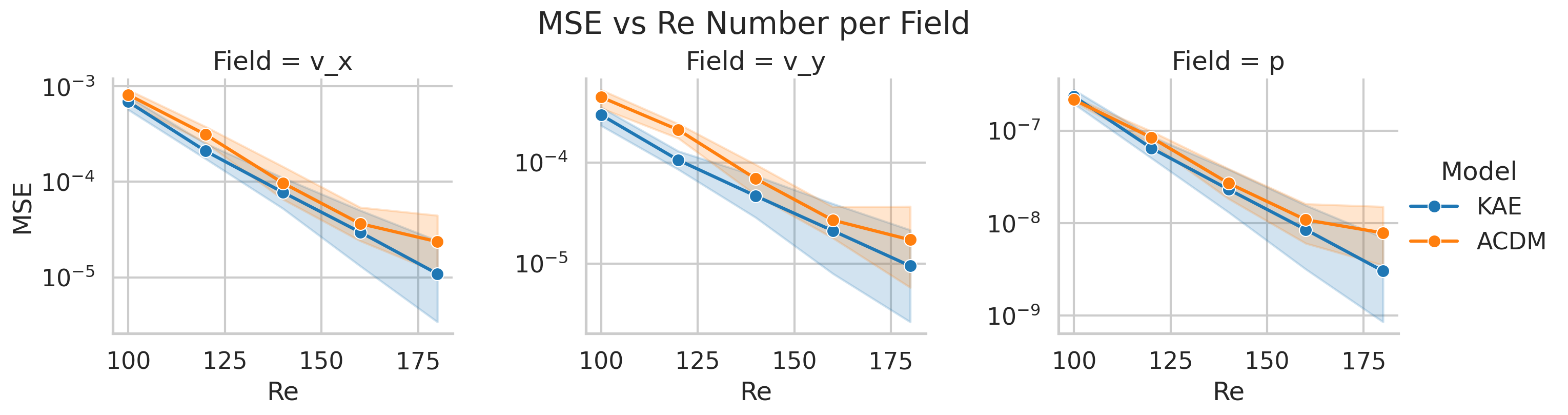}
        \caption{Low Reynolds regime}
        \label{fig:inc_fieldwise_lowRey}
    \end{subfigure}
    \vfill
    \begin{subfigure}{\textwidth}
        \centering
        \includegraphics[width=\linewidth]{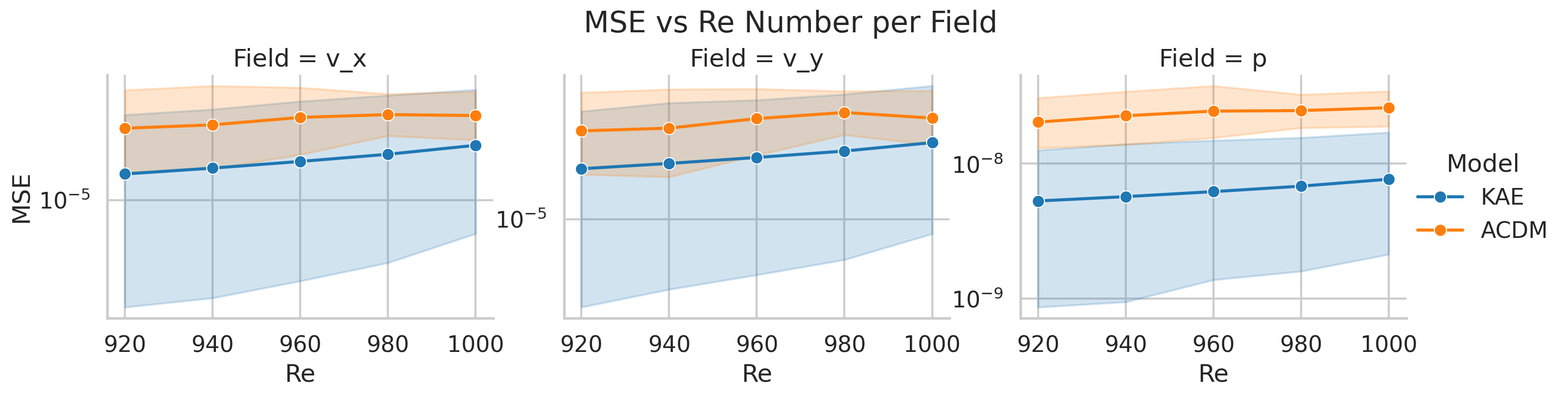}
        \caption{High Reynolds regime}
        \label{fig:inc_fieldwise_highRey}
    \end{subfigure}
    \caption{
    Field-wise MSE as a function of Reynolds number for incompressible flows.
    Velocity components and pressure are shown separately.
    KAE exhibits consistently smoother error scaling with Reynolds number, while ACDM shows increased sensitivity and variance, particularly in pressure-dominated regimes.
    }
    \label{fig:inc_fieldwise}
\end{figure*}

\begin{figure*}[ht!]
    \centering
    \begin{subfigure}{0.48\textwidth}
        \includegraphics[width=\linewidth]{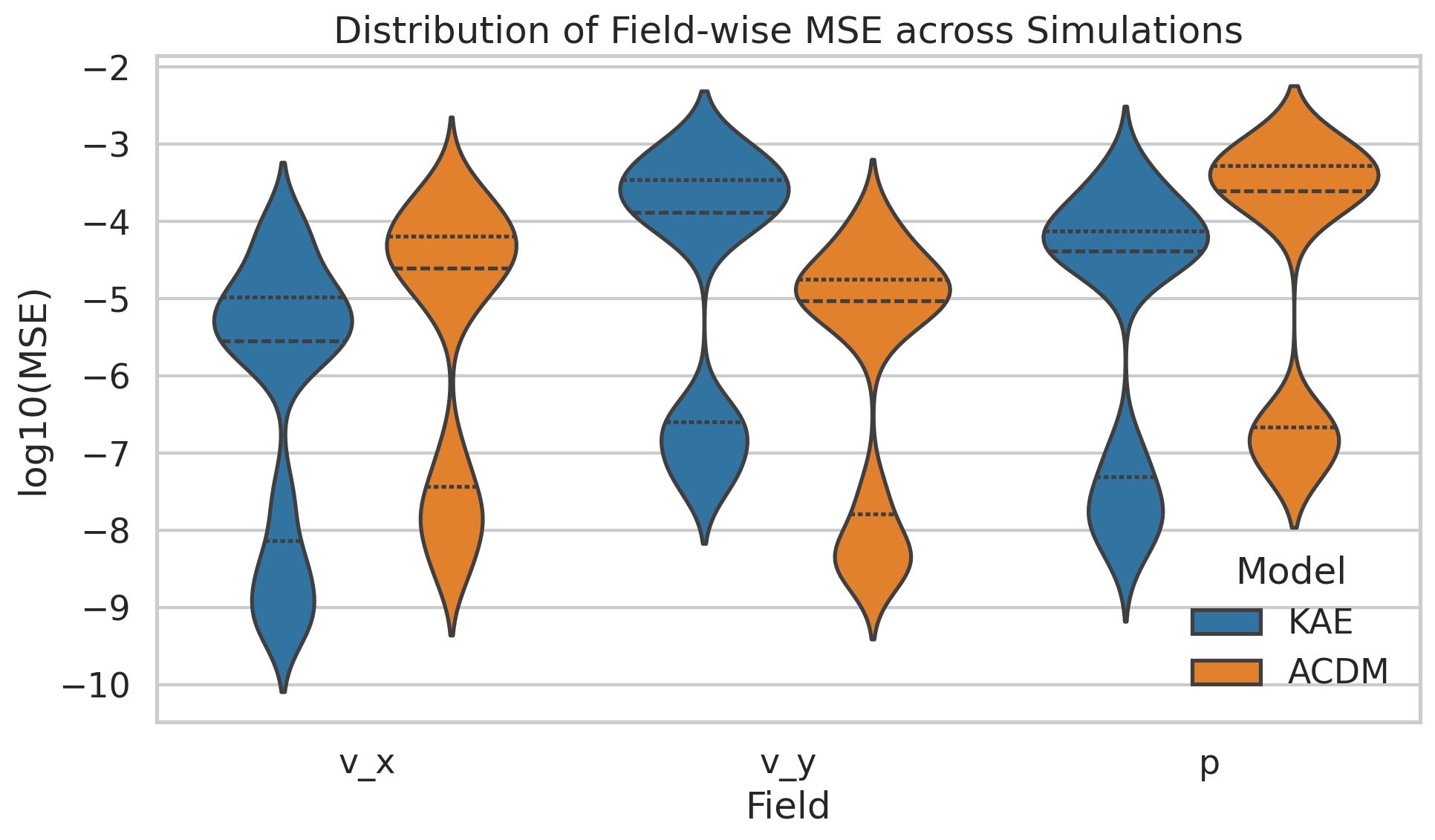}
        \caption{Low Reynolds number}
    \end{subfigure}
    \hfill
    \begin{subfigure}{0.48\textwidth}
        \includegraphics[width=\linewidth]{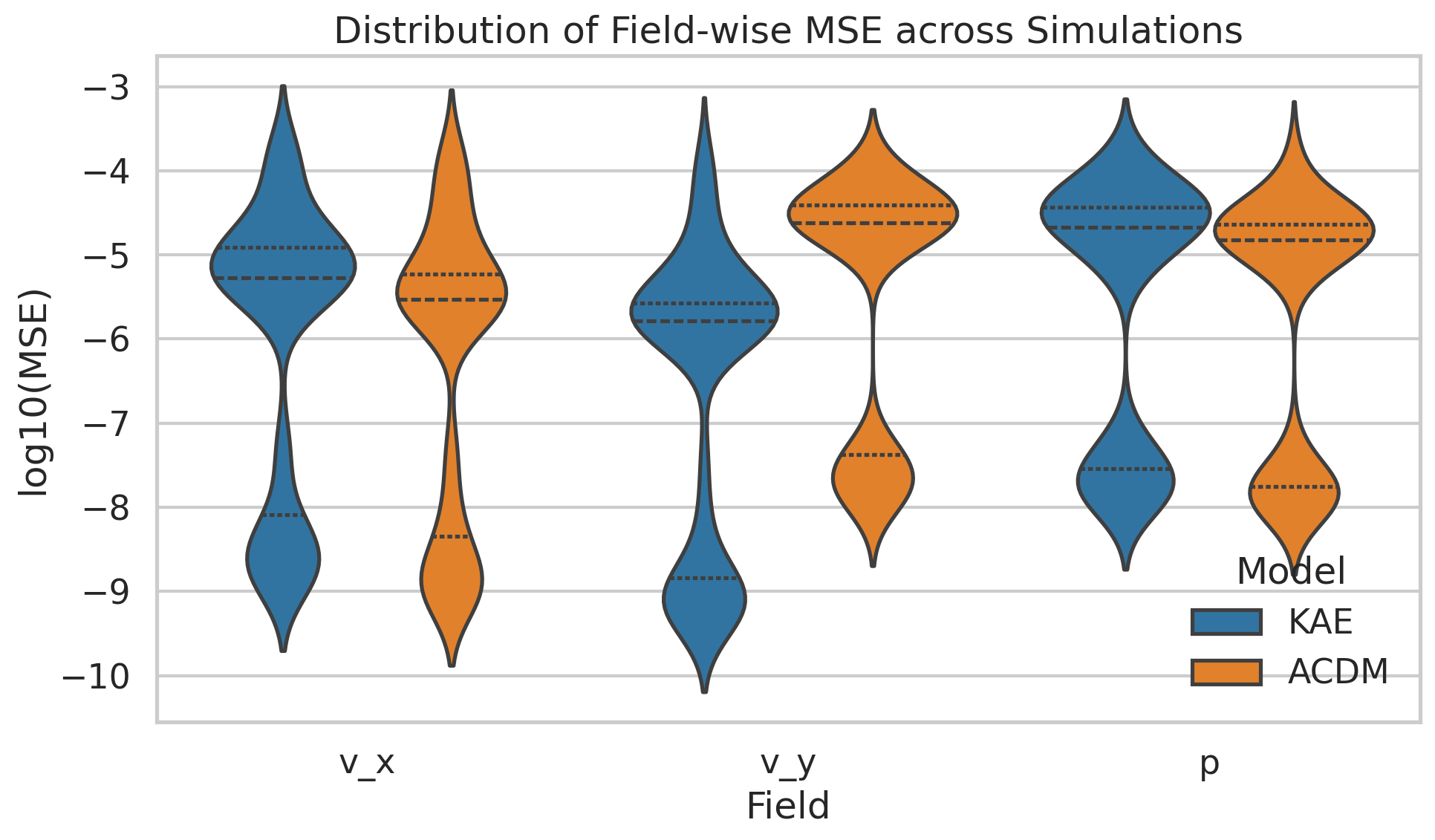}
        \caption{High Reynolds number}
    \end{subfigure}

    \caption{
    Error distributions under low and high Reynolds number regimes.
    While both models perform comparably at low Reynolds numbers, ACDM exhibits pronounced heavy-tailed error distributions at high Reynolds numbers.
    In contrast, KAE maintains controlled variance, demonstrating superior robustness in turbulent regimes.
    }
    \label{fig:inc_low_high_violin}
\end{figure*}

\section{Extended Stability and Spectral Analysis}
\label{appendix:extended_stability}

To thoroughly address the accuracy-stability trade-offs between deterministic latent dynamics and stochastic diffusion, this section provides an extended analysis of the models' spectral biases and their behavior under extreme autoregressive rollouts.

\subsection{Spectral Bias and Frequency Smoothing}\label{app:spectra_bias}

\begin{figure}[ht!]
    \centering
    \includegraphics[width=0.98\textwidth]{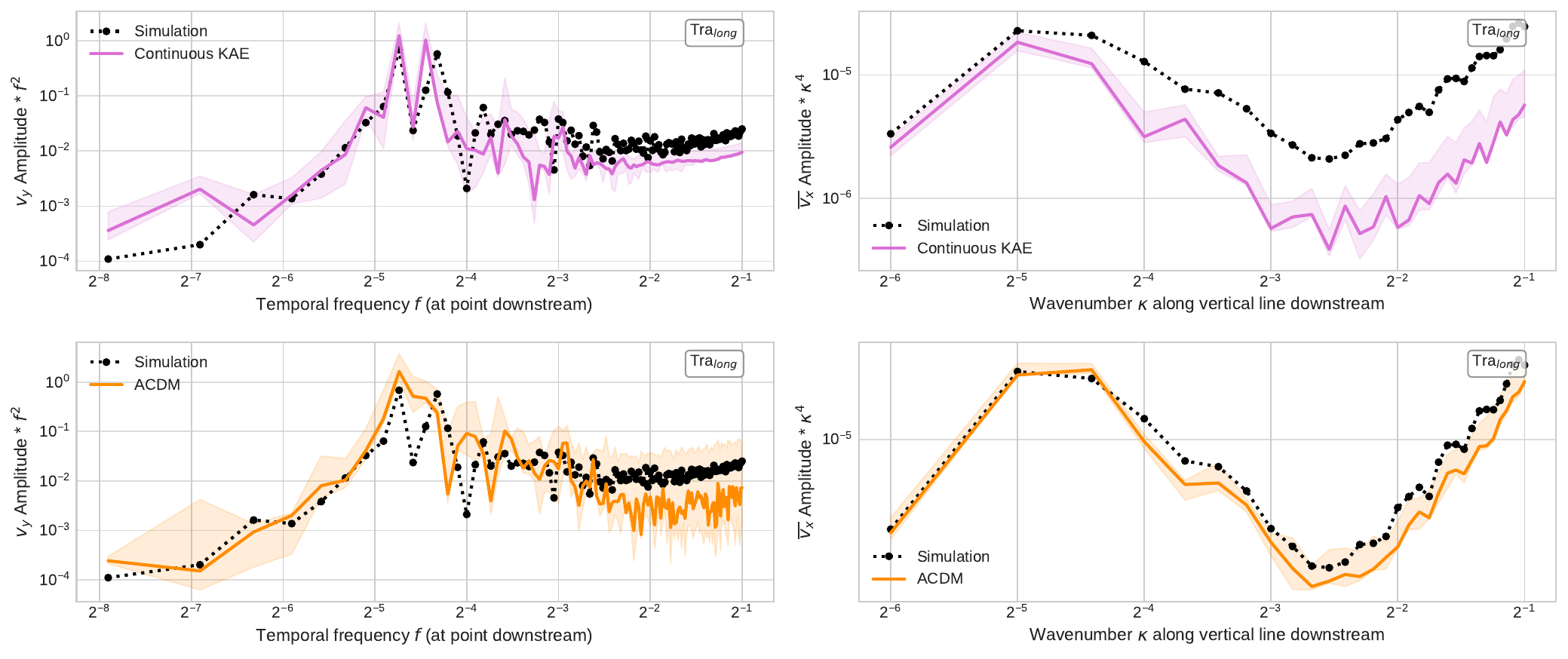}
    \caption{Temporal and spatial frequency analysis on a sequence from $Tra_{long}$. The shaded area represents the 5th to 95th percentile. KAE successfully captures the dominant physical frequencies but attenuates high-frequency turbulent noise compared to the stochastic baseline.}
    \label{fig:spectral_analysis}
\end{figure}

As established in turbulence research \citep{lorenz1969predictability}, even highly accurate numerical solvers eventually decorrelate from a target simulation over extended timeframes due to the chaotic nature of fluid dynamics. Therefore, we evaluate the temporal and spatial frequency spectra to measure whether the predicted trajectories statistically match the physical characteristics of the reference simulation. Figure \ref{fig:spectral_analysis} presents the temporal frequency of the vertical motion ($v_y$) and the spatial wavenumber of the horizontal motion ($v_x$) evaluated downstream.

The temporal analysis (Figure \ref{fig:spectral_analysis}, left) demonstrates that KAE accurately identifies and locks onto the primary vortex shedding frequency (the dominant energy peak), exhibiting a very tight variance band. ACDM also captures this primary frequency, though its stochastic formulation introduces slightly more variance. 

In the spatial domain (Figure \ref{fig:spectral_analysis}, right), ACDM accurately synthesizes the high-frequency turbulent textures. Conversely, KAE exhibits a steeper energy drop-off at high wavenumbers. This confirms that our model suppresses high-frequency chaotic structures in favor of a stable limit cycle. It acts as a low-pass physical filter, discarding the chaotic, unpredictable high-frequency turbulent cascade to guarantee that the primary, macro-scale shedding frequencies are preserved with near-zero variance. 

\subsection{Extreme Long-Horizon Stability ($T=1000$)} \label{appendix:long_rollout}
To test the absolute limits of the learned latent dynamics and investigate potential numerical divergence, we subjected both models to an extreme 1000-step autoregressive rollout (as visualized in Figure \ref{fig:very_long_rollout}). 

\begin{figure}[ht!]
    \centering
    \begin{subfigure}[b]{0.48\textwidth}
        \centering
        \includegraphics[width=\textwidth]{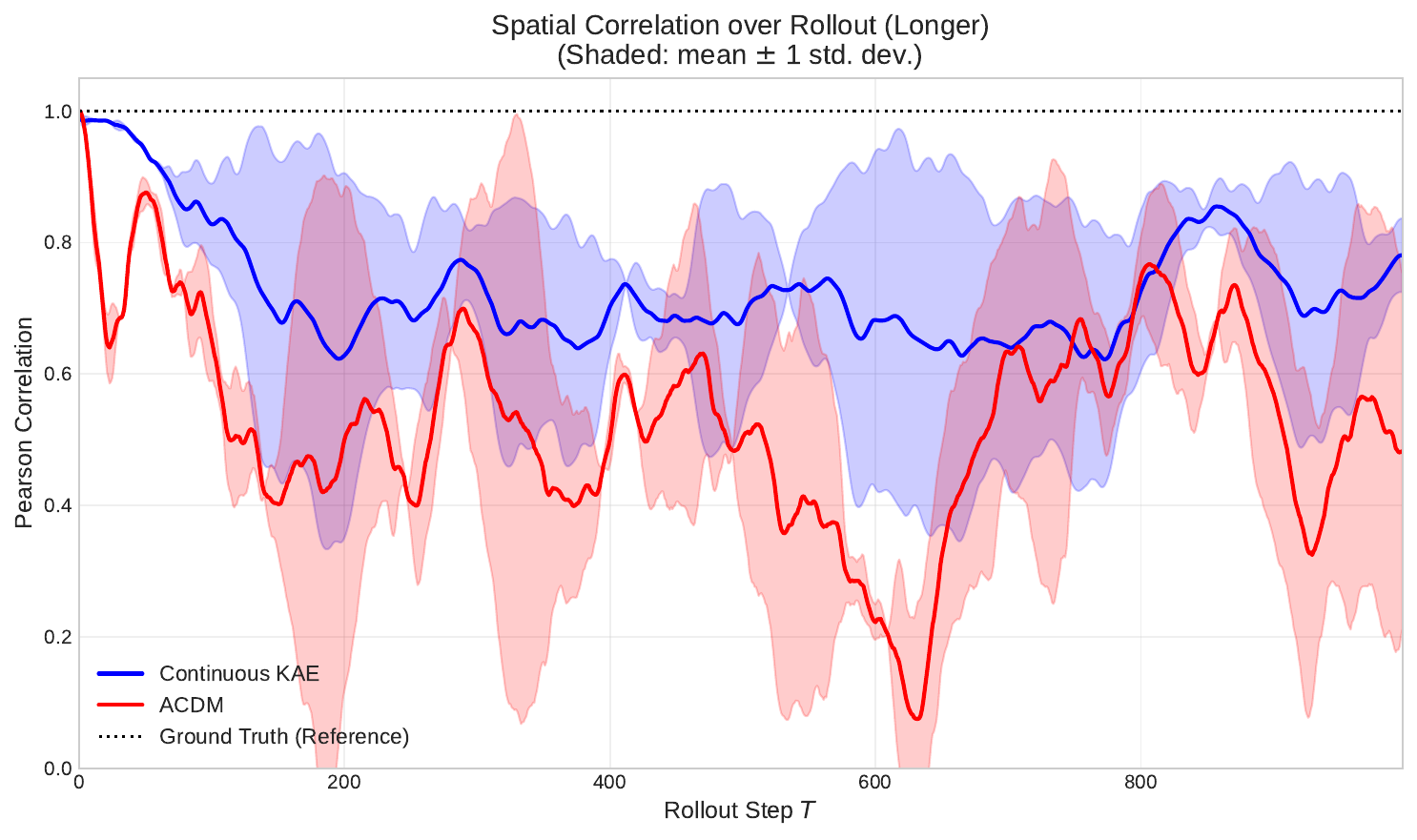}
        \caption{Spatial Correlation (1000 steps)}
    \end{subfigure}
    \hfill
    \begin{subfigure}[b]{0.48\textwidth}
        \centering
        \includegraphics[width=\textwidth]{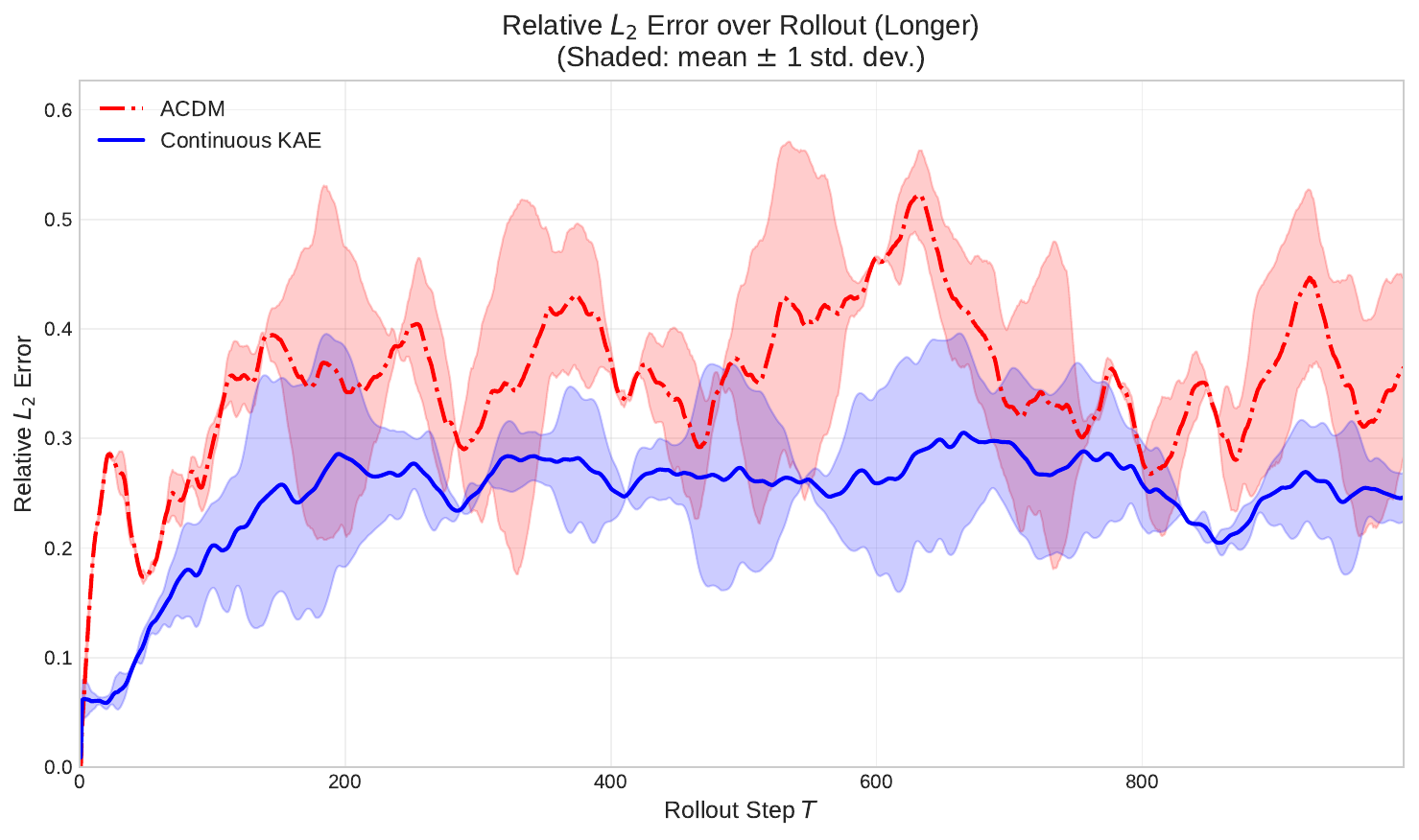}
        \caption{Relative $L_2$ Error (1000 steps)}
    \end{subfigure}
    \caption{Quantitative metrics over an extreme 1000-step rollout. ACDM exhibits severe instability and variance, while KAE remains strictly bounded by its linear latent dynamics.}
    \label{fig:1000_metrics}
\end{figure}

As shown in Figure \ref{fig:1000_metrics}, the danger of ACDM's textural hallucinations becomes catastrophic over extreme horizons. The stochastic nature of the diffusion model leads to severe phase divergence: its relative $L_2$ error spikes erratically with massive variance, and its spatial correlation completely collapses as the high-frequency details compound into unphysical numerical noise.

In contrast, KAE degrades gracefully. Because the latent space is governed by a globally stable linear Koopman operator, the system is asymptotically bounded. The transient, high-frequency modes naturally decay, leaving only the stable base flow. While fine-scale textures diffuse out over time, the model maintains a bounded $L_2$ error and a highly stable periodic correlation with the reference simulation, proving its reliability for robust long-term forecasting without structural collapse.

\subsection{Latent Dynamics Eigenvalue Spectrum}
\label{appendix:eigenvalues}

To understand the long-horizon stability of the continuous-time Koopman Autoencoder, we analyze the spectral properties of the learned latent dynamics. For a linear continuous-time system $dz/dt = \mathbf{K}z$, stability requires that the real parts of all eigenvalues of $\mathbf{K}$ are negative.

\begin{figure}[ht!]
    \centering
    \includegraphics[width=0.4\textwidth]{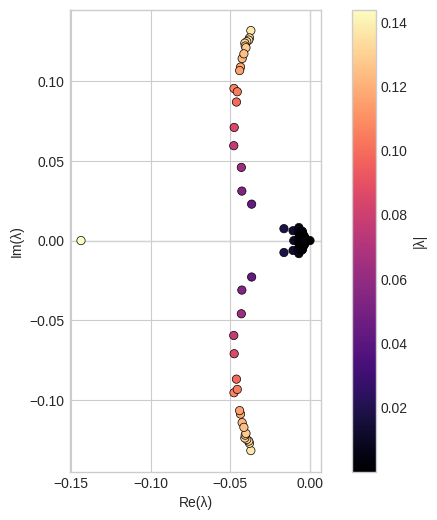}
    \caption{Eigenvalue spectrum of the learned Koopman generator across several conditioning parameters. The majority of eigenvalues lie in the stable region ($Re(\lambda) < 0$), indicating dissipative latent dynamics.}
    \label{fig:eig_spectrum}
\end{figure}

As shown in Figure~\ref{fig:eig_spectrum}, we plot the eigenvalue distribution of the learned generator $\mathbf{K}_{\mathrm{cont}}(\phi)$ across several conditioning parameters. The majority of eigenvalues lie in the left half of the complex plane ($\mathrm{Re}(\lambda) < 0$), reflecting a dissipative inductive bias that emerges from rollout training rather than being enforced as a hard constraint. This empirically stable spectrum suppresses high-frequency unstable growth modes, so numerical errors introduced during latent projection tend to decay over time rather than compound exponentially, producing bounded predictions over long horizons.

\section{Spatial Error Distribution and Difference Maps}
\label{appendix:difference_maps}

To analyze the physical nature of the prediction errors, we provide absolute spatial difference maps across all evaluated regimes. While raw MSE metrics often favor stochastic models in chaotic flows, these visual distributions highlight the trade-off between textural detail and structural coherence.

\begin{figure}[t]
    \centering
    \resizebox{0.5\textwidth}{!}{%
    \begin{subfigure}[b]{0.48\textwidth}
        \centering
        \includegraphics[width=\textwidth]{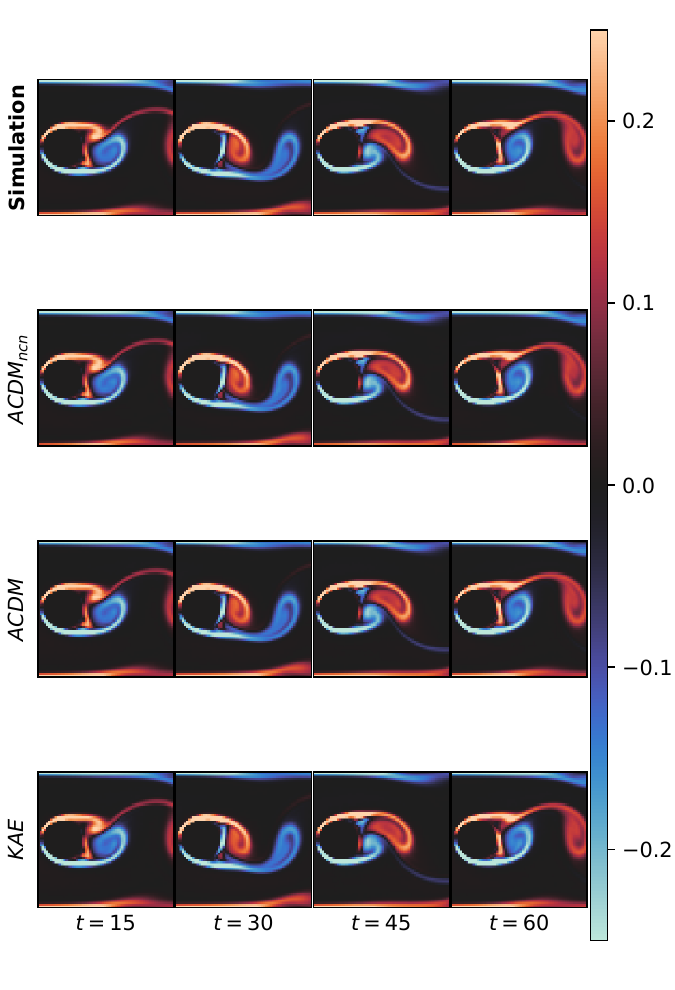}
        \caption{Vorticity for $Inc_{high}$ with $Re = 1000$}
        \label{fig:inc_high}
    \end{subfigure}
    \hfill
    \begin{subfigure}[b]{0.48\textwidth}
        \centering
        \includegraphics[width=\textwidth]{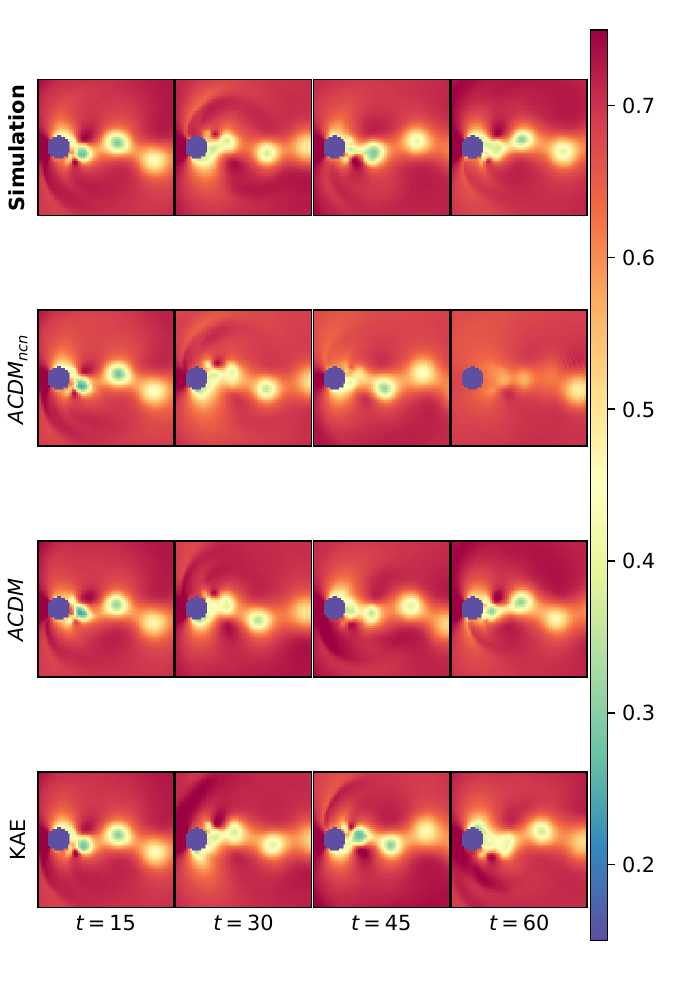}
        \caption{Pressure for $Tra_{ext}$ with $Ma = 0.50$}
        \label{fig:tra_extrap}
    \end{subfigure}
    }
    \caption{Qualitative validation rollouts for extrapolation regimes. (Left) Vorticity prediction for the incompressible wake flow in the high-Reynolds-number regime ($Inc_{high}, Re=1000$). (Right) Pressure prediction for the transonic cylinder flow in the low-Mach extrapolation regime ($Tra_{ext}, Ma=0.50$).
    }
    \label{fig:results_inc_high_tra_long}
\end{figure}

\subsection{Comparative Analysis: KAE vs. Diffusion Models}
The fundamental difference between KAE and the ACDM diffusion model lies in how they handle high-frequency information and temporal evolution:

\begin{itemize}
    \item \textbf{KAE (Deterministic Stability):} KAE is constrained by a linear Koopman operator in a continuous latent space. This structure ensures asymptotic stability, meaning errors do not grow exponentially over time. However, the deterministic $L_2$ objective leads to a "spectral bias" where the model prefers to smooth out high-frequency spatial discontinuities to ensure global structural alignment.
    \item \textbf{Diffusion Models (Stochastic Detail):} ACDM uses a stochastic process to synthesize fine-scale turbulent textures. While this allows the model to match the sharp 'look' of the ground truth, it lacks the rigid mathematical constraints of KAE. Consequently, minor stochastic errors compound over time, leading to widespread spatial noise and phase divergence.
\end{itemize}

\subsection{Visual Error Localization across All Regimes}
The difference maps reveal two distinct archetypes of model failure. Errors generated by KAE are highly localized along sharp discontinuities such as shock fronts and vortex edges. In contrast, errors from the ACDM baseline are more diffusely distributed throughout the spatial domain.

\begin{figure}[ht!]
    \centering
    \begin{subfigure}[b]{0.48\textwidth}
        \includegraphics[width=\textwidth]{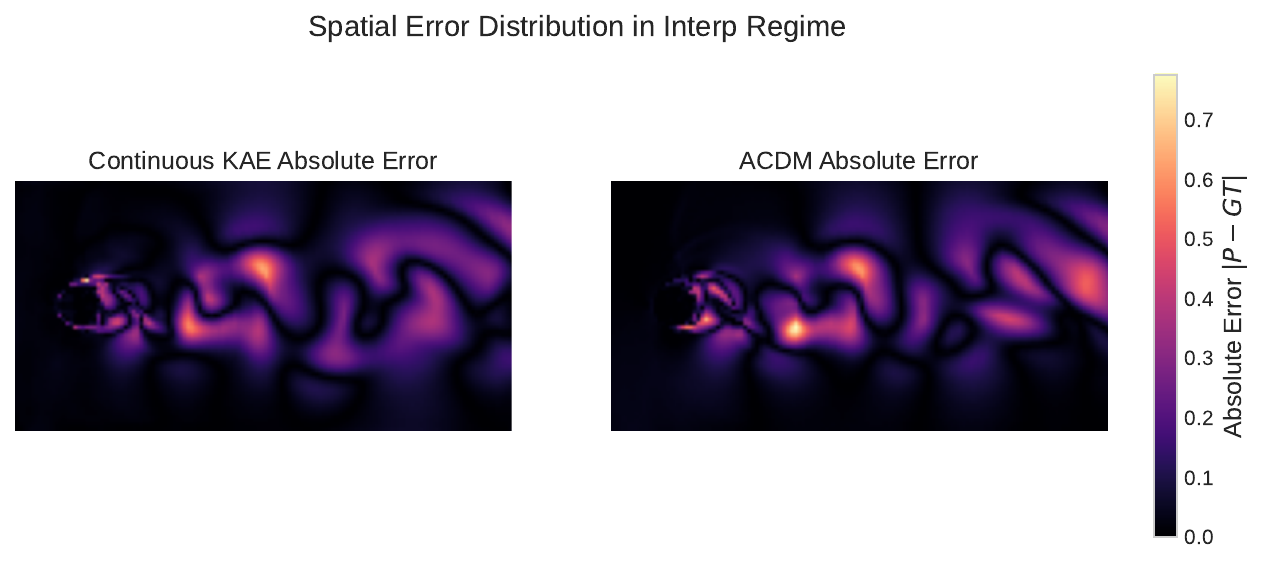}
        \caption{Transonic Interpolation}
    \end{subfigure}
    \hfill
    \begin{subfigure}[b]{0.48\textwidth}
        \includegraphics[width=\textwidth]{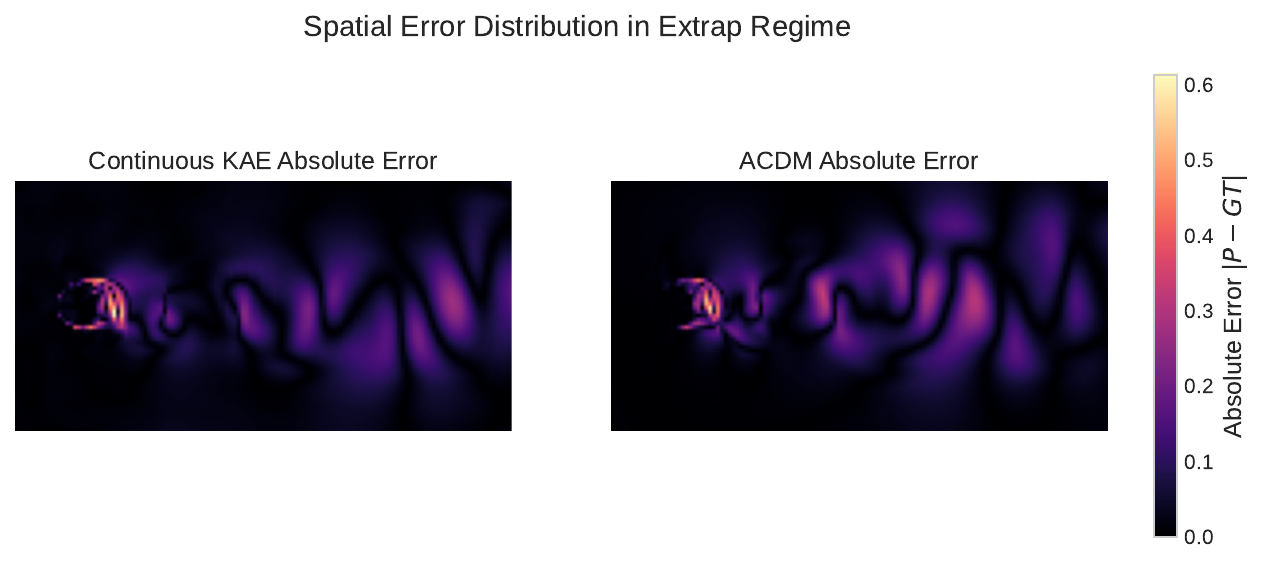}
        \caption{Transonic Extrapolation}
    \end{subfigure}
    \caption{Spatial error distribution in Transonic regimes. KAE errors are concentrated precisely at the sharp shock fronts, while ACDM exhibits broader spatial noise.}
    \label{fig:diff_maps_transonic_appx}
\end{figure}

\begin{figure}[ht!]
    \centering
    \includegraphics[width=0.9\textwidth]{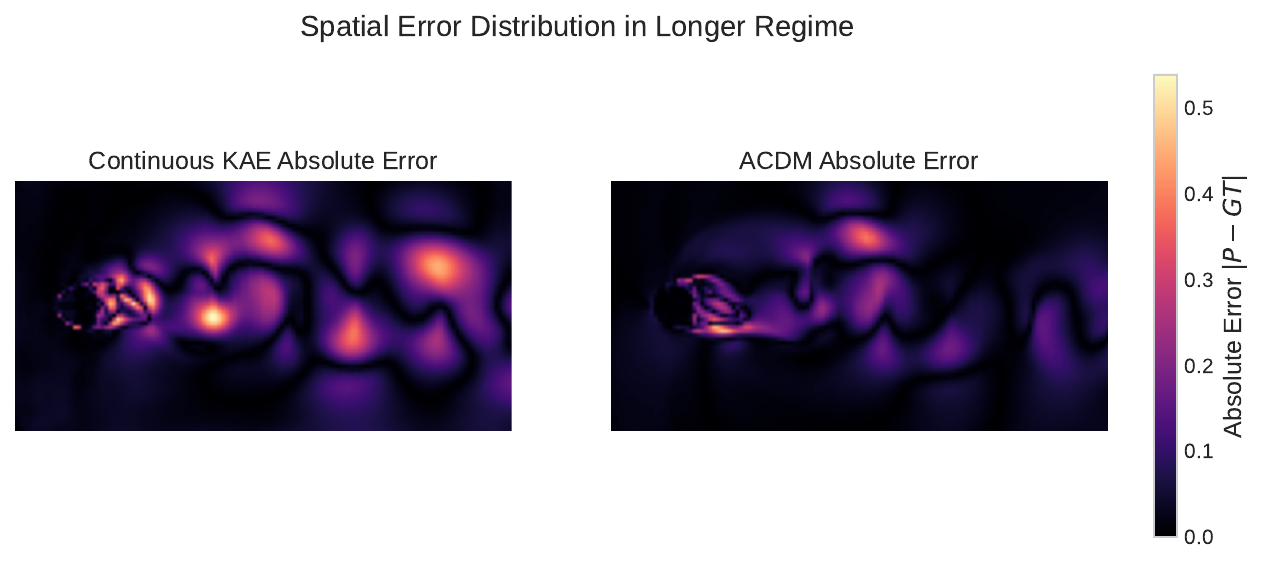}
    \caption{Spatial error distribution in the long-rollout regime. KAE maintains structural stability with localized errors, while ACDM shows diffuse error growth across the wake.}
    \label{fig:diff_maps_longer_appx}
\end{figure}

\begin{figure}[ht!]
    \centering
    \begin{subfigure}[b]{0.48\textwidth}
        \includegraphics[width=\textwidth]{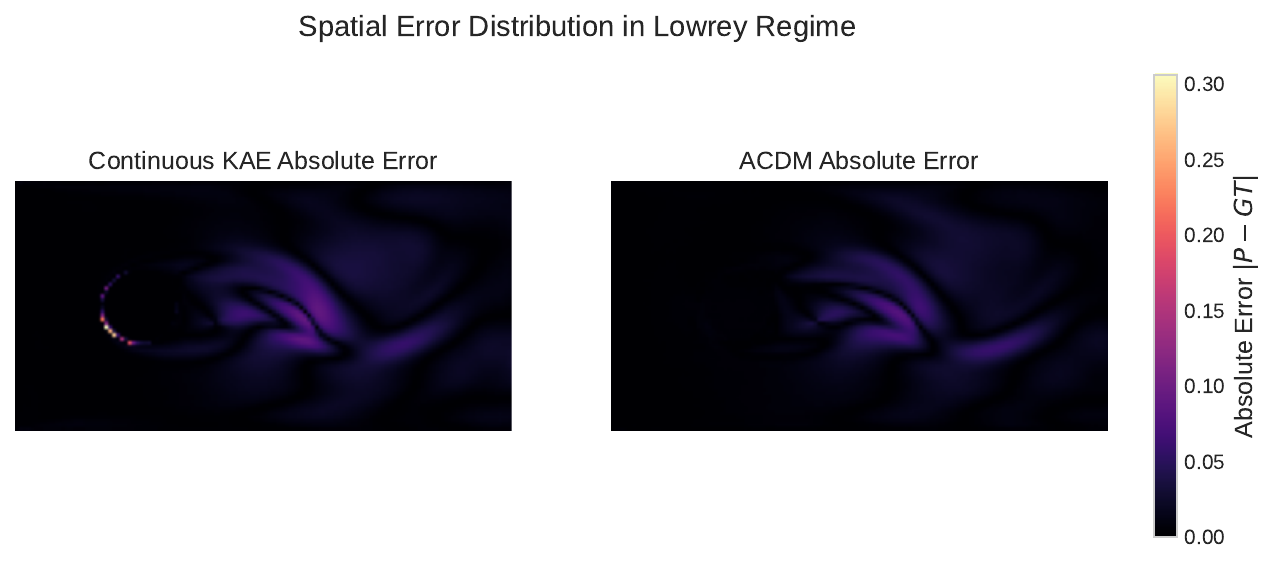}
        \caption{Incompressible (Low Reynolds)}
    \end{subfigure}
    \hfill
    \begin{subfigure}[b]{0.48\textwidth}
        \includegraphics[width=\textwidth]{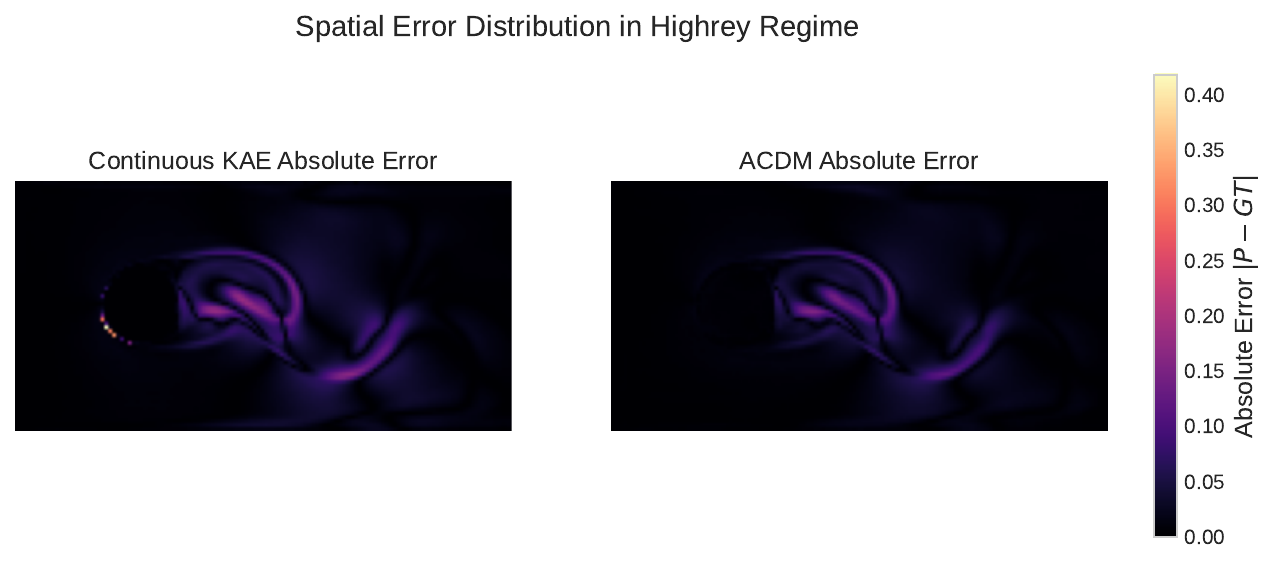}
        \caption{Incompressible (High Reynolds)}
    \end{subfigure}
    \caption{Comparison of error magnitudes across incompressible flow regimes. Note the higher error localization on the obstacle boundary for KAE.}
    \label{fig:diff_maps_incompressible_appx}
\end{figure}

\section{Data assimilation training and evaluation protocols}\label{app:da_protocol}

\subsection{Algorithms used}\label{app:da_algs}
 
\providecommand{\Kop}{\mathbf{K}}

\algrenewcommand{\algorithmicrequire}{\textbf{Input:}}
\algrenewcommand{\algorithmicensure}{\textbf{Output:}}

\begin{algorithm}[t]
\caption{KAE 4D-Var. All weights frozen; the control is the latent alone. The propagators
are built once, before the loop, because $\Kop$ does not change during it.}
\label{alg:kae}
\begin{algorithmic}[1]
\Require observations $\{y_i\}_{i=1}^{N}$, masks $\{M_i\}$, leads $\{\tau_i\}$; frozen
         $\mathcal{E},\mathcal{D},\Kop$; step size $\eta$; budget $T$
\Ensure  $\hat u(t_0)$
\State $\Phi_i \gets \exp(\Kop\,\tau_i)$ for $i=1,\dots,N$
\If{warm start}
  \State $\tilde y \gets y_1 + \text{null history}$
  \State $z \gets \exp(-\Kop\,\tau_\star)\,\mathcal{E}(\tilde y)$
\Else
  \State $z \gets \mathbf{0}$
\EndIf
\For{$t = 1,\dots,T$}
  \State $J \gets \tfrac{1}{N}\sum_{i=1}^{N}
          \big\|\,M_i \odot (\mathcal{D}(\Phi_i z) - y_i)\,\big\|^2$
  \State $z \gets \mathrm{Adam}(z,\;\nabla_z J,\;\eta)$
\EndFor
\State \Return $\mathcal{D}(z)$
\end{algorithmic}
\end{algorithm}

\begin{algorithm}[t]
\caption{Autoregressive 4D-Var (U-Net, FNO). The control is the $k$-frame window ending at
$t_0$. The rollout is rebuilt inside every iteration, and the leads $n_i$ must be integers.}
\label{alg:ar}
\begin{algorithmic}[1]
\Require observations $\{y_i\}_{i=1}^{N}$, masks $\{M_i\}$, integer leads $\{n_i\}$;
         frozen $F_\theta$; window $k$; step size $\eta$; budget $T$
\Ensure  $\hat u(t_0)$
\State $c \gets \mathbf{0} \in \mathbb{R}^{k\times C\times H\times W}$
\For{$t = 1,\dots,T$}
  \State $x_{1:k} \gets c$
  \For{$s = k,\dots,k+\max_i n_i-1$}
    \State $x_{s+1} \gets F_\theta(x_{s-k+1:s})$
  \EndFor
  \State $J \gets \tfrac{1}{N}\sum_{i=1}^{N}
          \big\|\,M_i \odot (x_{k+n_i} - y_i)\,\big\|^2$
  \State $c \gets \mathrm{Adam}(c,\;\nabla_c J,\;\eta)$
\EndFor
\State \Return $c_k$
\end{algorithmic}
\end{algorithm}

\subsection{Kuramoto--Sivashinsky}\label{app:da_ks}

\paragraph{Data.} $u_t = -uu_x-u_{xx}-u_{xxxx}$ on a periodic domain, of $64$ points, stored every $\Delta t = 0.1$. Training uses $100$ trajectories of $1000$ frames
split $80/10/10$. Assimilation uses a separate set of $64$ trajectories generated from the same
simulator configuration, plus $32$ trajectories of $5000$ frames for post-assimilation forecasts. There is no overlap between these and the training and validation trajectories. Lyapunov time is $T_L = 22.6$ time units, measured on $12$ held-out trajectories.

\paragraph{Training.} KAE trains for $200$ epochs, batch $1024$, Adam at $10^{-3}$ with $20$
epochs of warm-up and linear decay to $2\times10^{-5}$, on a $10$-frame rollout
(best validation loss $1.473\times10^{-3}$). The U-Net baseline uses the same data, settings, objective, rollout and optimiser schedule. The score network trains for $128$ epochs, batch $128$, Adam at
$2\times10^{-4}$ with weight decay $10^{-3}$, on segments of nine frames under the cosine VP
schedule. All three see the same exact training split.

\paragraph{Models.} \emph{KAE.} Because $\mathbf{K}_{cont}$ is frozen, the
$N$ matrices $e^{\mathbf{K}_{cont}\tau_i}$ are computed once before the optimisation loop, each iteration then
costs $N$ matrix--vector products and $N$ decoder evaluations, with no intermediate integration
state stored or differentiated through. Observation times do not need to lie on the training grid.
As an optimiser, we use Adam at $10^{-2}$, $8000$ iterations.

\emph{U-Net.} A $1$-D autoregressive map $F_\theta$ advancing the state by $\Delta t$. The
target is the $64$-point physical state $x_0$ and $\mathcal{H}_{x_0}(\tau)=F_\theta^{\tau/\Delta t}(x_0)$,
so the rollout sits inside every iteration and must be backpropagated. We use the same optimiser, Adam at $10^{-2}$, $8000$ iterations.

\emph{SDA.} A score network trained here on nine-frame segments
($k=4$), with Algorithm~2 of \citet{rozet2023score} composing the local scores into a score over
the whole trajectory. Observations use a Gaussian likelihood to return the Tweedie estimate
of the clean trajectory. Sampling uses $128$ exponential-integrator steps with four Langevin
corrections ($\tau=0.5$) and $\sigma_y=0.05$ for clean observations. There is no control variable
and no iteration count, so the reported value is independent of the optimization budget.

\begin{figure}[ht]
    \centering
    \includegraphics[width=1.0\linewidth]{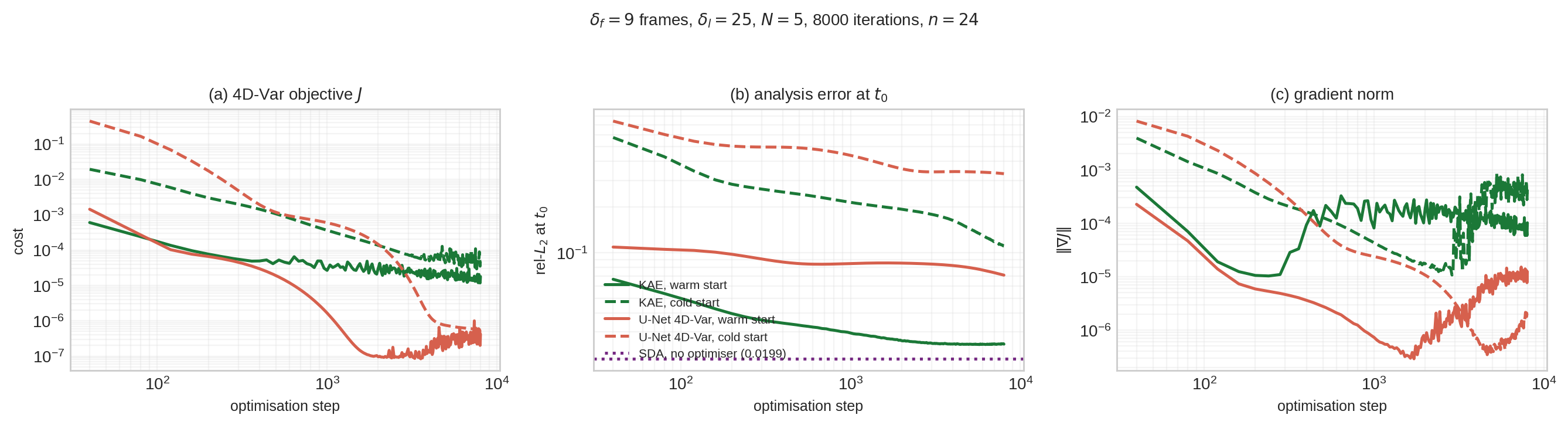}
    \caption{Convergence, analysis and gradient visualization for KAE and UNet 4D-Var optimization process for a cold and warm start in the optimization process, across 8000 steps.}
    \label{fig:ks_convergence_df9}
\end{figure}

\subsection{Transonic cylinder flow}\label{app:da:tra}
\paragraph{Data.} Compressible K\'arm\'an vortex streets past a cylinder at $Re = 10^4$,
$128\times64$, four fields ($\rho$, $p$, $v_x$, $v_y$) plus an obstacle mask, from
\citet{kohl2026benchmarking}. The raw interval $\Delta t = 0.05$\,s is subsampled to $0.1$\,s.
Training covers $Ma\in[0.53,0.63]\cup[0.69,0.89]$ ($32$ sequences of $1000$ frames). Assimilation
is run on $Tra_{int}$ ($Ma\in\{0.66,0.67,0.68\}$, $6$ sequences of $60$ frames), which sits inside the
gap in the training range and was seen by no model.

\paragraph{Training.} KAE is trained on the transonic training set with a latent space of size $128$. The four baselines are the pretrained checkpoints (on the same data splits) of \citet{kohl2026benchmarking}, unmodified: U-Net, FNO with $32$ modes, and ACDM and
ACDM$_{\mathrm{ncn}}$.

\paragraph{Models.} \emph{KAE.} As on KS, with two differences. Two conditioning frames only are encoded
separately and aggregated by a Transformer encoder with sinusoidal positional encodings, matching
the AR-2 conditioning of the diffusion baseline. The operator is conditioned on the Mach number through a LoRA update \citep{hu2022lora}, so $\mathbf{K}_{cont}$ is fixed per problem but differs between problems. The target is again $z_0\in\mathbb{R}^{128}$. For this, we use Adam at $10^{-1}$, $2000$ iterations.

\emph{U-Net and FNO 4D-Var.} Both are autoregressive maps, where the target is in the model's own normalised space and propagation is $\tau/\Delta t$ passes of the network inside each iteration. We use a tuned Adam at $3\times10^{-2}$, $2000$ iterations.

\emph{ACDM $+$ SDA.} ACDM is not run autoregressively here. Its network predicts $\epsilon$ for all
three frames it sees (two conditioning frames and the target), and all three outputs are kept as a
local score over a three-frame window, composed over $t_0,\dots,t_0+\dl$ by the same algorithm used
on KS. The likelihood covariance is $\sigma_y^2 + (\sigma^2/\mu^2)\Gamma$ with $\Gamma=0.3$ and
$\sigma_y = \max(\sigma,0.05)$. There are three deviations as compared to the sampler used on KS: a three-frame blanket instead of five, $20$ denoising steps instead of $256$, and ACDM's own DDPM schedule rather than an exponential integrator. These were chosed for consistency with the previous models and optimized for ACDM's best DA results.

\emph{ACDM$_{\mathrm{ncn}}$.} The same architecture trained with clean conditioning, so its outputs
on the conditioning frames are not a score and the composition is ill-posed. We report it once, at
$2.4$ analysis error, and exclude it from every sweep.

\section{Data Assimilation Ablations}\label{app:da_ablations}

\begin{table}[H]
\centering\scriptsize\setlength{\tabcolsep}{4pt}\renewcommand{\arraystretch}{1.35}
\begin{tabular}{lrrrr}
\toprule
TRA sweep ($\df=9$), first $\to$ last & KAE & ACDM & FNO & U-Net \\
\midrule
\shortstack[l]{$\dl$: $13\to50$ frames\\\strut} & \shortstack[r]{$0.094\!\pm\!0.005$\\$\to$ $0.105\!\pm\!0.006$} & \shortstack[r]{$0.248\!\pm\!0.005$\\$\to$ $0.254\!\pm\!0.004$} & \shortstack[r]{$0.294\!\pm\!0.002$\\$\to$ $0.313\!\pm\!0.010$} & \shortstack[r]{$0.241\!\pm\!0.002$\\$\to$ $0.272\!\pm\!0.004$} \\
\shortstack[l]{$N$: $2\to17$\\\strut} & \shortstack[r]{$0.093\!\pm\!0.011$\\$\to$ $0.090\!\pm\!0.003$} & \shortstack[r]{$0.253\!\pm\!0.005$\\$\to$ $0.255\!\pm\!0.006$} & \shortstack[r]{$0.273\!\pm\!0.002$\\$\to$ $0.304\!\pm\!0.003$} & \shortstack[r]{$0.266\!\pm\!0.002$\\$\to$ $0.281\!\pm\!0.005$} \\
\shortstack[l]{obs.\ fraction: $1\to0.005$\\\strut} & \shortstack[r]{$0.084\!\pm\!0.008$\\$\to$ $0.123\!\pm\!0.006$} & \shortstack[r]{$0.253\!\pm\!0.005$\\$\to$ $0.261\!\pm\!0.007$} & \shortstack[r]{$0.293\!\pm\!0.001$\\$\to$ $0.261\!\pm\!0.005$} & \shortstack[r]{$0.275\!\pm\!0.004$\\$\to$ $0.267\!\pm\!0.003$} \\
\shortstack[l]{noise $\sigma$: $0\to0.3$\\\strut} & \shortstack[r]{$0.084\!\pm\!0.008$\\$\to$ $0.091\!\pm\!0.006$} & \shortstack[r]{$0.253\!\pm\!0.005$\\$\to$ $0.269\!\pm\!0.004$} & \shortstack[r]{$0.293\!\pm\!0.001$\\$\to$ $0.329\!\pm\!0.005$} & \shortstack[r]{$0.275\!\pm\!0.004$\\$\to$ $0.272\!\pm\!0.005$} \\
\bottomrule
\end{tabular}
\caption{TRA, the same sweeps with $\df$ pinned at 9 frames; first and last point of
each sweep, mean $\pm$ standard error.}
\label{tab:tra_sweeps_df9}
\end{table}

\begin{figure}
    \centering
    \includegraphics[width=0.8\linewidth]{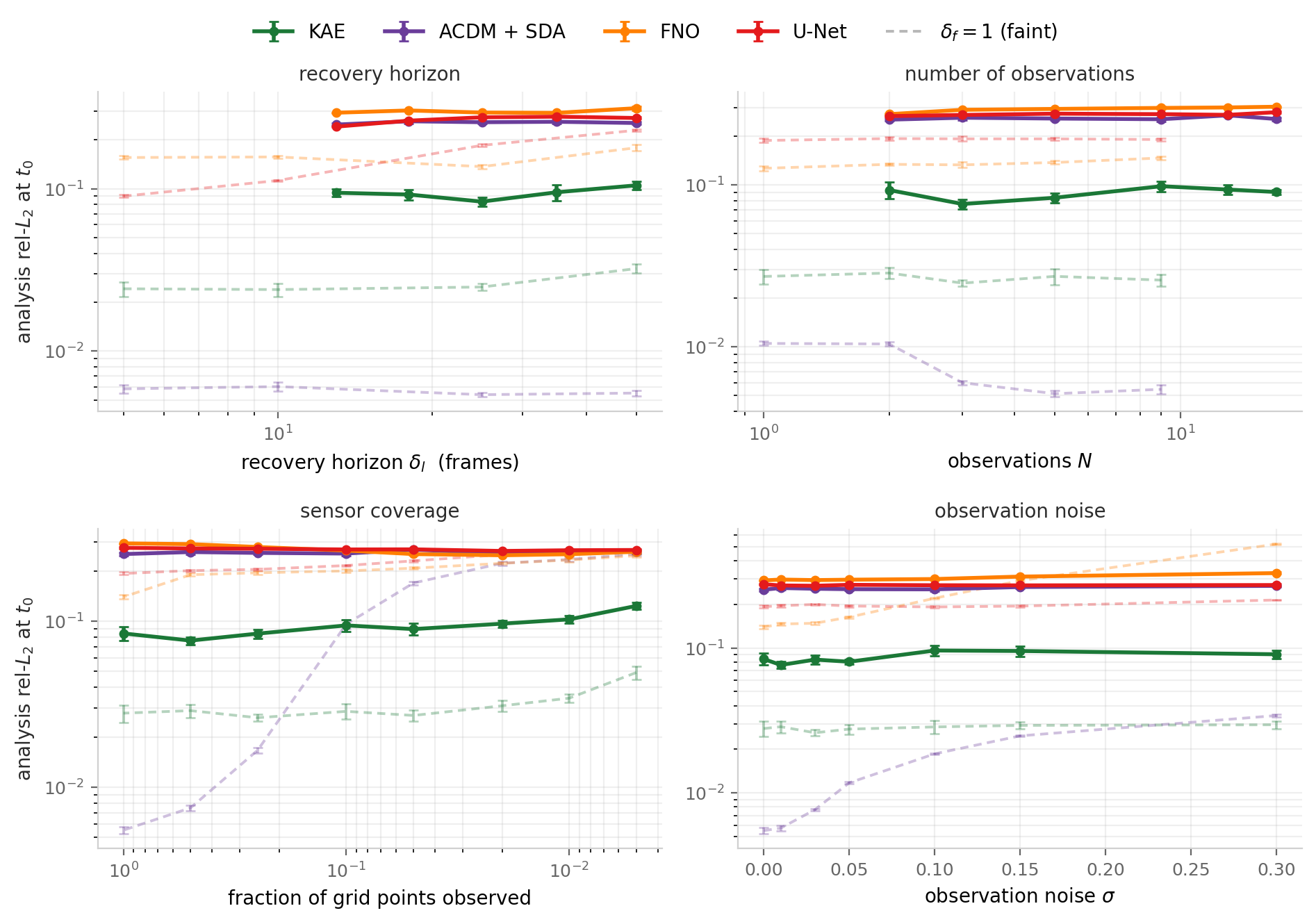}
    \caption{Tra Sweeps at $\df=1$ and $\df=9$.}
    \label{fig:tra_sweeps}
\end{figure}

\begin{table}[H]
\centering
\begin{minipage}[t]{0.49\linewidth}
\centering\scriptsize\setlength{\tabcolsep}{2.6pt}
$\df=1$\\[2pt]
\begin{tabular}{lrrr}
\toprule
$\dl$ (t.u.) & KAE & SDA & U-Net \\
\midrule
0.5 & $0.0027\!\pm\!0.0001$ & $0.0079\!\pm\!0.0003$ & $0.1877\!\pm\!0.0046$ \\
1 & $0.0027\!\pm\!0.0001$ & $0.0074\!\pm\!0.0002$ & $0.1911\!\pm\!0.0052$ \\
2.5 & $0.0033\!\pm\!0.0002$ & $0.0088\!\pm\!0.0003$ & $0.1924\!\pm\!0.0038$ \\
5 & $0.0040\!\pm\!0.0002$ & $0.0099\!\pm\!0.0003$ & $0.1886\!\pm\!0.0057$ \\
10 & $0.0059\!\pm\!0.0004$ & $0.0106\!\pm\!0.0004$ & $0.1593\!\pm\!0.0089$ \\
20 & $0.0068\!\pm\!0.0004$ & $0.0129\!\pm\!0.0006$ & $0.1688\!\pm\!0.0125$ \\
40 & $0.0082\!\pm\!0.0005$ & $0.0146\!\pm\!0.0008$ & $0.8355\!\pm\!0.0108$ \\
70 & $0.0096\!\pm\!0.0006$ & $0.0182\!\pm\!0.0013$ & $0.9930\!\pm\!0.0072$ \\
\bottomrule
\end{tabular}
\end{minipage}\hfill
\begin{minipage}[t]{0.49\linewidth}
\centering\scriptsize\setlength{\tabcolsep}{2.6pt}
$\df=9$\\[2pt]
\begin{tabular}{lrrr}
\toprule
$\dl$ (t.u.) & KAE & SDA & U-Net \\
\midrule
1.3 & $0.0234\!\pm\!0.0027$ & $0.0146\!\pm\!0.0007$ & $0.4230\!\pm\!0.0074$ \\
2.5 & $0.0191\!\pm\!0.0025$ & $0.0209\!\pm\!0.0013$ & $0.4576\!\pm\!0.0105$ \\
5 & $0.0244\!\pm\!0.0022$ & $0.0216\!\pm\!0.0014$ & $0.4452\!\pm\!0.0148$ \\
10 & $0.0294\!\pm\!0.0033$ & $0.0219\!\pm\!0.0013$ & $0.4291\!\pm\!0.0139$ \\
20 & $0.0424\!\pm\!0.0053$ & $0.0237\!\pm\!0.0018$ & $0.3565\!\pm\!0.0178$ \\
40 & $0.0583\!\pm\!0.0077$ & $0.0243\!\pm\!0.0016$ & $0.8788\!\pm\!0.0127$ \\
70 & $0.0766\!\pm\!0.0045$ & $0.0278\!\pm\!0.0023$ & $0.9900\!\pm\!0.0077$ \\
\bottomrule
\end{tabular}
\end{minipage}
\caption{KS, recovery horizon $\dl$. Mean $\pm$ standard error over the problems at each point, everything else pinned.}
\label{tab:ks_C2_delta_l}
\end{table}

\begin{table}[H]
\centering
\begin{minipage}[t]{0.49\linewidth}
\centering\scriptsize\setlength{\tabcolsep}{2.6pt}
$\df=1$\\[2pt]
\begin{tabular}{lrrr}
\toprule
$N$ & KAE & SDA & U-Net \\
\midrule
2 & $0.0089\!\pm\!0.0005$ & $0.0181\!\pm\!0.0008$ & $0.0873\!\pm\!0.0059$ \\
3 & $0.0026\!\pm\!0.0001$ & $0.0133\!\pm\!0.0006$ & $0.0931\!\pm\!0.0056$ \\
5 & $0.0022\!\pm\!0.0001$ & $0.0091\!\pm\!0.0003$ & $0.0965\!\pm\!0.0057$ \\
8 & $0.0022\!\pm\!0.0001$ & $0.0074\!\pm\!0.0002$ & $0.0994\!\pm\!0.0058$ \\
\bottomrule
\end{tabular}
\end{minipage}\hfill
\begin{minipage}[t]{0.49\linewidth}
\centering\scriptsize\setlength{\tabcolsep}{2.6pt}
$\df=9$\\[2pt]
\begin{tabular}{lrrr}
\toprule
$N$ & KAE & SDA & U-Net \\
\midrule
2 & $0.0626\!\pm\!0.0039$ & $0.0229\!\pm\!0.0013$ & $0.3189\!\pm\!0.0071$ \\
3 & $0.0261\!\pm\!0.0023$ & $0.0225\!\pm\!0.0013$ & $0.3245\!\pm\!0.0076$ \\
5 & $0.0105\!\pm\!0.0009$ & $0.0215\!\pm\!0.0012$ & $0.3293\!\pm\!0.0081$ \\
9 & $0.0112\!\pm\!0.0010$ & $0.0190\!\pm\!0.0010$ & $0.3318\!\pm\!0.0084$ \\
13 & $0.0115\!\pm\!0.0010$ & $0.0184\!\pm\!0.0010$ & $0.3333\!\pm\!0.0086$ \\
\bottomrule
\end{tabular}
\end{minipage}
\caption{KS, number of observations $N$. Mean $\pm$ standard error over the problems at each point, everything else pinned.}
\label{tab:ks_C3_n_obs}
\end{table}

\begin{table}[H]
\centering
\begin{minipage}[t]{0.49\linewidth}
\centering\scriptsize\setlength{\tabcolsep}{2.6pt}
$\df=1$\\[2pt]
\begin{tabular}{lrrr}
\toprule
$\sigma$ & KAE & SDA & U-Net \\
\midrule
0.0 & $0.0024\!\pm\!0.0001$ & $0.0098\!\pm\!0.0003$ & $0.0960\!\pm\!0.0057$ \\
0.01 & $0.0065\!\pm\!0.0002$ & $0.0038\!\pm\!0.0001$ & $0.0953\!\pm\!0.0036$ \\
0.05 & $0.0354\!\pm\!0.0017$ & $0.0135\!\pm\!0.0005$ & $0.1748\!\pm\!0.0059$ \\
0.1 & $0.0740\!\pm\!0.0040$ & $0.0242\!\pm\!0.0010$ & $0.2587\!\pm\!0.0071$ \\
0.2 & $0.1419\!\pm\!0.0074$ & $0.0417\!\pm\!0.0018$ & $0.3780\!\pm\!0.0102$ \\
0.4 & $0.2777\!\pm\!0.0131$ & $0.0770\!\pm\!0.0039$ & $0.5831\!\pm\!0.0152$ \\
\bottomrule
\end{tabular}
\end{minipage}\hfill
\begin{minipage}[t]{0.49\linewidth}
\centering\scriptsize\setlength{\tabcolsep}{2.6pt}
$\df=9$\\[2pt]
\begin{tabular}{lrrr}
\toprule
$\sigma$ & KAE & SDA & U-Net \\
\midrule
0.0 & $0.0142\!\pm\!0.0016$ & $0.0214\!\pm\!0.0007$ & $0.3671\!\pm\!0.0121$ \\
0.01 & $0.0554\!\pm\!0.0019$ & $0.0120\!\pm\!0.0003$ & $0.3695\!\pm\!0.0118$ \\
0.05 & $0.2199\!\pm\!0.0096$ & $0.0257\!\pm\!0.0011$ & $0.4962\!\pm\!0.0180$ \\
0.1 & $0.3500\!\pm\!0.0186$ & $0.0389\!\pm\!0.0019$ & $0.7886\!\pm\!0.0429$ \\
0.2 & $0.4821\!\pm\!0.0279$ & $0.0617\!\pm\!0.0037$ & $1.5046\!\pm\!0.0902$ \\
0.4 & $0.6686\!\pm\!0.0379$ & $0.1027\!\pm\!0.0066$ & $2.8842\!\pm\!0.1582$ \\
\bottomrule
\end{tabular}
\end{minipage}
\caption{KS, observation noise $\sigma$ (Gaussian). Mean $\pm$ standard error over the problems at each point, everything else pinned.}
\label{tab:ks_noise_law}
\end{table}

\begin{table}[H]
\centering
\begin{minipage}[t]{0.49\linewidth}
\centering\scriptsize\setlength{\tabcolsep}{2.6pt}
$\df=1$\\[2pt]
\begin{tabular}{lrrr}
\toprule
sensors & KAE & SDA & U-Net \\
\midrule
100\% & $0.0022\!\pm\!0.0001$ & $0.0094\!\pm\!0.0002$ & $0.0952\!\pm\!0.0060$ \\
50\% & $0.0067\!\pm\!0.0005$ & $0.0147\!\pm\!0.0005$ & $0.1084\!\pm\!0.0053$ \\
25\% & $0.0348\!\pm\!0.0029$ & $0.0257\!\pm\!0.0014$ & $0.1690\!\pm\!0.0077$ \\
10\% & $0.5734\!\pm\!0.0248$ & $0.2213\!\pm\!0.0368$ & $1.0289\!\pm\!0.0246$ \\
\bottomrule
\end{tabular}
\end{minipage}\hfill
\begin{minipage}[t]{0.49\linewidth}
\centering\scriptsize\setlength{\tabcolsep}{2.6pt}
$\df=9$\\[2pt]
\begin{tabular}{lrrr}
\toprule
sensors & KAE & SDA & U-Net \\
\midrule
100\% & $0.0121\!\pm\!0.0015$ & $0.0215\!\pm\!0.0009$ & $0.3491\!\pm\!0.0100$ \\
50\% & $0.0251\!\pm\!0.0015$ & $0.0281\!\pm\!0.0015$ & $0.3683\!\pm\!0.0118$ \\
25\% & $0.0836\!\pm\!0.0059$ & $0.0378\!\pm\!0.0023$ & $0.4110\!\pm\!0.0128$ \\
10\% & $0.5752\!\pm\!0.0241$ & $0.2471\!\pm\!0.0409$ & $1.0176\!\pm\!0.0224$ \\
\bottomrule
\end{tabular}
\end{minipage}
\caption{KS, sensor coverage (at $N=5$). Mean $\pm$ standard error over the problems at each point, everything else pinned.}
\label{tab:ks_joint_sparsity}
\end{table}

\begin{figure}[ht]
    \centering
    \includegraphics[width=0.8\linewidth]{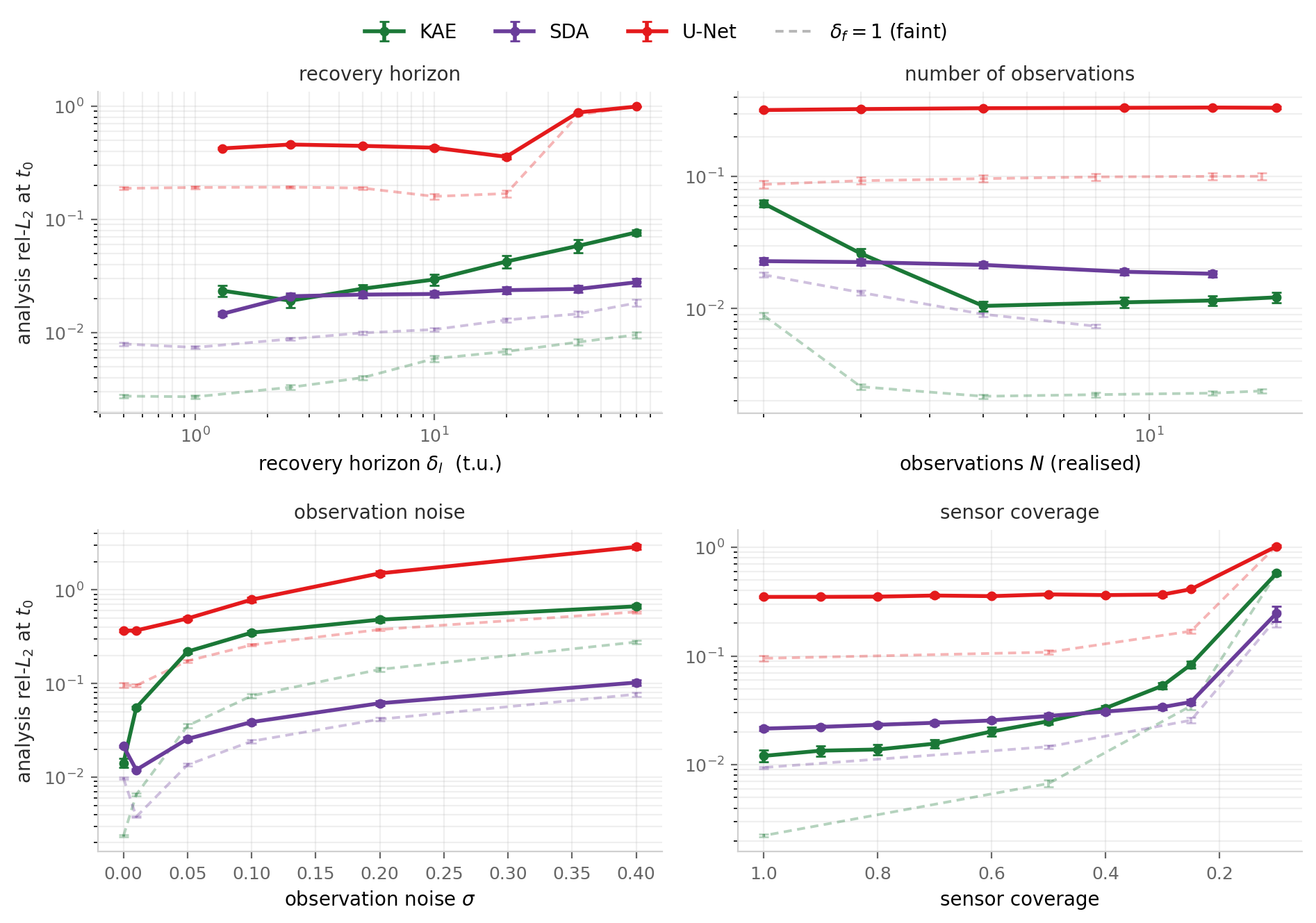}
    \caption{KS Sweeps at $\df=1$ and $\df=9$.}
    \label{fig:ks_sweeps}
\end{figure}

\begin{table}[H]
\centering\scriptsize\setlength{\tabcolsep}{3pt}
\begin{tabular}{lrrr}
\toprule
Sweep ($\df=9$) & KAE & ACDM & U-Net \\
\midrule
$\dl$ (frames) & $0.0944\!\pm\!0.0055$ $\to$ $0.1048\!\pm\!0.0060$ & $0.2481\!\pm\!0.0050$ $\to$ $0.2537\!\pm\!0.0040$ & $0.2409\!\pm\!0.0022$ $\to$ $0.2724\!\pm\!0.0037$ \\
$N$ & $0.0928\!\pm\!0.0109$ $\to$ $0.0902\!\pm\!0.0034$ & $0.2526\!\pm\!0.0053$ $\to$ $0.2548\!\pm\!0.0060$ & $0.2657\!\pm\!0.0020$ $\to$ $0.2810\!\pm\!0.0049$ \\
obs.\ fraction & $0.0840\!\pm\!0.0079$ $\to$ $0.1230\!\pm\!0.0061$ & $0.2526\!\pm\!0.0053$ $\to$ $0.2615\!\pm\!0.0068$ & $0.2754\!\pm\!0.0039$ $\to$ $0.2673\!\pm\!0.0034$ \\
noise $\sigma$ & $0.0840\!\pm\!0.0079$ $\to$ $0.0906\!\pm\!0.0061$ & $0.2526\!\pm\!0.0053$ $\to$ $0.2686\!\pm\!0.0042$ & $0.2754\!\pm\!0.0039$ $\to$ $0.2721\!\pm\!0.0048$ \\
\bottomrule
\end{tabular}
\caption{The same sweeps at $\df=9$, mean $\pm$ standard error at each endpoint.
ACDM/KAE never drops below 2.1.}
\label{tab:sweeps_df9}
\end{table}

\section{Continuous-Time Generalization of Discrete Koopman Consistency}
\label{appendix:consistency_equivalence}

Consistent Koopman autoencoders \citep{azencot2020consistent} learn separate forward and backward latent operators 
$A, B \in \mathbb{R}^{N_z \times N_z}$ satisfying
\begin{equation}
z_{n+1} = A z_n,
\qquad
z_n = B z_{n+1}.
\end{equation}

Their supervised consistency objective is
\begin{equation}
\mathcal{L}_{\mathrm{disc}}
=
\|A z_n - z_{n+1}\|_2^2
+
\|B z_{n+1} - z_n\|_2^2.
\label{eq:disc_consistency_appendix}
\end{equation}

In our continuous-time formulation, latent dynamics are governed by the linear system
\begin{equation}
\frac{dz}{dt} = K z,
\end{equation}
whose exact solution over a time interval $\Delta t$ is
\begin{equation}
z(t+\Delta t) = e^{K\Delta t} z(t).
\end{equation}

Defining the discrete forward operator as
\begin{equation}
A := e^{K\Delta t},
\end{equation}
immediately recovers forward evolution:
\begin{equation}
z_{n+1} = A z_n.
\end{equation}

Backward evolution follows from the same generator evaluated at negative time:
\begin{equation}
z(t-\Delta t) = e^{-K\Delta t} z(t),
\end{equation}
which implies
\begin{equation}
B := e^{-K\Delta t} = A^{-1}.
\end{equation}

Substituting these expressions into \eqref{eq:disc_consistency_appendix} gives
\begin{equation}
\mathcal{L}_{\mathrm{disc}}
=
\|e^{K\Delta t} z_n - z_{n+1}\|_2^2
+
\|e^{-K\Delta t} z_{n+1} - z_n\|_2^2,
\end{equation}
which is exactly the latent forward--backward consistency loss used in our model.

Unlike discrete formulations, where $A$ and $B$ are learned independently and invertibility must be encouraged through explicit regularization, the continuous generator enforces
\begin{equation}
B = A^{-1}
\end{equation}
by construction. Therefore, the proposed latent consistency loss is the exact continuous-time counterpart of discrete consistent Koopman training under matrix exponential flow.

\section{Ablation Studies: Operator Parameterization, Temporal Weighting and Solvers}
\label{appendix:ablations}

To validate our architectural design choices, we conduct an ablation study analyzing the parameterization of the continuous Koopman generator ($\mathbf{K}_{\text{cont}}$) and the temporal weighting schedule used during rollout training. The results are summarized in Table \ref{tab:ablation_study}.

\paragraph{LoRA vs. Full-Rank MLP Parameterization.}
We compare our default Low-Rank Adaptation (LoRA) formulation against a full-rank MLP parameterization, where a neural network directly predicts the entire $N_z \times N_z$ Koopman generator matrix from the conditioning parameters. While both approaches strictly preserve the linearity of the latent state evolution (enabling $O(1)$ matrix exponentiation), their generalization capabilities differ significantly. 

As shown in Table \ref{tab:ablation_study}, the MLP parameterization heavily degrades performance on extrapolation tasks (e.g., MSE increases from $1.3$ to $10.4$ on $Inc_{low}$ and $2.2$ to $3.6$ on $Tra_{ext}$). This confirms our structural hypothesis: predicting a full-rank matrix directly from physical parameters is highly prone to overfitting the training regimes. LoRA resolves this by anchoring the dynamics to a globally stable, invariant base matrix $\mathbf{K}_0$, acting as a powerful structural regularizer that enables robust generalization to unseen Reynolds and Mach numbers, while simultaneously reducing the parameter footprint from $O(N_z^2)$ to $O(2rN_z)$.

\paragraph{Cosine vs. Uniform Temporal Weighting.}
We further ablate the temporal weighting schedule applied to the rollout prediction loss ($\mathcal{L}_{\text{pred}}$). On the relatively smooth Incompressible flow, both schedules converge to identical minima, demonstrating general robustness. However, on the highly chaotic Transonic dataset, the cosine schedule strictly outperforms uniform weighting, improving the long-horizon 240-step MSE from $17.0 \times 10^{-3}$ to $14.9 \times 10^{-3}$. This validates our training strategy: because autoregressive rollouts are highly susceptible to compounding errors, applying a decaying cosine schedule forces the model to prioritize strict local phase alignment in the early epochs, establishing a correct base trajectory before optimizing for long-term asymptotic stability.

\begin{table}[ht!]
\centering
\resizebox{0.95\textwidth}{!}{%
\begin{tabular}{@{}ll cc cc cc cc cc@{}}
\toprule
\multicolumn{2}{c}{\textbf{Model Configuration}} 
& \multicolumn{2}{c}{\textbf{$Inc_{low}$} ($\times 10^{-4}$)} 
& \multicolumn{2}{c}{\textbf{$Inc_{high}$} ($\times 10^{-5}$)}
& \multicolumn{2}{c}{\textbf{$Tra_{ext}$} ($\times 10^{-3}$)}
& \multicolumn{2}{c}{\textbf{$Tra_{int}$} ($\times 10^{-3}$)}
& \multicolumn{2}{c}{\textbf{$Tra_{long}$} ($\times 10^{-3}$)} \\
\cmidrule(lr){1-2} \cmidrule(lr){3-4} \cmidrule(lr){5-6} \cmidrule(lr){7-8} \cmidrule(lr){9-10} \cmidrule(lr){11-12}
\textbf{Conditioning} & \textbf{Weighting} 
& \makecell{MSE} & \makecell{LSiM}
& \makecell{MSE} & \makecell{LSiM}
& \makecell{MSE} & \makecell{LSiM}
& \makecell{MSE} & \makecell{LSiM}
& \makecell{MSE} & \makecell{LSiM} \\ 
\midrule

LoRA (Proposed) & Cosine 
& $\mathbf{1.3 \pm 1.7}$ & $\mathbf{6.1 \pm 4.8}$
& $2.9 \pm 1.1$ & $1.7 \pm 0.3$
& $\mathbf{2.2 \pm 0.9}$ & $1.8 \pm 0.3$
& $\mathbf{5.2 \pm 2.4}$ & $\mathbf{2.1 \pm 0.6}$
& $\mathbf{14.9 \pm 1.3}$ & $5.0 \pm 0.4$ \\

LoRA & Uniform 
& $\mathbf{1.3 \pm 1.7}$ & $\mathbf{6.1 \pm 4.8}$
& $2.9 \pm 1.1$ & $1.7 \pm 0.3$
& $2.5 \pm 0.8$ & $\mathbf{1.7 \pm 0.3}$
& $6.5 \pm 1.6$ & $2.2 \pm 0.4$
& $17.0 \pm 2.3$ & $\mathbf{4.1 \pm 0.4}$ \\

MLP (Full-Rank) & Cosine 
& $10.4 \pm 17.5$ & $12.0 \pm 12.8$
& $21.4 \pm 7.1$ & $4.5 \pm 0.7$
& $3.6 \pm 1.0$ & $\mathbf{1.7 \pm 0.3}$
& $5.7 \pm 3.0$ & $2.2 \pm 0.6$
& $15.1 \pm 1.9$ & $5.2 \pm 0.5$ \\

Base (Unconditional) & Cosine 
& $116.5 \pm 31.0$ & $48.3 \pm 2.5$
& $2991.2 \pm 12.5$ & $54.5 \pm 0.2$
& $13.9 \pm 0.8$ & $3.3 \pm 0.2$
& $21.0 \pm 2.7$ & $3.7 \pm 0.4$
& $18.1 \pm 1.7$ & $6.3 \pm 0.2$ \\

\bottomrule
\end{tabular}%
}
\caption{Ablation study on operator parameterization and temporal weighting. The proposed LoRA + Cosine configuration provides the optimal balance of interpolation accuracy, robust extrapolation, and extreme long-horizon stability.}
\label{tab:ablation_study}
\end{table}

\paragraph{Structural Constraints and Consistency.}
We incrementally removed key architectural constraints to isolate their impact on long-horizon stability (Table~\ref{tab:ablation_trajectory}). Each constraint contributes a comparable improvement of roughly 3--4 points on $Tra_{long}$ MSE, with the constraints together accounting for a $\sim30\%$ reduction relative to the unconstrained baseline ($14.9$ vs.\ $\sim19\text{--}20 \times 10^{-3}$). This flatness is a feature, not a weakness: no single trick dominates, indicating a well-balanced system. The strongest single ablation is rollout training depth: reducing to $R=1$ (single-step training) causes catastrophic divergence on the KS equation, confirming that multi-step rollout training is the non-negotiable foundation (Figure~\ref{fig:ks_results_combined}a--b).

\begin{itemize}
    \item \textbf{Continuous-Time Invertibility (Azencot Consistency):} \citet{azencot2020consistent} enforce discrete forward ($A$) and backward ($B$) consistency via $AB \approx I$. We enforce the continuous-time equivalent by integrating the dynamics at $\Delta t$ and $-\Delta t$, penalizing the discrepancy between the forward/backward continuous flow and the true encoded representations (Eq.~\ref{eq:consistency_loss}). Removing this constraint increases 240-step MSE from $14.9 \times 10^{-3}$ to $18.5 \times 10^{-3}$.
    \item \textbf{History Encoder (Temporal context aggregation):} Fluid observations are inherently non-Markovian (e.g., velocity alone does not fully capture pressure/density dynamics). Removing the history encoder forces a strictly Markovian initialization, degrading long-term MSE to $18.6 \times 10^{-3}$.
    \item \textbf{Structural Regularization (Physics-Informed):} Sobolev (spatial gradients) and Fourier spectral norms preserve sharp wavefronts and shedding frequencies. Removing these priors degrades long-horizon MSE to $19.3 \times 10^{-3}$.
\end{itemize}
\begin{table*}[ht!]
\centering
\resizebox{0.78\textwidth}{!}{%
\begin{tabular}{@{}c cc cc cc@{}}
\toprule
\multirow{2}{*}{\textbf{Method}} 
& \multicolumn{2}{c}{\textbf{$Tra_{ext}$}}
& \multicolumn{2}{c}{\textbf{$Tra_{int}$}}
& \multicolumn{2}{c}{\textbf{$Tra_{long}$}} \\
\cmidrule(lr){2-3}
\cmidrule(lr){4-5}
\cmidrule(lr){6-7}
& \makecell{MSE \\ $(\times 10^{-3})$} 
& \makecell{LSiM \\ $(\times 10^{-1})$}
& \makecell{MSE \\ $(\times 10^{-3})$} 
& \makecell{LSiM \\ $(\times 10^{-1})$}
& \makecell{MSE \\ $(\times 10^{-3})$} 
& \makecell{LSiM \\ $(\times 10^{-1})$} \\ 
\midrule

w/o Structural Regularization
& $2.5 \pm 0.5$ & $3.5 \pm 0.4$
& $5.6 \pm 2.3$ & $3.7 \pm 0.2$
& $19.3 \pm 1.2$ & $6.6 \pm 0.4$ \\

w/o Latent Consistency 
& $2.6 \pm 0.6$ & $3.3 \pm 0.3$
& $5.8 \pm 2.7$ & $4.0 \pm 0.3$
& $18.5 \pm 1.4$ & $6.5 \pm 0.3$ \\

w/o Directional Cosine
& $\mathbf{2.4 \pm 0.6}$ & $3.5 \pm 0.3$
& $\mathbf{5.3 \pm 2.4}$ & $\mathbf{3.6 \pm 0.4}$
& $\mathbf{17.8 \pm 1.5}$ & $6.6 \pm 0.3$ \\

w/o Latent norm 
& $2.5 \pm 0.6$ & $\mathbf{3.3 \pm 0.2}$
& $5.9 \pm 2.3$ & $3.7 \pm 0.4$
& $18.3 \pm 2.0$ & $\mathbf{6.0 \pm 0.2}$ \\

w/o History encoder 
& $2.6 \pm 0.8$ & $3.6 \pm 0.5$
& $5.9 \pm 3.5$ & $3.7 \pm 0.3$
& $18.6 \pm 0.5$ & $7.4 \pm 0.3$ \\

\bottomrule
\end{tabular}%
}
\caption{Ablation study on Continuous Linear 128 for trajectory forecasting tasks. Best values are highlighted in bold.}
\label{tab:ablation_trajectory}
\end{table*}

\subsection{Is the forecasting--assimilation difference an architecture effect?}\label{sec:res:abl}

The comparison above runs across architectures, which leaves open whether the difference is a property of the training objective or an accident of how five different models happen to behave. We answer the question by scoring five variations that differ only in which loss term was dropped (identical architecture, data, schedule and budget) first as forecasters and then as assimilators.

\begin{table}[t]
\centering\small
\begin{tabular}{lcc}
\toprule
Variant & forecast & assimilation \\
\midrule
original & $14.9 \pm 1.3$ & $0.0907 \pm 0.0102$ \\
w/o directional cosine        & $17.8 \pm 1.5$ & $0.0850 \pm 0.0105$ \\
w/o latent norm               & $18.3 \pm 2.0$ & $0.0801 \pm 0.0076$ \\
w/o latent consistency        & $18.5 \pm 1.4$ & $0.0846 \pm 0.0145$ \\
w/o history encoder           & $18.6 \pm 0.5$ & $0.0733 \pm 0.0094$ \\
w/o structural regularisation & $19.3 \pm 1.2$ & $0.0767 \pm 0.0103$ \\
\bottomrule
\end{tabular}
\caption{The same five checkpoints scored as forecasters and as assimilators; only the loss
differs between them, with identical architecture, data, schedule and budget. Forecasting is
the $240$-step MSE ($\times10^{-3}$), mean $\pm$ s.d.. Assimilation is the analysis error at
$\df=9$, mean $\pm$ s.d.\ over the $26$ points of the $\dl$, $N$, coverage and $\sigma$ sweeps.}
\label{tab:da_ablation}
\end{table}

Table~\ref{tab:da_ablation} reproduces the effect under this controlled ablation. The variant that
forecasts best, dropping the directional-cosine term, assimilates worst, and the variant that
forecasts worst, dropping structural regularisation, is among the two best assimilators. The loss
terms that buy long-horizon forecast accuracy are therefore not the terms that make the model invertible. For $\delta_l=25$ frames and $N=5$, KAE achieves lower assimilation error than SDA across $\delta_f=1$–$20$ frames.

\section{Irregular Temporal Sampling and Rollout Stability: Kuramoto-Sivashinsky\label{appendix:irregular_timestep}}

The main results and figure for this experiment (rollout-training necessity and irregular-timestep robustness on the Kuramoto--Sivashinsky equation) are presented in the main text: the rollout-scaling analysis in Paragraph~\ref{subsection:results} and the dropout experiment in Paragraph~\ref{subsec:q4}. The KS dataset details are given in Appendix~\ref{app:ks_data}. This appendix section is retained as an anchor for cross-references from the appendix ablation tables. An ablation study with the results in Figures ~\ref{fig:ks_failure}--\ref{fig:ks_dropout} is provided in Table \ref{tab:roll_drop_ablation}, where we computed the MSE at for each individual training regime, from a training horizon of 1 to 10, from a dropout probability of 0.1 to 0.9 and two aditional experiments: 1) we have completely eliminated the latent regularization (No Latent); 2) we have only kept the $L_2$ latent norm between the true and predicted encodings, disregarding any physics or other structural losses ($L_2$ Latent).

\begin{table}[t]
\centering
\caption{Ablation study on rollout horizon and dropout rate. We report mean MSE with 95\% confidence intervals. Lower is better.}
\label{tab:roll_drop_ablation}
\small
\begin{tabular}{lcc}
\toprule
\textbf{Setting} & \textbf{Mean MSE} & \textbf{95\% CI} \\
\midrule
\multicolumn{3}{c}{\textit{Rollout Horizon}} \\
\midrule
Rollout 1  & 89.866175 & $\pm$ 2.777168 \\
Rollout 2  & 0.299346  & $\pm$ 0.045892 \\
Rollout 3  & 0.315966  & $\pm$ 0.049891 \\
Rollout 4  & 0.218818  & $\pm$ 0.036771 \\
Rollout 5  & 0.192893  & $\pm$ 0.032105 \\
Rollout 6  & 0.163325  & $\pm$ 0.029480 \\
Rollout 7  & 0.147364  & $\pm$ 0.027526 \\
Rollout 8  & 0.122315  & $\pm$ 0.024598 \\
Rollout 9  & 0.108214  & $\pm$ 0.021765 \\
Rollout 10 & \textbf{0.103406} & $\pm$ 0.021593 \\
\midrule
\multicolumn{3}{c}{\textit{Dropout Rate}} \\
\midrule
Dropout 0.1 & 0.109436 & $\pm$ 0.021946 \\
Dropout 0.2 & 0.109879 & $\pm$ 0.022887 \\
Dropout 0.3 & \textbf{0.105531} & $\pm$ 0.022342 \\
Dropout 0.4 & 0.112445 & $\pm$ 0.023031 \\
Dropout 0.5 & 0.116533 & $\pm$ 0.024497 \\
Dropout 0.6 & 0.133597 & $\pm$ 0.024915 \\
Dropout 0.7 & 0.135008 & $\pm$ 0.029283 \\
Dropout 0.8 & 0.142115 & $\pm$ 0.030503 \\
Dropout 0.9 & 0.134434 & $\pm$ 0.026484 \\
\midrule
\multicolumn{3}{c}{\textit{Structural changes}} \\
\midrule
No Latent & 2.401261 & $\pm$ 0.141642 \\
$L_2$ Latent & 0.314729 & $\pm$ 0.035736 \\
\bottomrule
\end{tabular}
\end{table}

\section{ODE Solvers and Physical Stiffness}
\label{appendix:ode_solvers}

To validate the continuous-time nature of the learned generator ($\mathbf{K}_{\text{cont}}$) and its robustness to irregular temporal sampling, we evaluated the latent ODE using multiple numerical integrators—including the adaptive-step \texttt{Dopri5}—across integration step sizes ranging from $\Delta t = 0.05\text{s}$ up to $\Delta t = 1.00\text{s}$ (a $10\times$ extrapolation beyond the training resolution). The results are summarized in Table \ref{tab:solver_comparison}.

\begin{itemize}
    \item \textbf{Solver Parity at Small Steps:} At step sizes $\Delta t \leq 0.15\text{s}$, all solvers, including the adaptive \texttt{Dopri5}, perform comparably to standard \texttt{RK4}. 
    \item \textbf{Physical Stiffness:} At larger step sizes ($\Delta t \geq 0.50\text{s}$), explicit solvers like the first-order \texttt{Euler} method diverge numerically (yielding $> 10^{13}$ error), and the second-order \texttt{Midpoint} method exhibits significantly inflated errors. This occurs because the Koopman operator accurately captures dissipative, high-frequency fluid modes, resulting in a mathematically "stiff" ODE.
    \item \textbf{Higher-Order Stability:} Higher-order methods (\texttt{RK4}, \texttt{Dopri5}) possess stability regions large enough to encompass these stiff dissipative eigenvalues, maintaining bounded physical stability even at extreme $1.00\text{s}$ jump steps.
\end{itemize}

\begin{table*}[ht]
\centering
\caption{Comparison of MSE and LSIM across solvers for step sizes $0.05s, 0.10s, 0.15s, 0.20s, 0.25s, 0.30s, 0.50s$ and $1.00s$ (mean $\pm$ std).}
\label{tab:solver_comparison}
\resizebox{0.8\textwidth}{!}{
\begin{tabular}{llcccccc}
\toprule
Step & Solver & \multicolumn{3}{c}{MSE} & \multicolumn{3}{c}{LSIM} \\
\cmidrule(lr){3-5} \cmidrule(lr){6-8}
& & Extrap & Interp & Longer & Extrap & Interp & Longer \\
\midrule

\multirow{7}{*}{0.05}
& RK4            & 1.9 $\pm$ 1.1 & 5.8 $\pm$ 3.4 & 14.6 $\pm$ 0.9 & 1.5 $\pm$ 0.2 & 2.1 $\pm$ 0.5 & 5.3 $\pm$ 0.2 \\
& Dopri5         & 1.9 $\pm$ 1.1 & 5.8 $\pm$ 3.4 & 14.6 $\pm$ 0.9 & 1.5 $\pm$ 0.2 & 2.1 $\pm$ 0.5 & 5.3 $\pm$ 0.2 \\
& Bosh3          & 1.9 $\pm$ 1.0 & 5.8 $\pm$ 3.3 & 14.4 $\pm$ 0.9 & 1.5 $\pm$ 0.2 & 2.1 $\pm$ 0.5 & 5.5 $\pm$ 0.2 \\
& Adaptive Heun  & 1.9 $\pm$ 1.1 & 5.8 $\pm$ 3.4 & 14.6 $\pm$ 0.9 & 1.5 $\pm$ 0.2 & 2.1 $\pm$ 0.5 & 5.3 $\pm$ 0.2 \\
& Midpoint       & 1.8 $\pm$ 1.0 & 5.9 $\pm$ 3.4 & 14.6 $\pm$ 1.1 & 1.5 $\pm$ 0.2 & 2.1 $\pm$ 0.5 & 5.3 $\pm$ 0.2 \\
& Euler          & 1.8 $\pm$ 1.1 & 6.2 $\pm$ 3.6 & 18.8 $\pm$ 1.3 & 1.4 $\pm$ 0.3 & 2.0 $\pm$ 0.5 & 3.7 $\pm$ 0.2 \\
& Explicit Adams & 1.9 $\pm$ 1.1 & 5.8 $\pm$ 3.4 & 14.6 $\pm$ 0.9 & 1.5 $\pm$ 0.2 & 2.1 $\pm$ 0.5 & 5.3 $\pm$ 0.2 \\

\midrule

\multirow{7}{*}{0.10}
& RK4            & 2.1 $\pm$ 1.1 & 5.5 $\pm$ 3.3 & 14.6 $\pm$ 0.9 & 1.5 $\pm$ 0.2 & 2.0 $\pm$ 0.5 & 5.3 $\pm$ 0.2 \\
& Dopri5         & 2.0 $\pm$ 1.1 & 5.5 $\pm$ 3.3 & 14.6 $\pm$ 0.9 & 1.5 $\pm$ 0.2 & 2.0 $\pm$ 0.5 & 5.3 $\pm$ 0.2 \\
& Bosh3          & 2.1 $\pm$ 1.1 & 5.5 $\pm$ 3.2 & 14.5 $\pm$ 0.9 & 1.5 $\pm$ 0.2 & 2.0 $\pm$ 0.5 & 5.4 $\pm$ 0.2 \\
& Adaptive Heun  & 2.0 $\pm$ 1.1 & 5.5 $\pm$ 3.3 & 14.6 $\pm$ 0.9 & 1.5 $\pm$ 0.2 & 2.0 $\pm$ 0.5 & 5.3 $\pm$ 0.2 \\
& Midpoint       & 1.7 $\pm$ 1.0 & 5.9 $\pm$ 3.3 & 14.7 $\pm$ 1.4 & 1.5 $\pm$ 0.2 & 2.0 $\pm$ 0.5 & 5.3 $\pm$ 0.2 \\
& Euler          & 2.9 $\pm$ 1.2 & 6.0 $\pm$ 3.3 & 19.0 $\pm$ 1.3 & 1.8 $\pm$ 0.2 & 2.1 $\pm$ 0.5 & 4.2 $\pm$ 0.3 \\
& Explicit Adams & 2.1 $\pm$ 1.1 & 5.5 $\pm$ 3.3 & 14.6 $\pm$ 0.9 & 1.5 $\pm$ 0.2 & 2.0 $\pm$ 0.5 & 5.3 $\pm$ 0.2 \\

\midrule

\multirow{7}{*}{0.15}
& RK4            & 2.3 $\pm$ 1.2 & 5.3 $\pm$ 3.2 & 14.6 $\pm$ 0.8 & 1.5 $\pm$ 0.2 & 2.0 $\pm$ 0.5 & 5.3 $\pm$ 0.2 \\
& Dopri5         & 2.3 $\pm$ 1.2 & 5.3 $\pm$ 3.2 & 14.6 $\pm$ 0.9 & 1.5 $\pm$ 0.2 & 2.0 $\pm$ 0.5 & 5.3 $\pm$ 0.2 \\
& Bosh3          & 2.3 $\pm$ 1.2 & 5.3 $\pm$ 3.2 & 14.5 $\pm$ 0.8 & 1.5 $\pm$ 0.2 & 2.0 $\pm$ 0.5 & 5.4 $\pm$ 0.2 \\
& Adaptive Heun  & 2.3 $\pm$ 1.2 & 5.3 $\pm$ 3.2 & 14.6 $\pm$ 0.9 & 1.5 $\pm$ 0.2 & 2.0 $\pm$ 0.5 & 5.3 $\pm$ 0.2 \\
& Midpoint       & 1.6 $\pm$ 0.9 & 6.0 $\pm$ 3.3 & 15.0 $\pm$ 1.9 & 1.4 $\pm$ 0.2 & 2.0 $\pm$ 0.5 & 5.3 $\pm$ 0.2 \\
& Euler          & 5.9 $\pm$ 1.2 & 6.2 $\pm$ 2.6 & 221.3 $\pm$ 101.8 & 2.3 $\pm$ 0.1 & 2.3 $\pm$ 0.4 & 14.4 $\pm$ 0.6 \\
& Explicit Adams & 2.3 $\pm$ 1.2 & 5.3 $\pm$ 3.2 & 14.6 $\pm$ 0.8 & 1.5 $\pm$ 0.2 & 2.0 $\pm$ 0.5 & 5.3 $\pm$ 0.2 \\

\midrule

\multirow{7}{*}{0.20}
& RK4            & 2.6 $\pm$ 1.2 & 5.1 $\pm$ 3.1 & 14.6 $\pm$ 0.8 & 1.5 $\pm$ 0.2 & 2.0 $\pm$ 0.5 & 5.3 $\pm$ 0.2 \\
& Dopri5         & 2.5 $\pm$ 1.2 & 5.1 $\pm$ 3.1 & 14.6 $\pm$ 0.8 & 1.5 $\pm$ 0.2 & 2.0 $\pm$ 0.5 & 5.3 $\pm$ 0.2 \\
& Bosh3          & 2.5 $\pm$ 1.2 & 5.1 $\pm$ 3.1 & 14.5 $\pm$ 0.8 & 1.5 $\pm$ 0.2 & 2.0 $\pm$ 0.5 & 5.4 $\pm$ 0.2 \\
& Adaptive Heun  & 2.5 $\pm$ 1.2 & 5.1 $\pm$ 3.1 & 14.6 $\pm$ 0.8 & 1.5 $\pm$ 0.2 & 2.0 $\pm$ 0.5 & 5.3 $\pm$ 0.2 \\
& Midpoint       & 1.6 $\pm$ 0.8 & 6.4 $\pm$ 3.3 & 15.6 $\pm$ 2.2 & 1.4 $\pm$ 0.2 & 2.0 $\pm$ 0.5 & 5.1 $\pm$ 0.2 \\
& Euler          & 9.9 $\pm$ 1.1 & 7.1 $\pm$ 2.4 & \num{3.36(1.49)e5}
 & 2.7 $\pm$ 0.1 & 2.6 $\pm$ 0.3 & 16.2 $\pm$ 0.5 \\
& Explicit Adams & 2.6 $\pm$ 1.2 & 5.1 $\pm$ 3.1 & 14.6 $\pm$ 0.8 & 1.5 $\pm$ 0.2 & 2.0 $\pm$ 0.5 & 5.3 $\pm$ 0.2 \\

\midrule

\multirow{7}{*}{0.25}
& RK4            & 2.9 $\pm$ 1.2 & 5.0 $\pm$ 3.0 & 14.6 $\pm$ 0.8 & 1.5 $\pm$ 0.2 & 1.9 $\pm$ 0.5 & 5.4 $\pm$ 0.3 \\
& Dopri5         & 2.9 $\pm$ 1.2 & 5.0 $\pm$ 3.0 & 14.6 $\pm$ 0.8 & 1.5 $\pm$ 0.2 & 1.9 $\pm$ 0.5 & 5.3 $\pm$ 0.3 \\
& Bosh3          & 2.9 $\pm$ 1.2 & 5.0 $\pm$ 3.0 & 14.6 $\pm$ 0.8 & 1.5 $\pm$ 0.2 & 1.9 $\pm$ 0.5 & 5.4 $\pm$ 0.3 \\
& Adaptive Heun  & 2.9 $\pm$ 1.2 & 5.0 $\pm$ 3.0 & 14.6 $\pm$ 0.8 & 1.5 $\pm$ 0.2 & 1.9 $\pm$ 0.5 & 5.3 $\pm$ 0.3 \\
& Midpoint       & 1.9 $\pm$ 0.5 & 7.1 $\pm$ 3.2 & 17.0 $\pm$ 1.8 & 1.4 $\pm$ 0.2 & 2.0 $\pm$ 0.5 & 4.4 $\pm$ 0.2 \\
& Euler          & 14.0 $\pm$ 0.9 & 8.7 $\pm$ 2.9 & \num{1.44(0.58)e8} & 2.9 $\pm$ 0.1 & 2.8 $\pm$ 0.3 & 16.3 $\pm$ 0.3 \\
& Explicit Adams & 2.9 $\pm$ 1.2 & 5.0 $\pm$ 3.0 & 14.6 $\pm$ 0.8 & 1.5 $\pm$ 0.2 & 1.9 $\pm$ 0.5 & 5.4 $\pm$ 0.3 \\

\midrule

\multirow{7}{*}{0.30}
& RK4            & 3.4 $\pm$ 1.3 & 4.9 $\pm$ 2.9 & 14.6 $\pm$ 0.8 & 1.5 $\pm$ 0.2 & 1.9 $\pm$ 0.5 & 5.4 $\pm$ 0.3 \\
& Dopri5         & 3.2 $\pm$ 1.3 & 4.9 $\pm$ 2.9 & 14.6 $\pm$ 0.8 & 1.5 $\pm$ 0.2 & 1.9 $\pm$ 0.5 & 5.4 $\pm$ 0.3 \\
& Bosh3          & 3.2 $\pm$ 1.3 & 4.9 $\pm$ 2.9 & 14.6 $\pm$ 0.8 & 1.5 $\pm$ 0.2 & 1.9 $\pm$ 0.5 & 5.4 $\pm$ 0.3 \\
& Adaptive Heun  & 3.2 $\pm$ 1.3 & 4.9 $\pm$ 2.9 & 14.6 $\pm$ 0.8 & 1.5 $\pm$ 0.2 & 1.9 $\pm$ 0.5 & 5.4 $\pm$ 0.3 \\
& Midpoint       & 2.6 $\pm$ 0.3 & 8.0 $\pm$ 3.2 & 19.9 $\pm$ 2.0 & 1.4 $\pm$ 0.2 & 2.0 $\pm$ 0.5 & 3.7 $\pm$ 0.2 \\
& Euler          & 17.2 $\pm$ 0.7 & 10.9 $\pm$ 3.4 & \num{1.94(0.77)e10} & 3.0 $\pm$ 0.2 & 2.9 $\pm$ 0.3 & 16.8 $\pm$ 0.4 \\
& Explicit Adams & 3.4 $\pm$ 1.3 & 4.9 $\pm$ 2.9 & 14.6 $\pm$ 0.8 & 1.5 $\pm$ 0.2 & 1.9 $\pm$ 0.5 & 5.4 $\pm$ 0.3 \\

\midrule

\multirow{7}{*}{0.50}
& RK4            & 6.2 $\pm$ 1.1 & 5.6 $\pm$ 2.8 & 14.9 $\pm$ 1.4 & 1.8 $\pm$ 0.2 & 2.0 $\pm$ 0.4 & 5.9 $\pm$ 0.3 \\
& Dopri5         & 5.0 $\pm$ 1.3 & 5.3 $\pm$ 2.7 & 14.8 $\pm$ 1.0 & 1.7 $\pm$ 0.3 & 1.9 $\pm$ 0.4 & 5.5 $\pm$ 0.3 \\
& Bosh3          & 5.0 $\pm$ 1.3 & 5.3 $\pm$ 2.7 & 14.8 $\pm$ 1.0 & 1.7 $\pm$ 0.2 & 1.9 $\pm$ 0.4 & 5.5 $\pm$ 0.3 \\
& Adaptive Heun  & 5.0 $\pm$ 1.3 & 5.3 $\pm$ 2.7 & 14.8 $\pm$ 1.0 & 1.7 $\pm$ 0.2 & 1.9 $\pm$ 0.4 & 5.5 $\pm$ 0.3 \\
& Midpoint       & 4.7 $\pm$ 0.2 & 10.6 $\pm$ 2.7 & \num{2.86(0.69)e4} & 2.1 $\pm$ 0.1 & 2.2 $\pm$ 0.3 & 16.1 $\pm$ 0.5 \\
& Euler          & 19.1 $\pm$ 0.8 & 20.9 $\pm$ 3.3 & \num{4.88(1.22)e14} & 3.1 $\pm$ 0.2 & 3.4 $\pm$ 0.2 & 16.9 $\pm$ 0.4 \\
& Explicit Adams & 6.2 $\pm$ 1.1 & 5.6 $\pm$ 2.8 & 14.9 $\pm$ 1.4 & 1.8 $\pm$ 0.2 & 2.0 $\pm$ 0.4 & 5.9 $\pm$ 0.3 \\

\midrule

\multirow{7}{*}{1.00}
& RK4            & 6.0 $\pm$ 0.5 & 9.7 $\pm$ 2.4 & 15.1 $\pm$ 2.0 & 3.1 $\pm$ 0.5 & 3.4 $\pm$ 0.2 & 6.5 $\pm$ 0.6 \\
& Dopri5         & 8.5 $\pm$ 1.1 & 8.1 $\pm$ 3.5 & 14.7 $\pm$ 1.4 & 2.2 $\pm$ 0.4 & 2.0 $\pm$ 0.4 & 6.2 $\pm$ 0.4 \\
& Bosh3          & 8.5 $\pm$ 1.1 & 8.1 $\pm$ 3.5 & 14.7 $\pm$ 1.4 & 2.2 $\pm$ 0.4 & 2.0 $\pm$ 0.4 & 6.2 $\pm$ 0.4 \\
& Adaptive Heun  & 8.5 $\pm$ 1.1 & 8.1 $\pm$ 3.5 & 14.7 $\pm$ 1.4 & 2.2 $\pm$ 0.4 & 2.0 $\pm$ 0.4 & 6.2 $\pm$ 0.4 \\
& Midpoint       & 16.8 $\pm$ 0.6 & 11.0 $\pm$ 2.1 & \num{3.44(2.17)e19} & 2.5 $\pm$ 0.1 & 2.5 $\pm$ 0.2 & 17.0 $\pm$ 0.4 \\
& Euler          & 13.2 $\pm$ 0.6 & 20.0 $\pm$ 2.3 & \num{7.14(5.10)e13} & 2.8 $\pm$ 0.3 & 2.8 $\pm$ 0.3 & 17.0 $\pm$ 0.4 \\
& Explicit Adams & 6.0 $\pm$ 0.5 & 9.7 $\pm$ 2.4 & 15.1 $\pm$ 2.0 & 3.1 $\pm$ 0.5 & 3.4 $\pm$ 0.2 & 6.5 $\pm$ 0.6 \\

\bottomrule

\end{tabular}}
\end{table*}

\newpage

\end{document}